\documentclass[runningheads]{llncs}
\usepackage[T1]{fontenc}
\usepackage{graphicx}
\usepackage{booktabs}
\usepackage[misc]{ifsym}
\newcommand{\corr}{(\Letter)}
\usepackage{mwe}

\usepackage{xurl}
\usepackage{enumitem}
\usepackage{placeins}
\usepackage{booktabs}
\usepackage{wrapfig}
\usepackage{float}
\usepackage{amsmath}
\usepackage{hyperref}
\usepackage{mathtools}
\usepackage[caption=false]{subfig}
\usepackage{chngcntr}
\usepackage{apptools}
\usepackage{amssymb}
\usepackage{xr}

\usepackage{cite}

\begin{document}

\title{Inspecting Training Dynamics of Similarity Development in Supervised Vision Networks}

\titlerunning{Training Dynamics of Similarity Development}

\author{Katarzyna Filus\inst{1} \corr \and
Mateusz Żarski\inst{1}}
\authorrunning{K. Filus and M. Żarski}

\institute{Institute of Theoretical and Applied Informatics, Polish Academy of Sciences, Gliwice, Poland, \email{kfilus@iitis.pl}}

\maketitle              

\begin{abstract}
For trustworthy and human-aware artificial intelligence, models should be evaluated beyond accuracy, among others through error predictability and semantic alignment. Similarity is central to these aspects, as it influences which classes a model considers related and confusable. Similarity manifests in multiple forms, including semantic similarity, which can serve as a proxy for human similarity perception. While similarity perception is often imposed in computer vision, little attention has been paid to its natural emergence during supervised training. Existing studies are largely limited to static and qualitative analyzes and lack a systematic, training-time perspective. Therefore, we analyze how similarity perception evolves and aligns with model error patterns and semantics in supervised vision networks. As an enabler, we introduce Deep Similarity Inspector (DSI) -- a systematic, training-time framework that unifies complementary views on similarity within a single methodology. Using DSI, we analyzed Convolutional and Transformer-based Networks and showed that both architectures develop rich similarity structures through three phases -- initial similarity surge, refinement, stabilization -- while exhibiting clear differences. We also identified the \textit{mistakes refinement phenomenon}, in which networks improve mistakes with time.

\end{abstract}

\section{Introduction}

Similarity aids categorization due to the correlated structure of the world \cite{bib:medin1993respects}. This makes similarity relevant for human-aware and trustworthy artificial intelligence, where accuracy evaluation should be complemented by analyzes of network perception and error severity. Similarity can be expressed through different notions, including perceptual and semantic similarity. While semantic similarity is often discussed in linguistics, it is also fundamentally grounded in visual similarities of the world, stemming from shared attributes, common functionality, or evolutionary traits, and thus is often treated as a reference \cite{bib:bertinetto2020making,bib:mopuri2020adversarial}. Although similarity's importance is recognized in machine perception as something to be imposed to support human-aware alignment \cite{bib:bertinetto2020making,bib:bilal2017convolutional,bib:chen2020simple,bib:caron2021emerging}, little attention has been paid to natural emergence of similarity during standard supervised training.

Similarity-based evaluation initially sparked some interest in computer vision, but rapid accuracy gains were prioritized within a homogeneous early model landscape \cite{bib:russakovsky2015imagenet,bib:bertinetto2020making}. Nevertheless, recent studies have begun to revisit similarity analysis \cite{bib:huang2021semantic,bib:bilal2017convolutional,bib:bertinetto2020making,bib:filus2025ecml}. The existing works provide only a limited view on similarity perception in networks, by focusing mostly on individual checkpoints rather than how similarity evolves during training \cite{bib:bilal2017convolutional,bib:bertinetto2020making,bib:filus2025ecml}. Aside from minor qualitative assessments \cite{bib:huang2021semantic}, no quantitative and systematic training analysis was performed. 
Moreover, prior work adopts a single perspective on similarity, without jointly leveraging complementary views -- direct and indirect   \cite{bib:medin1993respects} -- and is restricted to a functional view based on large-scale datasets, overlooking structural probing \cite{bib:mopuri2020adversarial,bib:filus2023netsat}. Also, Vision Transformers are mostly ignored \cite{bib:huang2021semantic,bib:bilal2017convolutional,bib:bertinetto2020making}.
Our previous work \cite{bib:filus2025ecml} aimed to solve some of these deficiencies, performing a first large-scale analysis of heterogeneous pretrained vision models, and examining their alignment with semantics and each other. While it represents a step forward, it remains limited to post-training, single-perspective analysis and does not address how similarity develops or aligns with model errors and semantics during training. Consequently, a systematic, multi-perspective inspection of similarity training dynamics remains an open challenge.

Motivated by these gaps, we conduct the first systematic training-time analysis of supervised vision networks, examining how similarity perception evolves and aligns with model error patterns and human semantics. To enable this analysis, we introduce \textit{\textbf{Deep Similarity Inspector (DSI)}}, which defines a structured protocol that unifies functional and structural, direct and indirect similarity estimation approaches into a coherent, training-time evaluation methodology. It represents a significant departure from the available, predominantly qualitative, approaches, offering a systematic and scalable solution. Using DSI, we provide new insights into the dynamics of supervised learning, contributing to a deeper understanding of model knowledge development, error behavior, and complementing accuracy evaluation.
Our key contributions are: \textbf{(1)} Definition of a structured, systematic framework for similarity-focused vision networks' training examination. \textbf{(2)} Formulation and unification of numerical metrics to examine how network's perception is developed during training, and aligns with semantics and error behavior. Using all the direct, indirect, functional and structural approaches to similarity measurements. \textbf{ (3)} Thorough examination and comparison of the training process of CNNs, ViTs and hybrid models from the similarity perception perspective and its alignment with errors and semantic similarity. We provide our code and reproducibility details at \url{zenodo.org/records/14764887}.

\section{Related Work}

Similarity is important in computer vision~\cite{bib:tang2017visual,bib:nayak2019zero,bib:muttenthaler2024improving,bib:chen2020simple} and spans multiple notions~\cite{bib:veit2017conditional}: visual,  representational \cite{bib:kornblith2019similarity}, contextual~\cite{bib:shi2019not}, adversarial~\cite{bib:elsayed2018adversarial,bib:filus2024similarity}, and semantic~\cite{bib:pedersen2004wordnet}. Vision methods often include similarity enforcement into training,~\emph{e.g.} via hierarchy enforcement~\cite{bib:bertinetto2020making,bib:bilal2017convolutional} or contrastive learning~\cite{bib:chen2020simple,bib:caron2021emerging,bib:he2020momentum}.
Network representations are aligned to similarity references such as human judgments~\cite{bib:muttenthaler2024improving,bib:roads2021enriching,bib:geirhos2021partial,bib:geirhos2020beyond} or semantic structure~\cite{bib:bilal2017convolutional}, typically via stimuli-based probes~\cite{bib:huang2021semantic,bib:kornblith2019similarity,bib:kriegeskorte2008rdm,bib:williamsequivalence}. In contrast, our goal is not to enforce a certain similarity behavior, but to examine how similarity perception emerges \emph{naturally} in standard supervised training, and how it aligns with semantic similarity and network mistakes. We use semantic similarity as a reference proxy for human similarity perception that is a dataset-agnostic baseline not relying on human polls~\cite{bib:geirhos2021partial,bib:geirhos2020beyond}, stemming from meaningful relations such as shared functionality or evolutionary traits, which are reflected in visual similarity through natural visual--semantic links~\cite{bib:deselaers2011visual}.

Existing semantic-alignment analyses are limited, mostly focusing on individual model checkpoints \cite{bib:bilal2017convolutional,bib:bertinetto2020making,bib:filus2025ecml}. Aside from minor qualitative assessments \cite{bib:huang2021semantic}, no quantitative and systematic training dynamics analysis was performed. Prior work adopts a single perspective on similarity, without jointly leveraging complementary estimation approaches  \cite{bib:medin1993respects}, and is restricted to a functional view based on large-scale datasets, overlooking structural probing  \cite{bib:mopuri2020adversarial,bib:filus2023netsat}. Vision Transformers are mostly ignored \cite{bib:huang2021semantic,bib:bilal2017convolutional,bib:bertinetto2020making}.
Our recent study~\cite{bib:filus2025ecml} moved toward a broader analysis by comparing heterogeneous pretrained vision models and their alignment with semantics and each other. However, it remains limited to post-training and a single similarity view. In contrast, current work unifies complementary views of similarity---functional and structural, direct and indirect---into a single training-time framework, enabling a systematic study of how similarity perception develops and relates to semantics and model errors during learning.

\section{Deep Similarity Inspector}

A core framework's data structure is a Class Similarity Matrix (CSM). It serves as a unifying representation of different notions of class similarity derived from substantially different data sources, and thus making them directly comparable within a common space. Let $ C = \{c_1, c_2, \dots, c_N\} $ be the set of classes in a given dataset $\mathcal{D}$ with $ N $ classes of model $\mathcal{M}$. $ CSM $ is an $ N \times N $ matrix that stores the pairwise similarities between classes. Each element $ CSM_{ij} $ of the matrix quantifies the similarity between classes $ c_i $ and $ c_j $ ($ CSM_{ii} = 1 $ (diagonal is excluded). We can define its 3 variants: Network Class Similarity Matrices (NCSMs), Confusion-based Class Similarity Matrices (CCSMs) and Semantic Class Similarity Matrices (SCSMs)

\textbf{Network Class Similarity Matrices (NCSMs)} represents a \textit{direct} and \textit{structural} approach to similarity estimation from the perspective of $\mathcal{M}$. We use the learned class representations from a final classifier of a deep network $\mathcal{M}$ \cite{bib:filus2025ecml,bib:mopuri2020adversarial,bib:nayak2019zero}. Each neuron $c$ of this layer corresponds to one of the classes - $c$. A vector $w_c$ of weights connecting neuron $c$ to the penultimate layer can be treated as a class template (representation in the layer's feature space) of class $c$. $w_c$ is represented as
    $w_c = [w_{c1}, w_{c2}, \ldots, w_{cn}]$,
    where each $w_{ci}$ corresponds to the weight connecting the $c$-th neuron to the $i$-th penultimate layer neuron. Similarity between classes $i$ and $j$ is computed with cosine similarity:
    $\text{CS}(i, j) = \frac{w_i^T w_j}{||w_i||||w_j||}$, and this becomes the value of $NCSM(i, j)$. We use cosine similarity as a simple scale-invariant alignment score and for consistency with works \cite{bib:filus2025ecml,bib:nayak2019zero,bib:filus2024similarity}, where cosine-based similarity proved functionally relevant for network similarity comparisons. We scale the NCSM values to range $\langle 0, 1 \rangle$. In contrast to  stimuli-based similarity estimation approaches, we do not need any test images to compute similarity, making the method significantly faster than the approach based on data samples \cite{bib:huang2021semantic} or confusion matrices \cite{bib:bilal2017convolutional}, while recent works have demonstrated that this similarity estimation is sufficient for multiple practical applications \cite{bib:filus2024similarity,bib:filus2025ecml,bib:filus2025doggest,bib:mopuri2020adversarial,bib:nayak2019zero}. 
    
     \textbf{Confusion-based Class Similarity Matrices (CCSMs)} represents an \textit{indirect} or \textit{functional} approach to similarity estimation from the perspective of $\mathcal{M}$. Let $ \mathbf{CM} $ be the confusion matrix, where $ \mathbf{CM}_{ij} $ is the number of instances of class $ c_i $ that are classified by $\mathcal{M}$ as class $ c_j $. $ \mathbf{CM} $ is an $ N \times N $ matrix. To create a $ \mathbf{CSM} $ from $ \mathbf{CM} $, we first normalize each row of $ \mathbf{CM} $ (each row sums to 1), and then fill the diagonal with value $1$. This results in a \textbf{CCSM}. \textbf{CCSM} is the only $ \mathbf{CSM} $ used in our experiments that can be asymmetric, meaning that $ \mathbf{CSM}_{ij} \neq \mathbf{CSM}_{ji} $ in general. The reason is that the confusion between class $ c_i $ being classified as class $ c_j $ may differ from class $ c_j $ being mistaken as class $ c_i $.
    
     \textbf{Semantic Class Similarity Matrices (SCSMs)} is a \textit{similarity reference} used by our framework. 
    Semantic relations approximate collective human similarity judgments, as they capture structured connections that humans intuitively recognize (common functionality, appearance, evolutionary traits). Semantic similarity is a relation between terms with a similar meaning \cite{bib:kolb2009experiments}. Semantic similarity can be measured \emph{e.g.} via WordNet \cite{bib:miller1998wordnet} similarity measures. We use path similarity \cite{bib:pedersen2004wordnet} in our study, as according to \cite{bib:kolb2009experiments}, it outperforms other measures in terms of correlation with human judgment of semantic relatedness. We use WordNet semantics instead of human judgments because its large lexical database is good for generalization and increases the chance of finding relevant representations. WordNet similarity is also clearly defined and provides consistent, objective scores derived from a fixed taxonomy. By computing the similarities between all categories in $\mathcal{D}$ (WordNet nodes), we obtain the \textbf{WordNet CSM (WNCSM)}, which is a special case of a SCSM.


\begin{figure}[h]
\centering
\includegraphics[width=0.9\textwidth]{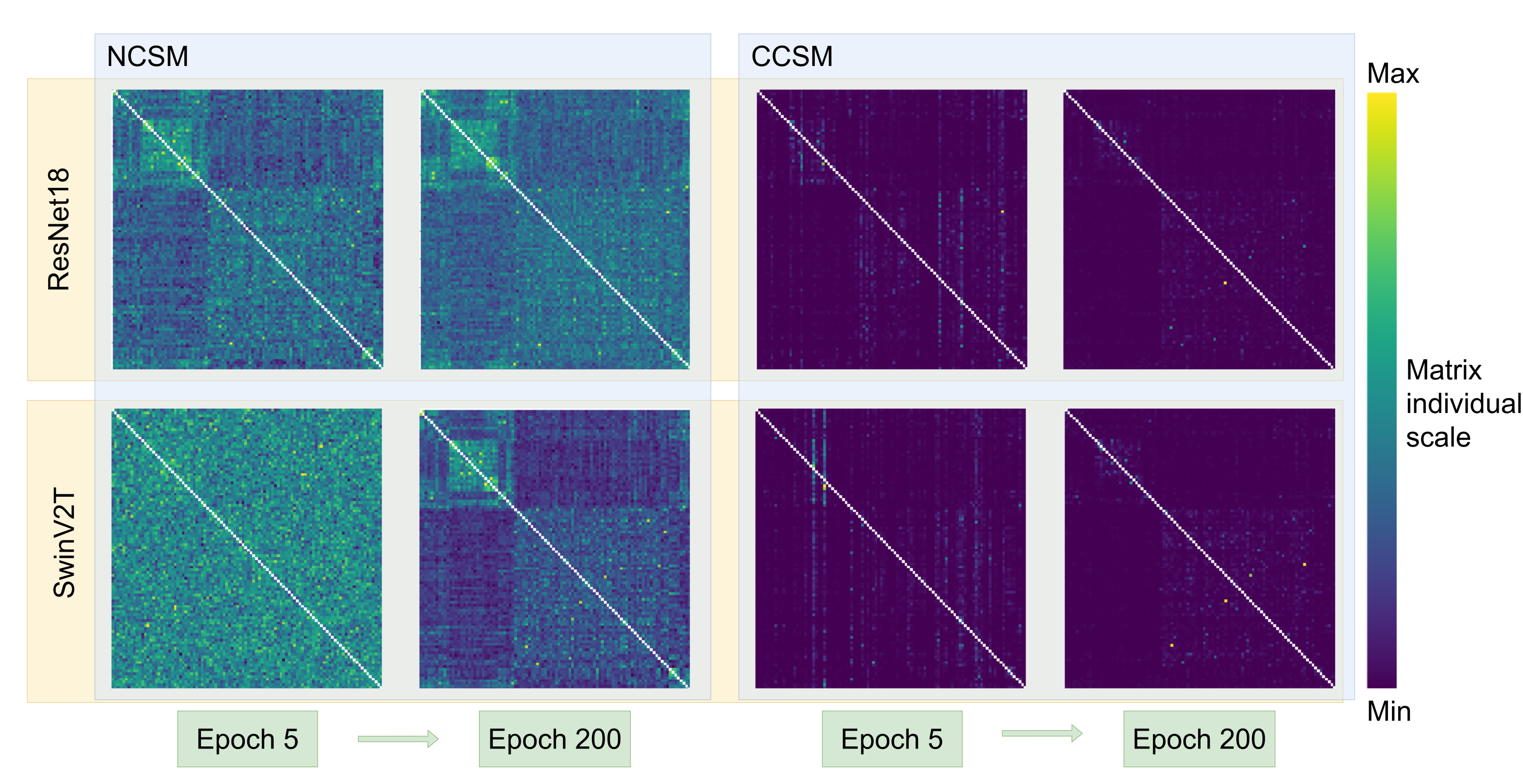} 
\caption{Mini-ImageNet: NCSMs and CCSMs of ResNet18 and SwinV2. }
\label{fig:ncsms_resnet_swin_mini_ccsm}
\end{figure}

Using different CSMs, we construct a set of key methods of our framework that can be used to examine networks. While our focus is on numerical analysis, which is non-existent in the literature, similarly to the available stimuli-based qualitative works \cite{bib:bilal2017convolutional,bib:huang2021semantic}, we also visualize class similarity matrices to graphically inspect and support our metrics. While SCSMs are constant in time, both the NCSMs and CCSMs differ between training steps. Their visualization shows how the perception changes with time and how it is developed from the beginning of the training to its end. In the main paper, we present some chosen matrices (measured after epochs 5 and 200), however the matrices for all epochs can be used to create a full animation of the similarity development during training.  

\textbf{Similarity Alignment Index (SAI) Curves} measure to what degree 2 Class Similarity Matrices are similar, thus how two different similarity perceptions are aligned. Let \( \mathbf{CSM}_1 \) and \( \mathbf{CSM}_2 \) be 2 matrices of size \( N \times N \). The first step is to exclude the diagonal elements from both matrices: $\mathbf{CSM}_1' = \mathbf{CSM}_1 - \text{diag}(\mathbf{CSM}_1), \mathbf{CSM}_2' = \mathbf{CSM}_2 - \text{diag}(\mathbf{CSM}_2)$. The second step is to Normalize the remaining elements of both matrices to the range \(\langle0, 1\rangle\). After applying these modifications, the two matrices can be compared with a chosen similarity measure to obtain $\mathbf{SAI}(\mathbf{CSM}_1, \mathbf{CSM}_2)$. We chose Cosine Similarity due to its frequent usage for high-dimensional data. We name the plots of SAI as a function of a number of epoch as \textbf{Similarity Alignment Index (SAI) Curves}. They allow to observe how the similarity perception of a given network changes during training. Possible variants below:
        \begin{itemize}
            \item \textbf{SAI(NCSM, SCSM)} - measures the alignment between the direct similarity perception of a network with the semantic similarity, indicating if its understanding of relationships aligns with semantic reasoning. 
            Low \textbf{SAI(NCSM, SCSM)} values suggest that a network's perception does not align with semantics. Its high values imply they are well aligned. Changes in SAI indicate whether a network's perception converge with semantic similarity during learning or diverge from it, suggesting the use of more machine-like reasoning for categorization;
            \item \textbf{SAI(NCSM, CCSM)} - evaluates how well network’s similarity perception aligns with its mistakes to determine whether the errors stem from perceived similarity (more predictable errors), thus model-centric error severity;
            \item \textbf{SAI(CCSM, SCSM)} - measures how closely the mistake patterns align with semantic similarity, indicating whether its errors make sense semantically (more reasonable errors), thus semantic error severity. 
        \end{itemize}

 \textbf{Inverse Dissimilarity Metric (IDM) Curves} are based on the Dissimilarity Metric introduced in work \cite{bib:filus2023netsat} for the  adversarial attacks evaluation. It can be interpreted as the mean similarity shift between the ground truth label and the post attack label. In this work, we acknowledge its usefulness for similarity development and error examination and generalize it to a variant that computes the mean similarity shift between the ground truth label and the predicted label on clear data. We propose to use an inverse version of this metric as an extension of accuracy assessment that measures the severity of network errors. The metric can be computed as follows. To generate the standard DM values, a given $\mathbf{CSM}$ is used. Each row is sorted in a descending order and a matrix with ids of the classes belonging to the particular similarity values is obtained -- Sorted Class Similarity Matrix (SoCSM). Each $c$-th row stores classes with the decreasing similarity values for the $c$-th class. For each sample, we take the ground truth label $i$ and the prediction $j$. We check in the SoCSM at which index label $j$ is placed in the $i$-th row. The larger the index, the more dissimilar the label is to the ground truth. As our target is to use DM for accuracy inspection, we transform it to obtain the Inverse variant: $IDM = 1 - DM$. We consider only the cases in which a given network returned an incorrect prediction (\textbf{IDM's errors only variant}). This can be approximated as $\frac{DM}{1 - accuracy}$. Plotting the IDM value as a function of the epoch number shows how the accuracy and the mistakes being made change in relation to similarity. Depending on which $\mathbf{CSM}$ is used to obtain the $\mathbf{SoCSM}$, we define:
\begin{itemize}
    \item \textbf{Network-based IDM (NIDM)} - The inverse version of the metric (DM metric) proposed in \cite{bib:filus2023netsat}. Network CSM reflects inter-class similarities and NIDM shows how accurate is the network in terms of the perceived similarity -- it can be treated as indicator of mistakes' predictability based on model-centric error severity. The errors-only variant measures whether the network's similarity perception and its mistakes are related, and to what extent this perception impacts the mistakes. This is a more local approach than SAI(NCSM,CCSM). Increasing NIDM suggests that the network starts to make mistakes between categories it perceives as increasingly similar.;
    \item \textbf{WordNet-based IDM (WIDM)} - it is our original modification of DM. Instead of NCSM, it uses WCSM, therefore it can be treated as a semantic version of IDM. It says how accurate is the network (even if it makes mistakes) in terms of semantic similarity. The errors-only variant focuses on the semantic similarity of the mistakes (semantic severity). Increasing WIDM suggests that the network starts to make mistakes between categories that are increasingly semantically similar. 
\end{itemize}

 \textbf{Weights Similarity Index (WSI) Curves} describe how the relationships between weights during training change. We define Weights Similarity Index (WSI) as a mean of some specific elements of an NCSM. Below, we formulate and interpret different WSIs variants: 

\begin{itemize}
    \item \textbf{Mean WSI} is computed as $WSI_{\mu} = \frac{2}{N(N-1)} \sum_{i=1}^{N-1} \sum_{j=i+1}^{N} \mathbf{NCSM}_{ij}$. Upper triangle of the similarity matrix is considered. Its curves show how the similarity of weights changes overall. If it increases, it means that the representations of classes are pulled towards each other. In the opposite case -- they are pushed away from each other; 
    \item \textbf{Max WSI} represents the mean maximum similarity of classes. 
    For each class $i$, let $\mathcal{S}_{\geq Q_i(0.95)}(i)$ represent the set of similarities that are larger or equal than the quantile $0.95$, excluding self-similarity ($S_{ii}$). Max WSI is the mean value of the averaged similarities for each class $i$: $WSI_{max} = \frac{1}{N} \sum_{i=1}^{N} \frac{1}{|\mathcal{S}_{\geq Q_i(0.95)}(i)|} \sum_{j \in \mathcal{S}_{\geq Q_i(0.95)}(i)} S_{ij}$. Its curves represent how the similarity of classes perceived as the most similar by a given network changes, showing the changes in the local similarity of classes. Increases suggest discovering highly similar classes (representations pulled towards each other);
    \item\textbf{Min WSI} - represents  mean similarity of the most dissimilar classes for each class. 
    For class $i$, let $\mathcal{S}_{\leq Q_i(0.05)}(i)$ represent the set of similarities that are less or equal than quantile $0.05$. Min WSI is the mean value of averaged similarities per class $i$: $WSI_{max} = \frac{1}{N} \sum_{i=1}^{N} \frac{1}{|\mathcal{S}_{\leq Q_i(0.05)}(i)|}  \sum_{j \in \mathcal{S}_{\leq Q_i(0.05)}(i)} S_{ij}$. Min WSI represents how networks learn most distinct inter-class dissimilarities (representations pushed away).
\end{itemize}
    
We also provide supplementary materials (Appendix \ref{appendix_metrics}) that may serve as a visual and terminological aid, including visualizations of the methods and a brief dictionary of the main matrices and metrics used throughout the paper.

\section{Experiments}

In our experiments, we examine the similarity perception of 2 standard CNNs (ResNet18 \cite{bib:he2016deep}, MobileNetV2 \cite{bib:sandler2018mobilenetv2}), 1 CNN modernized with ViT-inspired techniques (ConvNeXt-T \cite{bib:liu2022convnet}), 2 ViTs (ViTB \cite{bib:dosovitskiy2020image}, SwinV2T \cite{bib:liu2022swin}) and 1 hybrid model (MaxViTT \cite{bib:tu2022maxvit}). We experiment with the Mini-ImageNet \cite{bib:vinyals2016matching} dataset. All models were trained on a largely unified setup, with architecture-specific modifications where required: we used normalized \(224\times224\) inputs, Adam optimizer, categorical cross-entropy loss, batch sizes of $96$/$128$, learning rates of order \(10^{-3}\) to \(10^{-4}\), 400 epochs. We applied 10 epochs of linear warm-up followed by a Reduce-on-Plateau scheduler. Training used a 10-epoch linear warm-up (\(\text{decay}=0.01\)), followed by a Reduce-on-Plateau scheduler with model-specific factor (\(0.6\text{--}0.9\)) and patience (\(10\text{--}30\)). We use standard architecture-appropriate data augmentation. For CNNs: Random Resized Crop/Horizontal Flip/Rotation, Gaussian Blur, Color Jitter, Random Perspective, and Random Affine transformations. For ViTs, we additionally apply CutMix. Full model-specific configurations are available in our Zenodo repository. 
We also provide supplementary materials that may help contextualize and further illustrate the main results, including additional details on the datasets and models (Appendix~\ref{appendix_materials}), the WordNet reference SCSM and standard accuracy/loss curves for all networks (Appendix~\ref{appendix_wordnet_accuracy}), more CSM and example visualizations for the mini-ImageNet results (App.~\ref{appendix_mini}, App.~\ref{appendix_mistakes}), and an analogous analysis on CIFAR100 with results consistent with our main experiments (Appendix~\ref{appendix_cifar}).
 \textbf{The aim of our experiments is to answer the following research questions:} \textbf{(RQ1)} How does the direct similarity perception change during training?; \textbf{(RQ2)} How does the similarity perception align with semantic similarity during training?; \textbf{(RQ3)} Do the confusion patterns follow the network's similarity perception during training? \textbf{(RQ4)} Do the confusion patterns follow semantic similarity during training?

\begin{figure}[h]
\begin{center}
\subfloat[][Mean]{\includegraphics[width=0.33\textwidth]{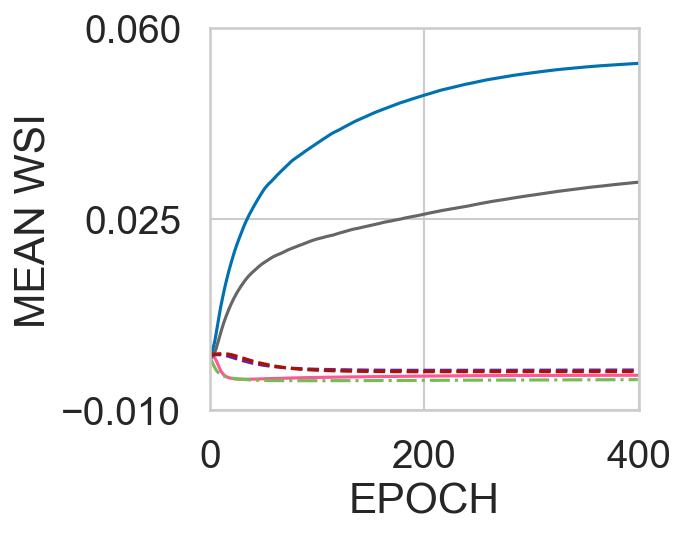}}
\subfloat[][Max]{\includegraphics[width=0.33\textwidth]{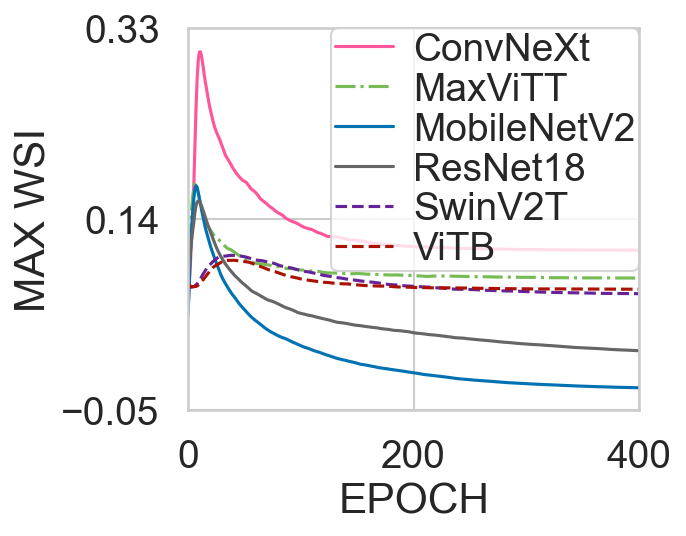}}
\subfloat[][Min]{\includegraphics[width=0.33\textwidth]{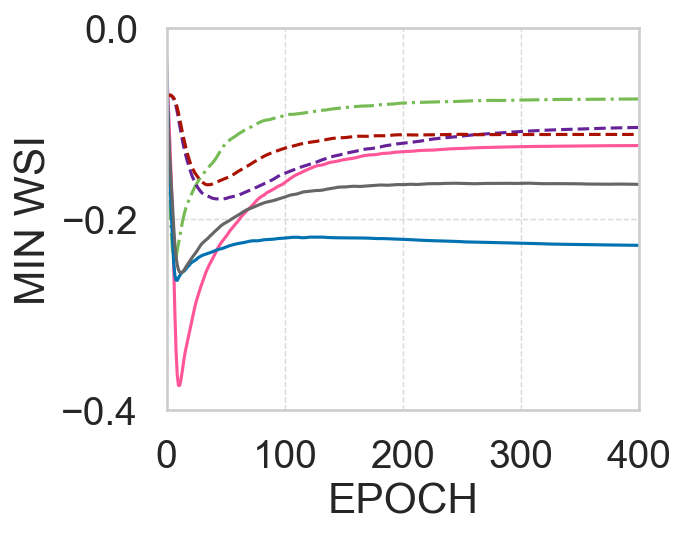}}
\end{center}
\caption{Mini-ImageNet: Weights Similarity Index (WSI). Common legend used.}
\label{fig:WSI_Mini}
\end{figure}

\subsection{How does the similarity perception change during training?}\label{sec:sai_sem}

Weights Similarity Index (WSI) Curves show how inter-class similarities change during training (Fig. \ref{fig:WSI_Mini}). For the majority of networks, mean similarity initially slightly increases to quickly decrease toward the negative cosine similarity. The minimum values are close to 0, suggesting the pursuit to achieve orthogonality. 2 standard CNNs behave significantly different from other models (ResNet18, MobileNetV2) with Mean WSIs growing in a logarithmic fashion with values close to 0 (results for 2 more CNNs in App. \ref{appendix_wsi_cnn} support that this shape of Mean WSI is characteristic for CNNs). For Max WSI, networks show an increase at the beginning of training. It reaches a peak at app. 15 epochs for CNNs, and 40 for ViTs). The curves are mirrored with respect to the x-axis for Min WSI. The results for the hybrids are closer in terms of the curvature to the ones of CNNs, and in terms of the final values - to ViTs. It shows that hybrid models combine dynamics of both architectures. This suggests the existence of a \textit{phase of network's rapid discovery of the most similar categories and the most dissimilar categories}, showing an effort to push the first ones to, and pull the latter from each other. It occurs during the early epochs, when networks obtain low accuracies. After initial gains/drops for the Max/Min variants, the similarities start to decrease/grow, suggesting a 2nd phase, in which \textit{ differences are discovered between similar classes, and similarities between dissimilar classes} (vector orthogonality pursuit). After some epochs, the perception reaches a relative stability (we call it the \textit{stability phase}). 

\subsection{Does the network similarity perception align with semantic similarity during training?}\label{sec:sai}

\begin{figure}[h]
\centering

\subfloat[][SAI(NCSM, SCSM)\label{fig:sai_network_wordnet}]{
  \includegraphics[width=0.33\textwidth]{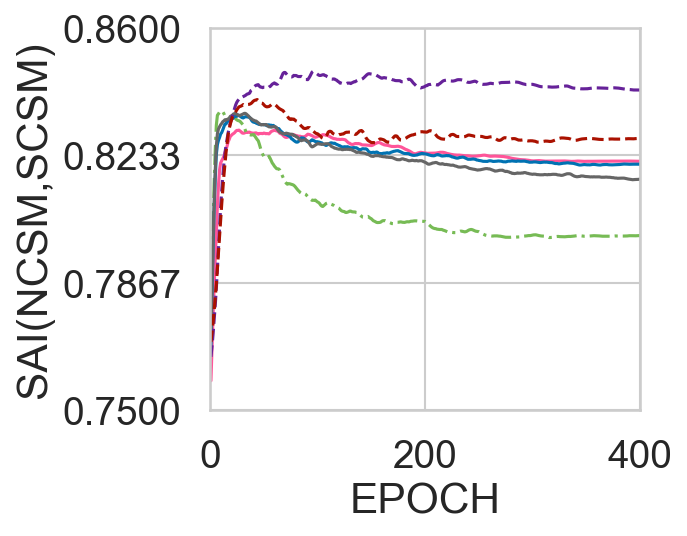}
}
\subfloat[][SAI(NCSM, CCSM)\label{fig:sai_ncsm_ccsm}]{
  \includegraphics[width=0.33\textwidth]{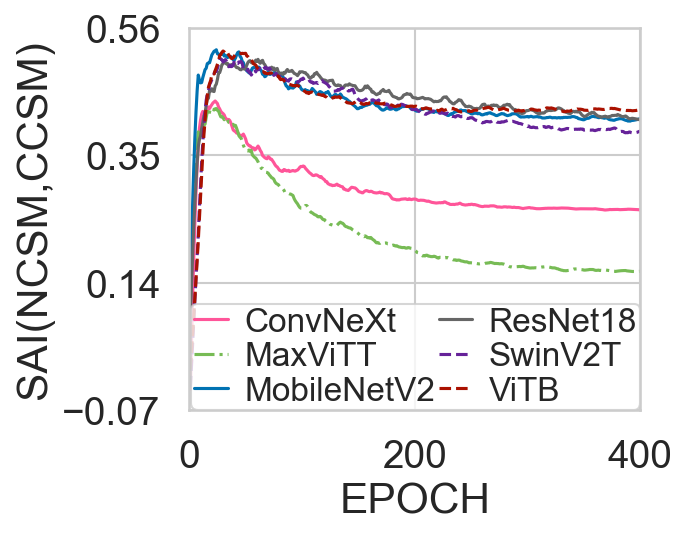}
}
\subfloat[][SAI(CCSM, SCSM)\label{fig:sai_ccsm_scsm}]{
  \includegraphics[width=0.33\textwidth]{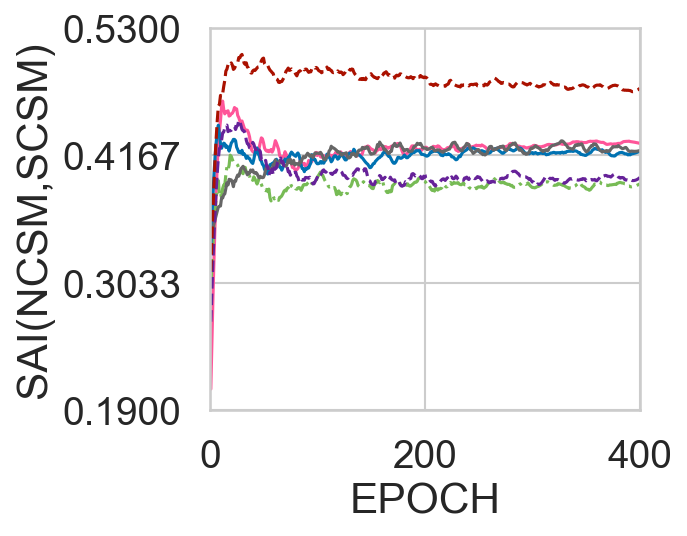}
}
\caption{Mini-ImageNet: SAI curves for different similarity pairings.}
\label{fig:sai_miniimagenet}
\end{figure}

We analyze whether the similarity perception of CNNs and ViTs aligns with semantics and its changes during training. In Fig.~\ref{fig:sai_network_wordnet} we present the Cosine- and Structure-based \textbf{SAI(NCSM, SCSM)}. For both, a rapid increase in the alignment between the network and semantic similarity can be observed for the examined networks. This increase is faster for CNNs/the hybrid than for ViTs. This matches the 1st training phase observed while analyzing the WSI plots. It suggests that the dynamic learning of inter-class similarities is due to the actual semantic similarities and highly correlated structure of the world. These are discovered and learned to understand, which is in line with the categorization principles from cognitive psychology presented in the introduction. Again, after this initial growth, the alignment slightly decreases (suggesting the similarity perception `refinement') with visible `bumps' in the curve to practically stabilize in the later epochs (slightly earlier for CNNs/the hybrid than for ViTs). These non-monotonic fluctuations occur while the models are still being actively optimized and are consistent with transitions in the learning dynamics rather than meaning a reversal in the overall trend. Supplementary material's Appendix~\ref{appendix_bumps} provides a more detailed illustration of this effect on an example CNN by jointly plotting SAI(NCSM,SCSM), training loss, and learning rate. After reaching its plateau, the alignment persists to be higher for ViTs than for CNNs/hybrids. We support these numerical results with visualizations of NCSMs for the 5th and 200th training epochs for 2 models in Fig. \ref{fig:ncsms_resnet_swin_mini_ccsm}.
Already after 5 epochs, the CNN exhibits a clear emerging hierarchical similarity structure, visible as a block-diagonal pattern whose two largest upper-left clusters correspond to the basic-level categories of animals and artifacts, whereas in SwinV2 this organization remains much weaker. In epoch 200, almost the same structure is visible for both models. For reference, the corresponding WordNet matrix is shown in Fig.~\ref{fig:wordnet_csm} of App. \ref{appendix_wordnet_accuracy}, while additional NCSMs for all other models and different epochs (1, 5, 25, 200) are provided in Appendix~\ref{appendix_mini} in our Supplementary Materials for further visualization.




Notably, the strongest semantic-alignment gains occur within roughly the first 25 epochs, whereas the largest improvements in standard optimization metrics (accuracy, loss) extend longer, over approximately the first 50-100 epochs. This suggests that early performance gains are accompanied by a rapid discovery of inter-class similarity structure, while the subsequent phase is dominated more by refinement and differentiation among nearby classes than by further increases in semantic alignment. The later slight decrease in SAI may reflect discovering relations not well captured by WordNet, such as containment or meronymy \cite{bib:bilal2017convolutional,bib:filus2025ecml}, or a task-driven separation of highly similar classes (visible in Fig. \ref{fig:WSI_Mini}) to reduce confusions. Moreover, models reaching similar accuracy can still differ in semantic-alignment level, indicating that similarity-based measures may provide a complementary criterion for model selection beyond accuracy if such alignment is desired. The discussed accuracy and loss curves are provided in Appendix~\ref{appendix_wordnet_accuracy} in Supplementary Materials. The referenced accuracy plot indicates that models ConvNeXt, MobileNetV2, SwinV2T, ResNet18 obtained almost the same testing accuracy at epoch 400, while MaxViT and ViTB obtained slightly higher and lower accuracy respectively. 

\subsection{Does the confusion match network similarity perception during training? }\label{sec:idm_confusion}

To examine how errors align with network's perception, we use the \textbf{SAI(NCSM, CCSM)} curves in Fig. \ref{fig:sai_ncsm_ccsm} showing that both similarity perception estimates (direct - NCSM, indirect - CCSMs) align closely, peaking around epoch 25. Sparser structure of confusion-based CSMs than NCSMs' results in lower SAI values (visible in Fig. \ref{fig:ncsms_resnet_swin_mini}). The ViT-CNN hybrid and ViT-inspired CNN behave similarly. The curves of CNNs are initially steeper than for ViTs. Then, alignment decrease occurs and stabilization, stemming from the smaller mistakes number. 

We also provide a visual aid -- Confusion-based CSMs  -- in Fig. \ref{fig:ncsms_resnet_swin_mini}). At later epochs, CCSMs start to reflect a similar box-diagonal structure as NCSMs (Fig. \ref{fig:ncsms_resnet_swin_mini}) with lower density. However, it needs more epochs to be clearly developed, even for ResNet18, whose NCSM already exhibited  a clearly box-diagonal structure at the 5th epoch. At early epochs, confusion (thus the indirect similarity estimation) can be observed even out of the basic-level categories (see the off-diagonal `noise' and visible vertical `stripes' in the CCSMs). It suggests that while the networks quickly discover inter-class similarities, they need more epochs to align their mistakes with their similarity judgments, \emph{e.g.} via improving on more `atypical'/`difficult' samples. 



\begin{figure}[h]
\begin{center}
\subfloat[][Network-based\label{fig:idm_network:a}]{
  \includegraphics[width=0.45\textwidth]{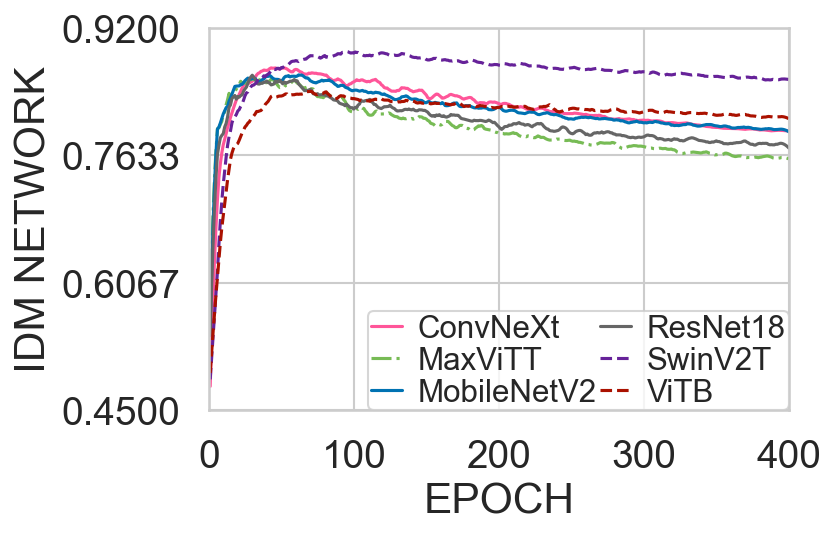}
}
\subfloat[][WordNet-based\label{fig:idm_wordnet:a}]{
  \includegraphics[width=0.45\textwidth]{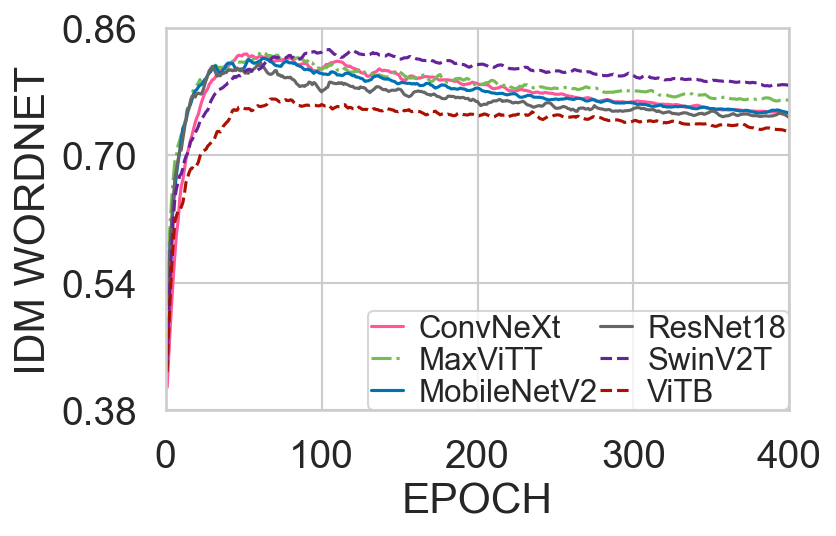}
}
\end{center}
\caption{Mini-ImageNet: Inverse Dissimilarity Metric (IDM) .}
\label{fig:idm_network}
\label{fig:idm_wordnet}
\end{figure}

\begin{figure}[h]
    \centering
    \includegraphics[width=0.95\linewidth]{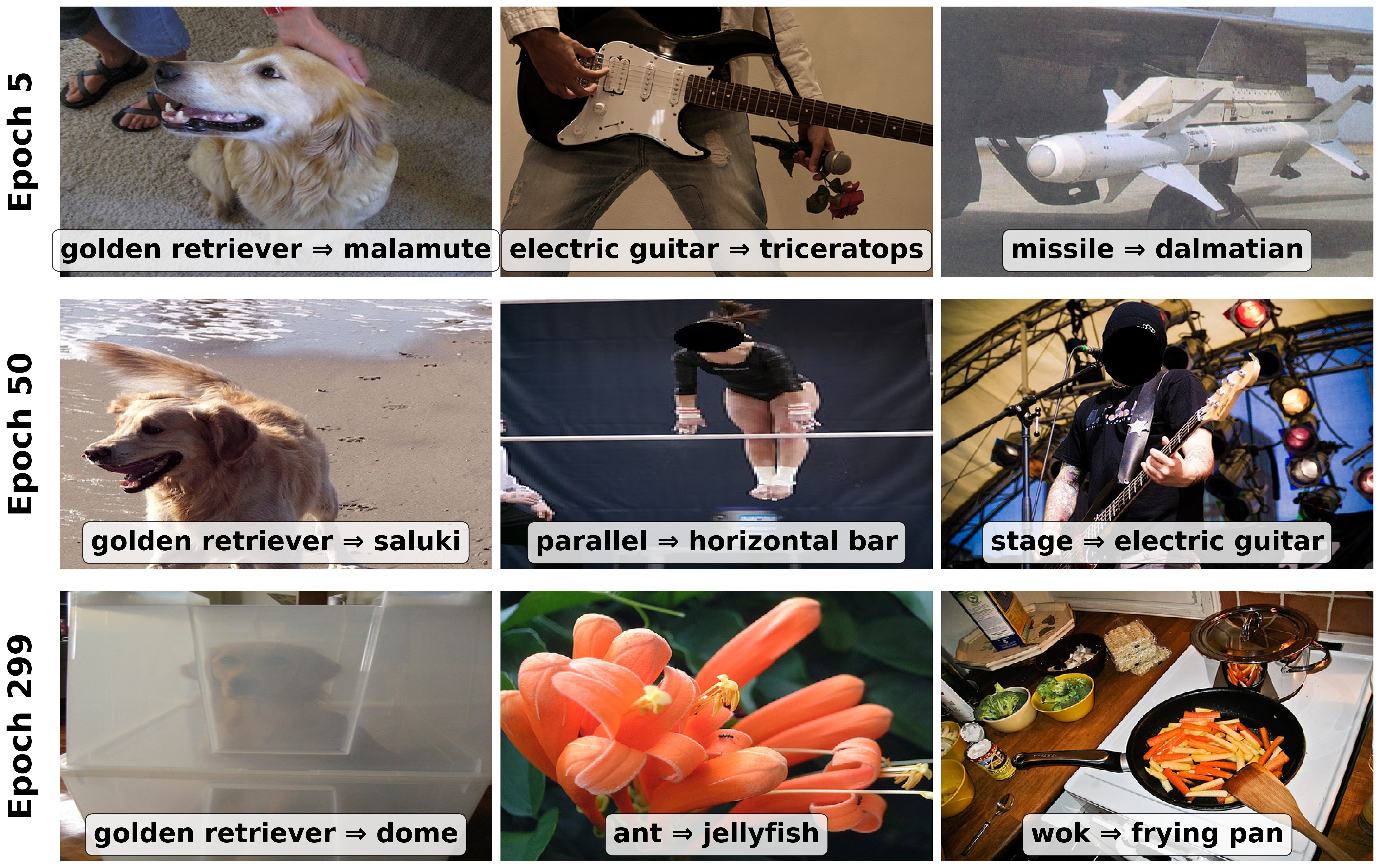}
    \caption{Inspection of MobileNetV2's mistakes on Mini-ImageNet at epoch \textbf{5} (top row),  \textbf{50} (middle row)  and \textbf{299} (bottom row). At epoch 5 mistakes with very dissimilar classes dominate even for typical images with (already) some mistakes with similar classes, e.g. two breeds of dogs. At epoch 50, mistakes tend to be more refined, caused by different semantic relations (e.g. similarity, containment). At epoch 299, network rather makes mistakes on atypical, difficult images (a dog hidden in a box or very small target objects (an ant) overwhelmed by different objects. The impact of minor mislabeling is also visible (wok, pan). }
    \label{fig:examples_mistakes}
\end{figure}

Fig. \ref{fig:idm_network} shows the Network-based IDM Curves. One can use them to analyze the network's mistakes more locally (how distant are the predictions and ground truth in the perceived class space). The basic IDM plot confirms that networks quickly start to predict classes from the closest perceived class neighborhood (decisions guided by similarity perception). The plots reach their peak and stabilize after around 100 epochs for all networks. Surprisingly, the errors-only IDM reveals that after reaching its peak, values slightly drop, indicating the network is making errors between less similar classes. Despite the drop, networks still mostly confuse classes perceived as similar. It may be because the accuracy is already high, having eliminated more 'obvious' mistakes, and tackling challenging and less typical samples, or even mislabeling noise, and thus decreasing the IDM values. In Fig. \ref{fig:examples_mistakes}, we present example images for which MobileNetV2 made mistakes at different similarity development phases (epochs 5, 50, 299). Early in training, the errors are often semantically distant, but the first signs of similarity refinement are already visible in occasional confusions within related categories. As training progresses toward the IDM peak, mistakes concentrate among highly similar classes and increasingly reflect more atypical, low-quality inputs, co-occurring objects, or label noise, and thus supporting the fact that the learned representations become more semantically structured over time. We describe these examples more thoroughly in App. \ref{appendix_mistakes}. Due to the alignment between network similarity perception with semantics discussed in Section \ref{sec:sai}, these examples also support the results on the error-semantics alignment presented in the next section. Again, the results show that models at similar accuracy levels (ConvNeXt, MobileNetV2, SwinV2T, ResNet18) can obtain different values of confusion-similarity alignment, suggesting that these metrics can be used to define additional criteria for model selection along with accuracy comparisons.

\subsection{Do the confusion patterns match semantic similarity during training?}

To answer this question, we use the Similarity Alignment Index Curve for the Confusion-based similarity and the WordNet-based similarity - \textbf{SAI\\(CCSM, SCSM)} as a more global measure, and WordNet-based IDM as a more local measure. In Fig. \ref{fig:sai_ccsm_scsm}, we present the SAI(CCSM, SCSM) curves. The figure shows that not only do the confusion patterns follow the network's similarity perception, but also the semantic similarity. The rapid growth in the alignment also includes the first epochs of training, peaking around the 25th epoch. Then, it either decreases and stabilizes or immediately reaches stabilization. Furthermore, the initial rapid growth of both the Network-based and WordNet-based SAI variants is yet another, now indirect, proof that the network's similarity perception partially aligns with the semantic similarity in the very beginning of the training. The results of the WordNet-based IDM (see Fig.~\ref{fig:idm_wordnet}) also align with these results. For the general variant, a rapid increase in the early epochs can be noticed, followed by a quick stabilization at a high level (app. 0.9 or more). For the errors-only variant, it is visible that the mistakes initially tend to become semantically related to the ground truth, peaking before slightly declining (to a relatively high level of app. 0.75). The results show that although the accuracy is low early in training, the networks are already very semantically close to the ground truth prediction. It results in a relatively good quality of a classifier (semantically 'reasonable' decisions).

\section{Discussion}

The analysis performed in this study allowed us to observe some interesting phenomena regarding the learning of vision networks. The results show that CNNs and ViTs develop a rich and hierarchical similarity perception during the course of training for standard object recognition on natural datasets. 
Similarity perception emerges very early in the training. 
This process is significantly more dynamic for CNNs/hybrid models than for ViTs. CNNs develop similarity perception rapidly and then refine it more dynamically after reaching their peak. ViTs take longer to mature their similarity perception, but once achieved, it is maintained more consistently.
Using 
hybrid models results in the blending of both architecture features.
Nevertheless, the overall dynamics of the changes in the similarity perception are close for both architectures and 3 phases of similarity development during training can be drawn: \textbf{(1)} \textit{Initial Similarity Surge} - models rapidly discover similarities 
and dissimilarities between 
categories.;
\textbf{(2)} \textit{Similarity Refinement Phase} - models discover dissimilarities between similar classes and push them from each other, while finding 
relations between less similar categories.;
\textbf{(3)} \textit{Similarity Stabilization Phase} - the similarity perception becomes steady and networks focus on further elimination of mistakes 
Our results also show that both CNNs and ViTs
make improvements in the quality of their mistakes, which we call the \textit{mistakes refinement phenomenon}. It is particularly evident in the initial similarity development phase, in which the mistakes tend to increase similarity to the ground truth.

Moreover, as our results showed, models that obtained similar levels of accuracy can exhibit different levels of similarity alignment between the model perception and its mistakes, but also with semantics. This suggests that our similarity-based metrics can be used for model selection along with standard accuracy. At comparable accuracy levels, models with higher values of our metrics can be used. Models with higher SAI(NCSM, SCSM) store complex information structures in a way that can be better explained by semantics and high-level visual category overlaps. Although semantic and visual similarity are not identical, for many categories they overlap substantially (due to common functionality, environment, genetic origin), making such alignment a plausible indicator of semantically reasonable internal representations. Moreover, errors of models with high values of SAI(CCSM, WordNet SCSM) and WordNet-based IDM can be better explained with semantics. may be preferable to those with lower alignment, as they suggest more structured and potentially more predictable error patterns, partly resembling cognition consistency observed in humans. Our methods show potential for practical usage, as they offer a unified framework, compatible with standard procedures, fast computation (e.g. the additional complexity of computing our new metrics accounted for approximately 0.29\% of epoch time computation for a small MobileNetV2, and 0.11\% for a larger model -- SwinV2T).


Our main paper presented the core framework formulation and evidence on supervised training on Mini-ImageNet. For interested readers, the supplementary material provides additional analyses that broaden the perspective on the reported results and further support the main findings. As we focused on supervised training on mini-ImageNet, which  could limit generalizability, we provided additional results on CIFAR100 and partial results on CIFAR10, COCO, ImageNet and other supervised tasks (\emph{e.g.} object detection, segmentation) in Appendices \ref{appendix_cifar} and \ref{appendix_otherdatasets}. These results corroborated our main findings. We described how to extend our framework to self-supervised (DINOv2 \cite{bib:oquab2023dinov2}) and text-image (CLIP \cite{bib:radford2021learning}) tasks in App. \ref{appendix_contrastive}. While we use cosine similarity as a simple scale-invariant measure, due to its functional relevance in works \cite{bib:filus2025ecml,bib:nayak2019zero, bib:filus2024similarity}, other measures are also possible. Our additional analysis with standard distances such as Mean Squared/Absolute Errors on SAI(NCSM,SCSM) for Mini-ImageNet presented in Appendix \ref{appendix_wordnet_accuracy} showed similar overall trends, though with less pronounced differences between networks, likely due to the limitations of distance-based measures in high-dimensional settings. As we focused on relatively accurate networks, and ignored networks struggling to learn, to provide additional insights, we additionally compared a ``good'' and a ``bad'' ViT in Appendix \ref{appendix_bad}. The results showed that though differences in their perception occur, similarity is still an important driver even for inaccurate models. We also provide visualizations and more detailed equations for our framework and term dictionary in Appendix \ref{appendix_materials}.


\section{Conclusions}

We introduced Deep Similarity Inspector (DSI), a systematic framework for training-time analysis of similarity development in supervised vision networks. By unifying direct and indirect, as well as structural and functional, views of similarity, DSI enables the numerical analysis of how class-level similarity develops during learning and how it relates to semantics and model errors. Our experiments show that both CNNs and Transformer-based models develop non-trivial similarity structure early in training, but differ in dynamics: CNNs develop this structure more rapidly, whereas ViTs -- more gradually and then maintain it more consistently. Across models, we observed three recurring phases of similarity development, and found that error patterns become more aligned with both the network’s own similarity perception and semantics. These results suggest that similarity perception evolves over time and reflects structured regularities in visual data. Our framework and findings improve the understanding of model training dynamics and error behavior that are not reflected by accuracy alone. By analyzing whether mistakes follow semantic or model-perceived similarity, DSI provides a step toward severity-aware error analysis, where nature and interpretability of errors are considered. Therefore, DSI may serve as a lightweight tool for trustworthy model inspection and, when semantic or perception-error alignment is desired, for risk-aware model selection in human-critical settings.

\FloatBarrier

\begin{credits}

\subsubsection{Preprint notice}
This is a preprint version. A shorter version of this paper has been accepted for presentation and publication in the post-workshop proceedings of the Joint Workshop on Security, Human Awareness, and Risk Mitigation in AI-driven Systems (SHIELD-AI 2026), co-located with ECML~PKDD~2026. The appendix is included only in this preprint and is not part of the peer-reviewed proceedings paper.

\subsubsection{\ackname} We gratefully acknowledge Poland's high-performance Infrastructure \-PLGrid ACC Cyfronet AGH for providing computer facilities and support within computational grant no PLG/2024/017513. This research was supported by the~START Scholarship of the Foundation for Polish Science (FNP) for outstanding young scholars, agreement No START 017.2025 and the~Polish Minister of Science and Higher Education scholarship for outstanding young researchers, agreement No SMN/20/1546/2024. The authors utilized ChatGPT for language polishing under human supervision. The authors remain fully responsible for the manuscript.


\subsubsection{\discintname}
The authors have no competing interests. 

\end{credits}
%
%
%
\bibliographystyle{splncs04}
\bibliography{bibiliography}

@article{bib:russakovsky2015imagenet,
  title={Imagenet large scale visual recognition challenge},
  author={Russakovsky, Olga and Deng, Jia and Su, Hao and Krause, Jonathan and Satheesh, Sanjeev and Ma, Sean and Huang, Zhiheng and Karpathy, Andrej and Khosla, Aditya and Bernstein, Michael and others},
  journal={{International Journal of Computer Vision}},
  volume={115},
  pages={211--252},
  year={2015}
}

@inproceedings{bib:he2016deep,
  title={Deep residual learning for image recognition},
  author={He, Kaiming and Zhang, Xiangyu and Ren, Shaoqing and Sun, Jian},
  booktitle={{Conference on Computer Vision and Pattern Recognition}},
  pages={770--778},
  year={2016}
}

@article{bib:dosovitskiy2020image,
  title={An image is worth 16x16 words: Transformers for image recognition at scale},
  author={Dosovitskiy, Alexey},
  journal={arXiv preprint arXiv:2010.11929},
  year={2020}
}

@inproceedings{bib:liu2022swin,
  title={Swin transformer v2: Scaling up capacity and resolution},
  author={Liu, Ze and Hu, Han and Lin, Yutong and Yao, Zhuliang and Xie, Zhenda and Wei, Yixuan and Ning, Jia and Cao, Yue and Zhang, Zheng and Dong, Li and others},
  booktitle={{Conference on Computer Vision and Pattern Recognition}},
  pages={12009--12019},
  year={2022}
}

@inproceedings{bib:yun2019cutmix,
  title={Cutmix: Regularization strategy to train strong classifiers with localizable features},
  author={Yun, Sangdoo and Han, Dongyoon and Oh, Seong Joon and Chun, Sanghyuk and Choe, Junsuk and Yoo, Youngjoon},
  booktitle={{International Conference on Computer Vision}},
  pages={6023--6032},
  year={2019}
}

@article{bib:rosch1978cognition,
  title={Cognition and categorization.},
  author={Rosch, Eleanor Ed and Lloyd, Barbara B},
  year={1978},
  publisher={Lawrence Erlbaum}
}

@article{bib:medin1993respects,
  title={Respects for similarity.},
  author={Medin, Douglas L and Goldstone, Robert L and Gentner, Dedre},
  journal={{Psychological Review}},
  volume={100},
  number={2},
  pages={254},
  year={1993},
  publisher={American Psychological Association}
}

@article{bib:kriegeskorte2008rdm,
  title={Representational similarity analysis - connecting the branches of systems neuroscience},
  author={Kriegeskorte, Nikolaus and Mur, Marieke and Bandettini, Peter},
  journal={Frontiers in Systems Neuroscience},
  volume={2},
  number={4},
  year={2008},
  doi={10.3389/neuro.06.004.2008}
}

@inproceedings{bib:filus2025ecml,
	title = {How CNNs and ViTs perceive similarities between categories},
	year = {2025},
	author = {Katarzyna Filus and Joanna Doma{\'n}ska},
	booktitle={European Conference on Machine Learning and Principles and Practice of Knowledge Discovery in Databases}
}

@inproceedings{bib:kornblith2019similarity,
  title={Similarity of neural network representations revisited},
  author={Kornblith, Simon and Norouzi, Mohammad and Lee, Honglak and Hinton, Geoffrey},
  booktitle={International conference on machine learning},
  pages={3519--3529},
  year={2019},
  organization={PMLR}
}

@inproceedings{bib:williamsequivalence,
  title={Equivalence between representational similarity analysis, centered kernel alignment, and canonical correlations analysis},
  author={Williams, Alex H},
  booktitle={UniReps: 2nd Edition of the Workshop on Unifying Representations in Neural Models}
}

@article{bib:oquab2023dinov2,
  title={Dinov2: Learning robust visual features without supervision},
  author={Oquab, Maxime and Darcet, Timoth{\'e}e and Moutakanni, Th{\'e}o and Vo, Huy and Szafraniec, Marc and Khalidov, Vasil and Fernandez, Pierre and Haziza, Daniel and Massa, Francisco and El-Nouby, Alaaeldin and others},
  journal={arXiv preprint arXiv:2304.07193},
  year={2023}
}

@article{bib:geirhos2021partial,
  title={Partial success in closing the gap between human and machine vision},
  author={Geirhos, Robert and Narayanappa, Kantharaju and Mitzkus, Benjamin and Thieringer, Tizian and Bethge, Matthias and Wichmann, Felix A and Brendel, Wieland},
  journal={Advances in Neural Information Processing Systems},
  volume={34},
  pages={23885--23899},
  year={2021}
}

@article{bib:geirhos2020beyond,
  title={Beyond accuracy: quantifying trial-by-trial behaviour of CNNs and humans by measuring error consistency},
  author={Geirhos, Robert and Meding, Kristof and Wichmann, Felix A},
  journal={Advances in Neural Information Processing Systems},
  volume={33},
  pages={13890--13902},
  year={2020}
}

@inproceedings{bib:radford2021learning,
  title={Learning transferable visual models from natural language supervision},
  author={Radford, Alec and Kim, Jong Wook and Hallacy, Chris and Ramesh, Aditya and Goh, Gabriel and Agarwal, Sandhini and Sastry, Girish and Askell, Amanda and Mishkin, Pamela and Clark, Jack and others},
  booktitle={International conference on machine learning},
  pages={8748--8763},
  year={2021},
  organization={PMLR}
}

@inproceedings{bib:chen2020simple,
  title={A simple framework for contrastive learning of visual representations},
  author={Chen, Ting and Kornblith, Simon and Norouzi, Mohammad and Hinton, Geoffrey},
  booktitle={International conference on machine learning},
  pages={1597--1607},
  year={2020},
  organization={PMLR}
}

@article{bib:margalit2024unifying,
  title={A unifying framework for functional organization in early and higher ventral visual cortex},
  author={Margalit, Eshed and Lee, Hyodong and Finzi, Dawn and DiCarlo, James J and Grill-Spector, Kalanit and Yamins, Daniel LK},
  journal={Neuron},
  year={2024},
  publisher={Elsevier}
}

@article{bib:vinyals2016matching,
  title={Matching networks for one shot learning},
  author={Vinyals, Oriol and Blundell, Charles and Lillicrap, Timothy and Wierstra, Daan and others},
  journal={{Advances in Neural Information Processing Systems}},
  volume={29},
  year={2016}
}

@article{bib:filus2025doggest,
  title={What is the doggest dog? Examination of typicality perception in ImageNet-trained networks},
  author={Filus, Katarzyna and Doma{\'n}ska, Joanna},
  journal={Neural Networks},
  volume={188},
  pages={107425},
  year={2025},
  publisher={Elsevier}
}

@inproceedings{bib:lin2014microsoft,
  title={Microsoft coco: Common objects in context},
  author={Lin, Tsung-Yi and Maire, Michael and Belongie, Serge and Hays, James and Perona, Pietro and Ramanan, Deva and Doll{\'a}r, Piotr and Zitnick, C Lawrence},
  booktitle={Computer Vision--ECCV 2014: 13th European Conference, Zurich, Switzerland, September 6-12, 2014, Proceedings, Part V 13},
  pages={740--755},
  year={2014},
  organization={Springer}
}

@article{bib:krizhevsky2009learning,
  title={Learning multiple layers of features from tiny images},
  author={Krizhevsky, Alex and Hinton, Geoffrey and others},
  year={2009},
  publisher={Toronto, ON, Canada}
}

@inproceedings{bib:sandler2018mobilenetv2,
  title={Mobilenetv2: Inverted residuals and linear bottlenecks},
  author={Sandler, Mark and Howard, Andrew and Zhu, Menglong and Zhmoginov, Andrey and Chen, Liang-Chieh},
  booktitle={{Conference on Computer Vision and Pattern Recognition}},
  pages={4510--4520},
  year={2018}
}

@inproceedings{bib:tu2022maxvit,
  title={Maxvit: Multi-axis vision transformer},
  author={Tu, Zhengzhong and Talebi, Hossein and Zhang, Han and Yang, Feng and Milanfar, Peyman and Bovik, Alan and Li, Yinxiao},
  booktitle={European conference on computer vision},
  pages={459--479},
  year={2022},
  organization={Springer}
}

@inproceedings{bib:he2020momentum,
  title={Momentum contrast for unsupervised visual representation learning},
  author={He, Kaiming and Fan, Haoqi and Wu, Yuxin and Xie, Saining and Girshick, Ross},
  booktitle={{Conference on Computer Vision and Pattern Recognition}},
  pages={9729--9738},
  year={2020}
}

@inproceedings{bib:caron2021emerging,
  title={Emerging properties in self-supervised vision transformers},
  author={Caron, Mathilde and Touvron, Hugo and Misra, Ishan and J{\'e}gou, Herv{\'e} and Mairal, Julien and Bojanowski, Piotr and Joulin, Armand},
  booktitle={{International Conference on Computer Vision}},
  pages={9650--9660},
  year={2021}
}

@inproceedings{bib:bertinetto2020making,
  title={Making better mistakes: Leveraging class hierarchies with deep networks},
  author={Bertinetto, Luca and Mueller, Romain and Tertikas, Konstantinos and Samangooei, Sina and Lord, Nicholas A},
  booktitle={{Conference on Computer Vision and Pattern Recognition}},
  pages={12506--12515},
  year={2020}
}

@article{bib:bilal2017convolutional,
  title={Do convolutional neural networks learn class hierarchy?},
  author={Bilal, Alsallakh and Jourabloo, Amin and Ye, Mao and Liu, Xiaoming and Ren, Liu},
  journal={{IEEE Transactions on Visualization and Computer Graphics}},
  volume={24},
  number={1},
  pages={152--162},
  year={2017}
}

@inproceedings{bib:liu2022convnet,
  title={A convnet for the 2020s},
  author={Liu, Zhuang and Mao, Hanzi and Wu, Chao-Yuan and Feichtenhofer, Christoph and Darrell, Trevor and Xie, Saining},
  booktitle={{Conference on Computer Vision and Pattern Recognition}},
  pages={11976--11986},
  year={2022}
}

@article{bib:huang2021semantic,
  title={Semantic relatedness emerges in deep convolutional neural networks designed for object recognition},
  author={Huang, Taicheng and Zhen, Zonglei and Liu, Jia},
  journal={{Frontiers in Computational Neuroscience}},
  volume={15},
  pages={625804},
  year={2021}
}

@inproceedings{bib:mopuri2020adversarial,
  title={Adversarial Fooling Beyond" Flipping the Label"},
  author={Mopuri, Konda Reddy and Shaj, Vaisakh and Babu, R Venkatesh},
  booktitle={{Conference on Computer Vision and Pattern Recognition Workshops}},
  pages={778--779},
  year={2020}
}

@inproceedings{bib:filus2023netsat,
  title={NetSat: Network Saturation Adversarial Attack},
  author={Filus, Katarzyna and Domanska, Joanna},
  booktitle={{IEEE International Conference on Big Data}},
  pages={5038--5047},
  year={2023}
}

@article{bib:tang2017visual,
  title={Visual and semantic knowledge transfer for large scale semi-supervised object detection},
  author={Tang, Yuxing and Wang, Josiah and Wang, Xiaofang and Gao, Boyang and Dellandr{\'e}a, Emmanuel and Gaizauskas, Robert and Chen, Liming},
  journal={{IEEE Transactions on Pattern Analysis and Machine Intelligence}},
  volume={40},
  number={12},
  pages={3045--3058},
  year={2017}
}

@inproceedings{bib:nayak2019zero,
  title={Zero-shot knowledge distillation in deep networks},
  author={Nayak, Gaurav Kumar and Mopuri, Konda Reddy and Shaj, Vaisakh and Radhakrishnan, Venkatesh Babu and Chakraborty, Anirban},
  booktitle={{International Conference on Machine Learning}},
  pages={4743--4751},
  year={2019}
}

@article{bib:muttenthaler2024improving,
  title={Improving neural network representations using human similarity judgments},
  author={Muttenthaler, Lukas and Linhardt, Lorenz and Dippel, Jonas and Vandermeulen, Robert A and Hermann, Katherine and Lampinen, Andrew and Kornblith, Simon},
  journal={{Advances in Neural Information Processing Systems}},
  volume={36},
  year={2023}
}

@inproceedings{bib:deselaers2011visual,
  title={Visual and semantic similarity in imagenet},
  author={Deselaers, Thomas and Ferrari, Vittorio},
  booktitle={{Conference on Computer Vision and Pattern Recognition}},
  pages={1777--1784},
  year={2011}
}

@inproceedings{bib:kolb2009experiments,
  title={Experiments on the difference between semantic similarity and relatedness},
  author={Kolb, Peter},
  booktitle={{Nordic Conference of Computational Linguistics}},
  pages={81--88},
  year={2009}
}

@book{bib:miller1998wordnet,
  title={WordNet: An electronic lexical database},
  author={Miller, George A},
  year={1998},
  publisher={MIT press}
}

@inproceedings{bib:pedersen2004wordnet,
  title={WordNet:: Similarity-Measuring the Relatedness of Concepts.},
  author={Pedersen, Ted and Patwardhan, Siddharth and Michelizzi, Jason and others},
  year={2004}
}

@inproceedings{bib:shi2019not,
  title={Not all frames are equal: Weakly-supervised video grounding with contextual similarity and visual clustering losses},
  author={Shi, Jing and Xu, Jia and Gong, Boqing and Xu, Chenliang},
  booktitle={Conference on Computer Vision and Pattern Recognition},
  pages={10444--10452},
  year={2019}
}

@article{bib:elsayed2018adversarial,
  title={Adversarial examples that fool both computer vision and time-limited humans},
  author={Elsayed, Gamaleldin and Shankar, Shreya and Cheung, Brian and Papernot, Nicolas and Kurakin, Alexey and Goodfellow, Ian and Sohl-Dickstein, Jascha},
  journal={Advances in neural information processing systems},
  volume={31},
  year={2018}
}

@inproceedings{bib:roads2021enriching,
  title={Enriching imagenet with human similarity judgments and psychological embeddings},
  author={Roads, Brett D and Love, Bradley C},
  booktitle={Conference on Computer Vision and Pattern Recognition},
  pages={3547--3557},
  year={2021}
}

@inproceedings{bib:veit2017conditional,
  title={Conditional similarity networks},
  author={Veit, Andreas and Belongie, Serge and Karaletsos, Theofanis},
  booktitle={Conference on Computer Vision and Pattern Recognition},
  pages={830--838},
  year={2017}
}

@article{bib:filus2024similarity,
  title={Similarity-driven adversarial testing of neural networks},
  author={Filus, Katarzyna and Doma{\'n}ska, Joanna},
  journal={Knowledge-Based Systems},
  volume={305},
  pages={112621},
  year={2024},
  publisher={Elsevier}
}

@book{bird2009natural,
  title={Natural language processing with Python: analyzing text with the natural language toolkit},
  author={Bird, Steven and Klein, Ewan and Loper, Edward},
  year={2009},
  publisher={" O'Reilly Media, Inc."}
}

\appendix
\section*{Appendices}
\addcontentsline{toc}{section}{Appendices} 
\setcounter{section}{0}

\renewcommand{\thefigure}{\thesection.\arabic{figure}}
\setcounter{figure}{0}

\section{Datasets and models used in the study} \label{appendix_materials}

\subsection{Datasets used in the study}

In the main part of our paper, we use \textbf{Mini-ImageNet} \cite{bib:vinyals2016matching}. It is a version of ImageNet \cite{bib:russakovsky2015imagenet} with 100 classes randomly chosen from the original dataset. ImageNet and its versions are well suited for the similarity related research, as they were created based on the semantic hierarchy of WordNet \cite{bib:miller1998wordnet}. While original ImageNet included both the internal and leaf WordNet nodes, ImageNet-1k \cite{bib:russakovsky2015imagenet}  and Mini-ImageNet \cite{bib:vinyals2016matching} consist of only leaves.  This results in ImageNet-1k and Mini-ImageNet not including any built-in hierarchy due to their labels being at one hierarchy level, and not different hierarchy levels (which occurs in the original version). Therefore, smaller ImageNet versions are suitable for studying how vision networks represent complex information hierarchies and the similarity between their concepts. Their direct connection with WordNet eliminates any ambiguities caused by the ambiguity of text labels of other datasets, for which the WordNet IDs are not given. Moreover, they are both current and important benchmarks in computer vision.

\textbf{CIFAR100} \cite{bib:krizhevsky2009learning} is a dataset of a similar size as Mini-ImageNet, however with significantly smaller original sizes of the images (only 32x32 pixels). With its relatively high number of classes, it is a good alternative to Mini-ImageNet; therefore, we use it to examine whether our observations based on the Mini-ImageNet generalize to other datasets for object recognition.

Additionally, we also use some example models trained on \textbf{CIFAR10} \cite{bib:krizhevsky2009learning} and \textbf{COCO 2017} \cite{bib:lin2014microsoft}. The first dataset is a small version of CIFAR with only 10 classes. Its classes are also at a more basic-level of the semantic hierarchy than the ones of CIFAR100. COCO 2017, on the other hand, is a dataset that includes the labels for object detection and segmentation, representing a dataset used for other tasks. It includes 80 different natural categories. We use the models trained on these datasets to further (qualitatively) examine the generalizability of our findings (whether the hierarchical similarity structure develops in their NCSMs).

\subsection{Models used in the study}

In our main experiments, we use the following state-of-the-art models: 2 standard CNNs (ResNet18 \cite{bib:he2016deep}, MobileNetV2 \cite{bib:sandler2018mobilenetv2}), 1 CNN 'modernized' with the techniques from the ViT domain - ConvNeXt-T \cite{bib:liu2022convnet}, 2 ViTs (ViTB \cite{bib:dosovitskiy2020image}, SwinV2T \cite{bib:liu2022swin} and 1 hybrid model (MaxViTT \cite{bib:tu2022maxvit} - it uses an attention model blended with convolutions). We use their implementations provided via the torchvision Python library. They represent older and more recent CNNs and ViTs (and models that use techniques borrowed from the contrasting architecture). We train all the models from scratch to examine how their perception of similarity changes during training under the assumption that no knowledge was present in a given classifier before training. We do not use any techniques to enforce the development of similarity perception nor hierarchy to examine whether these phenomena do and how they self-emerge in networks trained with standard training procedures. We use standard data augmentation techniques suitable for a given model architecture: for CNNs, we use Random Resized Crop, Random Horizontal Flip, Random Rotation, Gaussian Blur, Color Jitter, Random Perspective Transformation, and Random Affine Transformation. For ViTs, we also use Cutmix \cite{bib:yun2019cutmix}. We use a scheduler with a linear warmup and Reduce On Plateau (reproducibility: see our GitHub repository - supplementary materials and Zenodo during the revision stage - for the specific values of their parameters and random seeds). We also use some additional models in the additional experiments in the appendices. We provide their names and links to access them in the sections regarding experiments with these particular models for clarity.

Our experiments were performed on a Linux-based system within a high-powered computer center computation grid, with 2x GPGPU NVIDIA A100 and 80 GB RAM per task. The average training time per experiment was approximately 26 hours for 400 epochs. The additional complexity of computing our new metrics accounted for approximately 0.29\% of epoch time computation for MobileNetV2, and it stayed consistent for other networks as well (see Tab. \ref{tab:epoch_time} with the results for 3 other models), making it an insignificant addition to the total training time.

\begin{table}[h!]
\centering
\caption{Computational time of our methods as a percentage of the epoch time computation for an example CNN and ViT.}
\begin{tabular}{lc}
\toprule
\textbf{Network} & \textbf{Epoch Time Percentage (\%)} \\
\midrule
MobileNetV2 & 0.29 \\
SwinV2T      & 0.11 \\
\bottomrule
\end{tabular}
\label{tab:epoch_time}
\end{table}

\newpage

\section{WordNet reference matrices and performance metrics for Mini-ImageNet and CIFAR100} \label{appendix_wordnet_accuracy}

In this appendix, we provide the generated WordNet semantic similarity matrices obtained for Mini-ImageNet (which we use to experiment with in the main paper) and CIFAR100 (used in additional experiments in Appendix \ref{appendix_cifar}). We also provide the test accuracy and loss plots for these two datasets to enable establishing the connection between the similarity and standard performance metrics.

\subsection{Semantic Class Similarity Matrices obtained for WordNet and the examined datasets}

\begin{figure}[h]
\begin{center}
\subfloat[][Mini-ImageNet]{\includegraphics[width=0.49\textwidth]{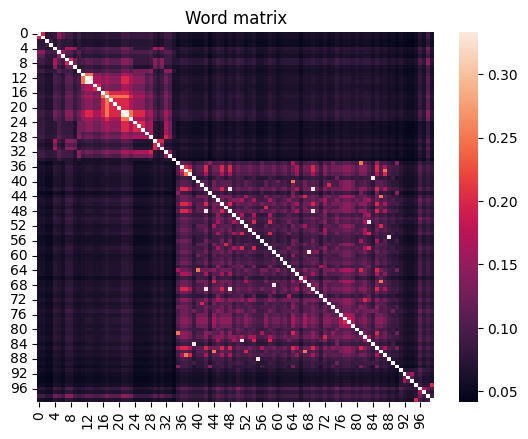}}
\subfloat[][CIFAR100]{\includegraphics[width=0.49\textwidth]{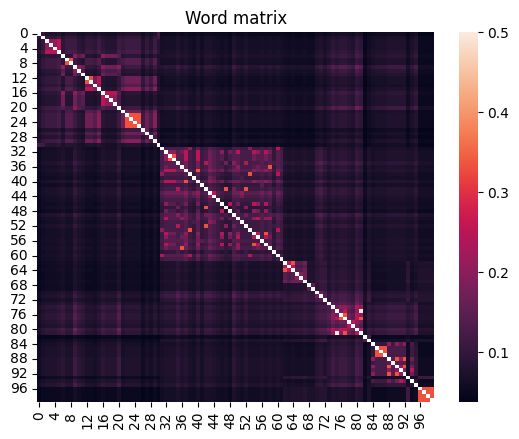}}
\end{center}
   \caption{Semantic Class Similarity Matrices (SCSM), precisely the WordNet Class Similarity Matrices (WCSM) obtained for Mini-ImageNet and CIFAR100. A clear hierarchical structure is visible for both datasets. }
\label{fig:wordnet_csm}
\end{figure}

For computing semantic similarity and obtaining Semantic Class Similarity Matrices, we use the \textbf{NLTK} framework~\cite{bird2009natural} along with WordNet linguistic taxonomy. The semantic similarity is computed for each pair of the classes in the dataset as an inverse of the distance of the shortest path connecting them in the linguistic taxonomy (path connecting two nodes in the WordNet taxonomy tree). This metric can take the values from 0 to 1, where higher values indicate semantically closer concepts.

In Figure \ref{fig:wordnet_csm}, we present the computed Semantic Class Similarity Matrices (SCSM), precisely the WordNet Class Similarity Matrices (WCSM) obtained for Mini-ImageNet and CIFAR100. There are two main basic-level semantic groups in WordNet: the first one (in the left upper corner of the matrix) contains different animals, and the second one - artificial objects. In the case of CIFAR100, the square in the left upper corner also corresponds to different living organisms, while the subsequent squares to different subgroups of the artificial objects and formations. Both heatmaps show the hierarchical nature of the semantic relations in the two examined datasets.

\subsection{The course of the networks training from the perspective of standard metrics}

\begin{figure}[t]
\begin{center}
\subfloat[][Testing accuracy - Mini-ImageNet]{\includegraphics[width=0.49\textwidth]{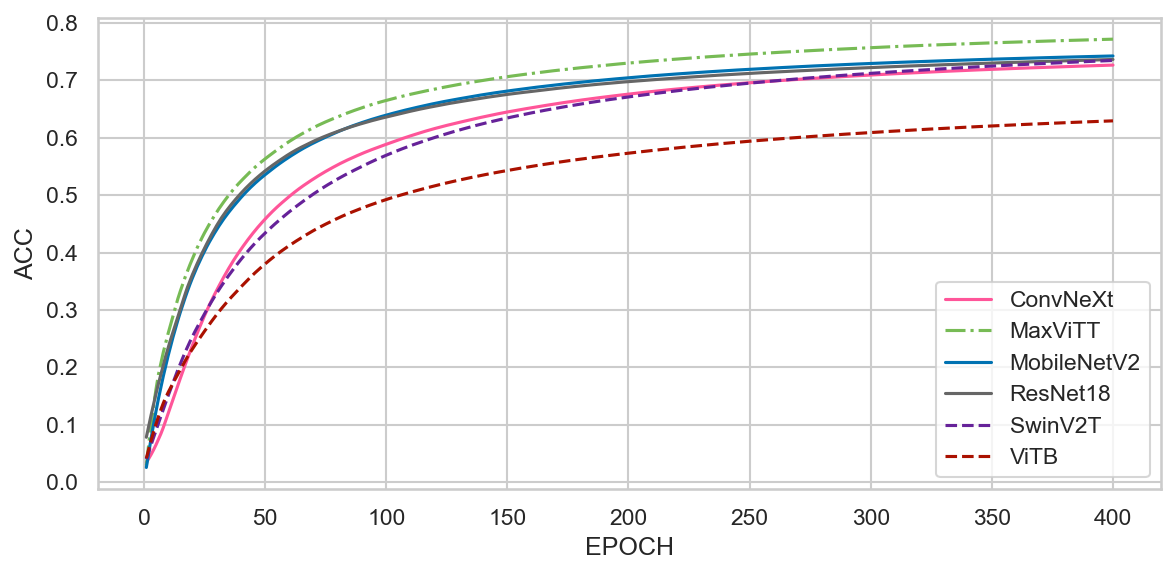}}
\subfloat[][Loss - Mini-ImageNet]{\includegraphics[width=0.49\textwidth]{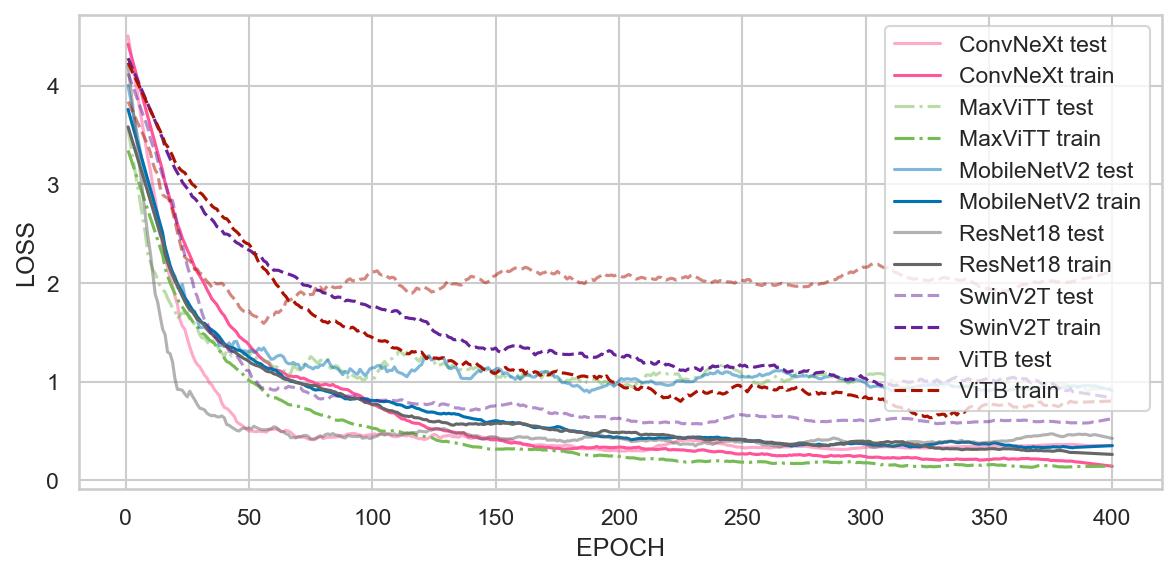}}
\\
\subfloat[][Testing accuracy - CIFAR100]{\includegraphics[width=0.49\textwidth]{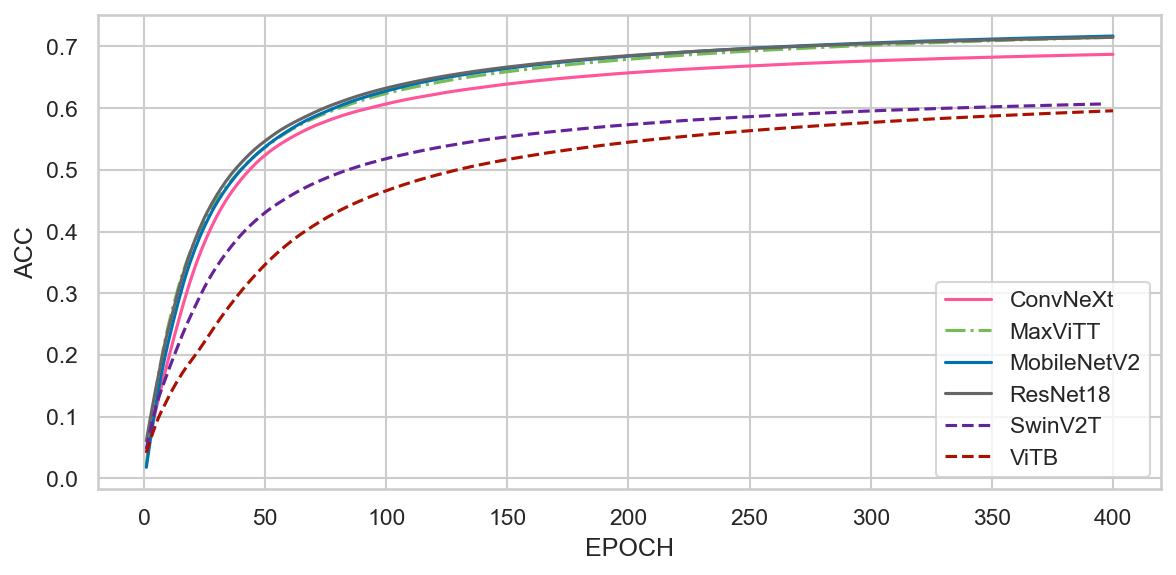}}
\subfloat[][Loss - CIFAR100]{\includegraphics[width=0.49\textwidth]{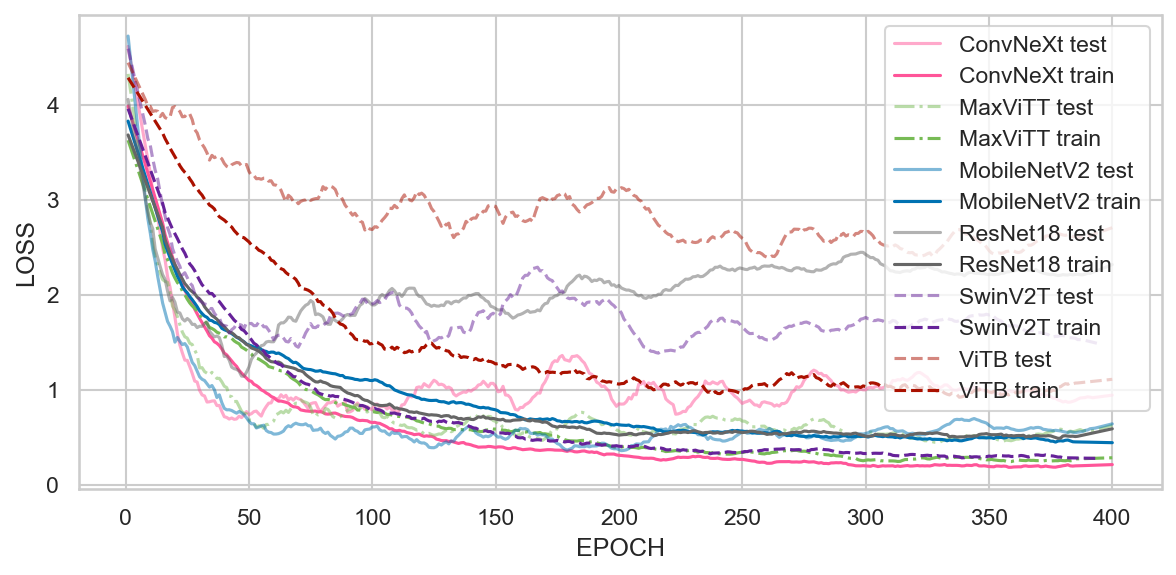}}
\end{center}
   \caption{Testing accuracy and Train/Test loss value curves for the two examined datasets: Mini-ImageNet and CIFAR100.}
\label{fig:accuracy_loss}
\end{figure}

In Fig. \ref{fig:accuracy_loss}, we provide the plots of testing accuracy and train/test loss curves for both the examined datasets - Mini-ImageNet and CIFAR100. We used the same hyperparameters to train the models on CIFAR100 that we initially used for Mini-ImageNet for the comparison. It is visible that while the testing accuracy of all networks is rather similar for the two datasets (one exception is the SwinV2T model on CIFAR100), the loss plots show higher overfitting of the models trained on CIFAR100 compared to Mini-ImageNet.

\section{Additional numerical results obtained on Mini- \\ ImageNet}\label{appendix_mini}

\begin{figure}[htbp]
    \centering
    \subfloat[Mean Squared Error]{
        \includegraphics[width=0.48\textwidth]
        {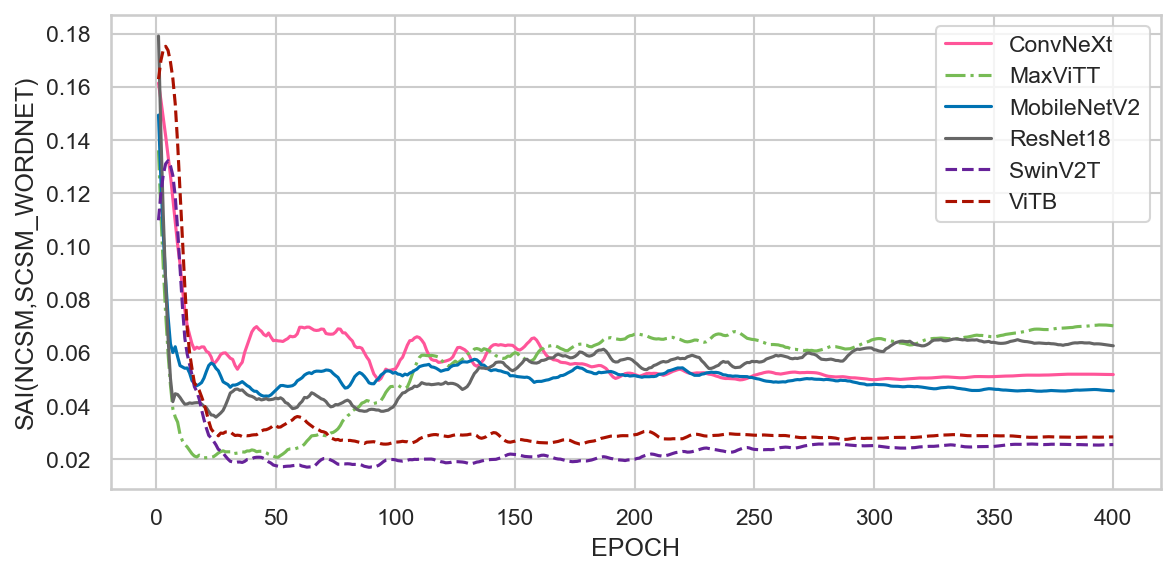}
    }
    \hfill
    \subfloat[Mean Absolute Error]{
        \includegraphics[width=0.48\textwidth]
        {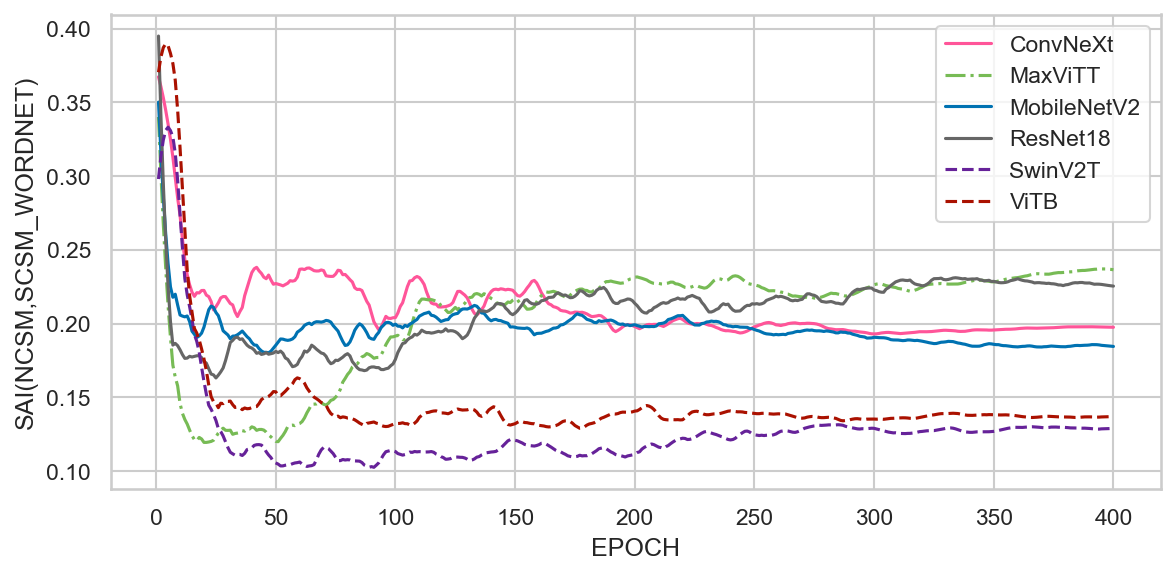}
    }

    \caption{Mini-ImageNet: Similarity Alignment Index curves based on
    distance measures for network and WordNet similarity perception,
    \textbf{SAI(NCSM, SCSM)}. Both measures show that networks quickly
    develop a similarity perception that largely aligns with semantic
    relations. Excluding some minor drops, this alignment persists as
    training continues.}
    \label{fig:mse_mae_mini}
\end{figure}

In this appendix, we provide the remaining results obtained for the Mini-ImageNet. In Fig. \ref{fig:mse_mae_mini}, we present the obtained \textbf{SAI(NCSM, SCSM)} for the distance measures: Mean Squared Error (MSE) and Mean Absolute Error (MAE). The results show that these measures also reflect the changes in the similarity alignment between the Network's and the Semantic perception of similarity.

\begin{figure}[h]
\centering
\subfloat[][ResNet18 (1)]{\includegraphics[width=0.25\textwidth]{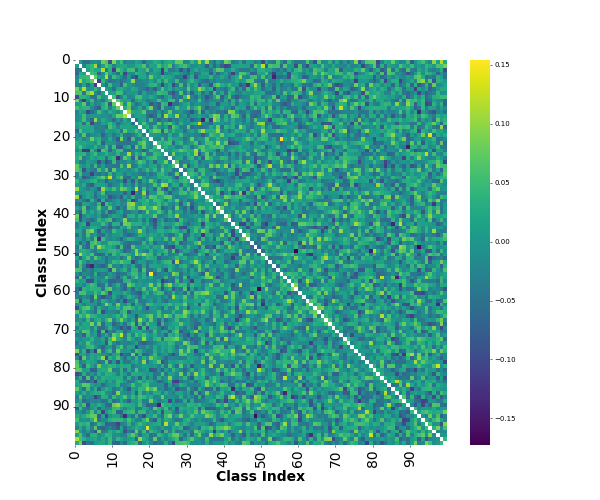}}
\subfloat[][ResNet18 (5)]{\includegraphics[width=0.25\textwidth]{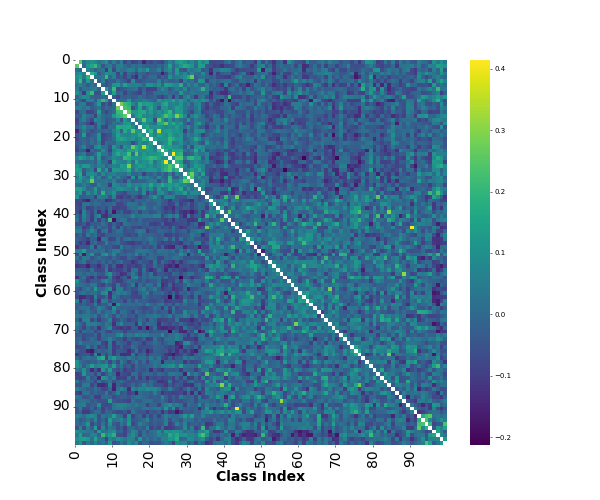}}
\subfloat[][ResNet18 (25)]{\includegraphics[width=0.25\textwidth]{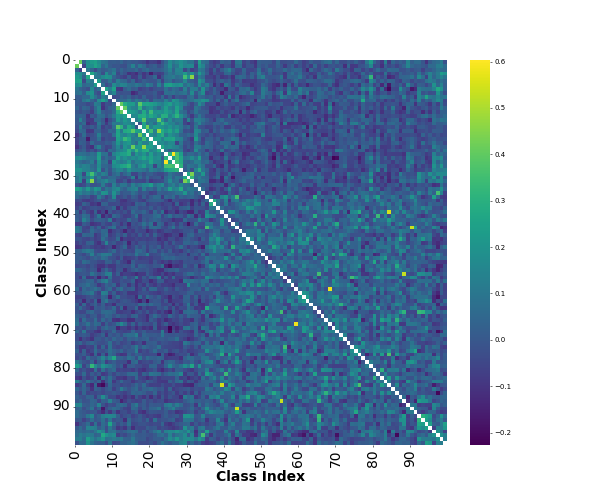}}
\subfloat[][ResNet18 (200)]{\includegraphics[width=0.25\textwidth]{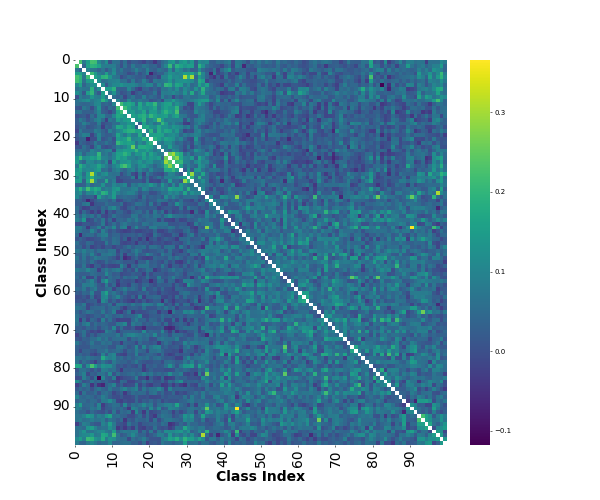}}
\\[-1em]
\subfloat[][SwinV2 (1)]{\includegraphics[width=0.25\textwidth]{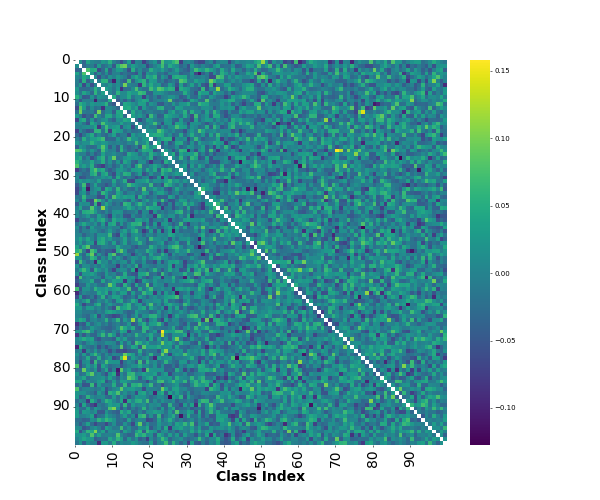}}
\subfloat[][SwinV2 (5)]{\includegraphics[width=0.25\textwidth]{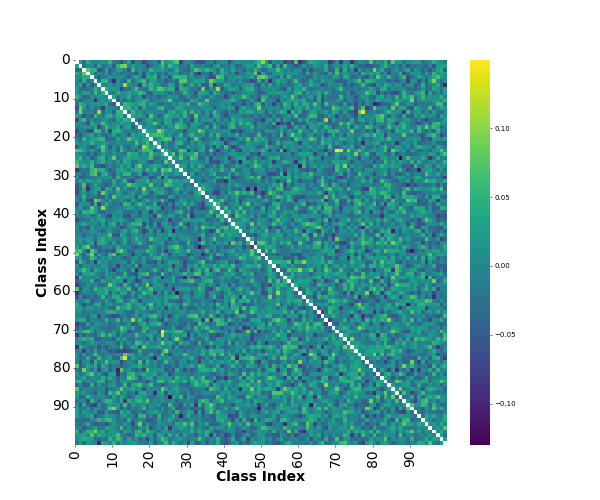}}
\subfloat[][SwinV2 (25)]{\includegraphics[width=0.25\textwidth]{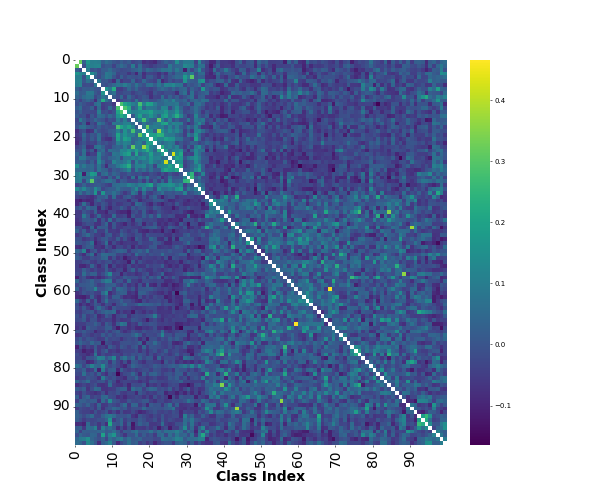}}
\subfloat[][SwinV2 (200)]{\includegraphics[width=0.25\textwidth]{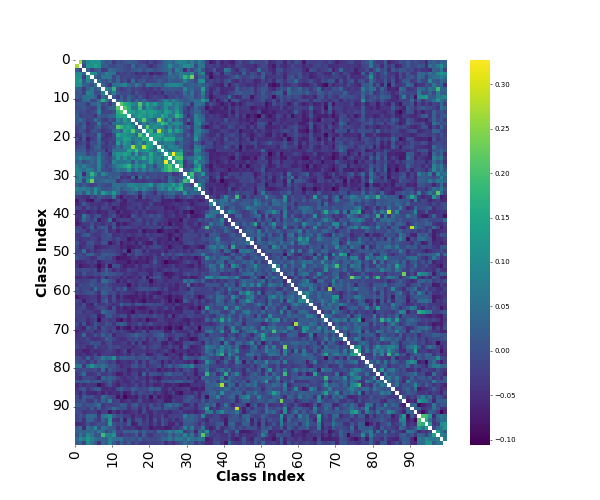}}
\caption{Mini-ImageNet: NCSMs of ResNet18 and SwinV2 (epoch number).  }
\label{fig:ncsms_resnet_swin_mini}
\end{figure}

\begin{figure}[h]
\centering
\subfloat[][ResNet18 (1)]{\includegraphics[width=0.25\textwidth]{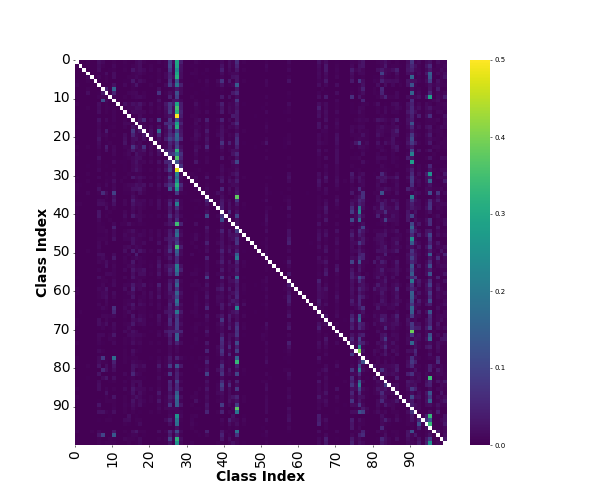}}
\subfloat[][ResNet18 (5)]{\includegraphics[width=0.25\textwidth]{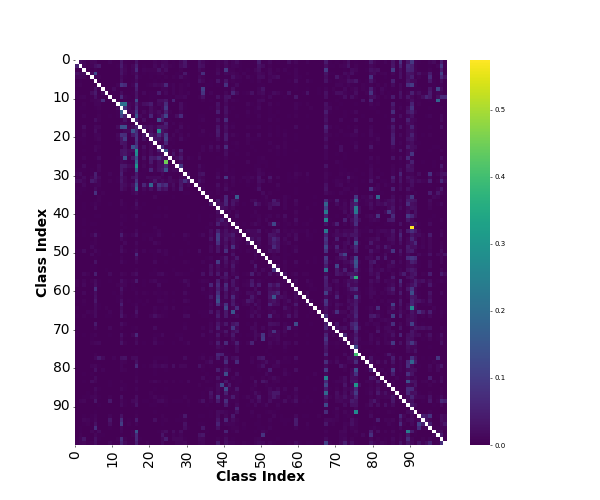}}
\subfloat[][ResNet18 (25)]{\includegraphics[width=0.25\textwidth]{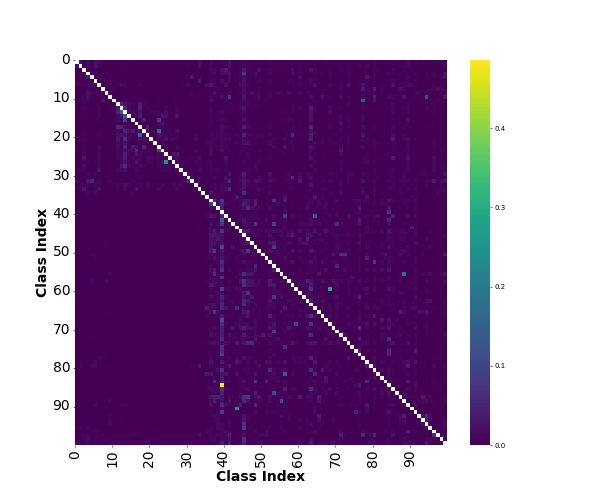}}
\subfloat[][ResNet18 (200)]{\includegraphics[width=0.25\textwidth]{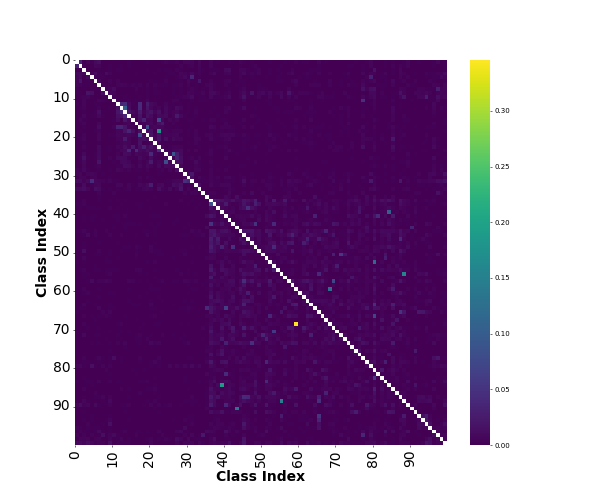}}
\\[-1em]
\subfloat[][SwinV2 (1)]{\includegraphics[width=0.25\textwidth]{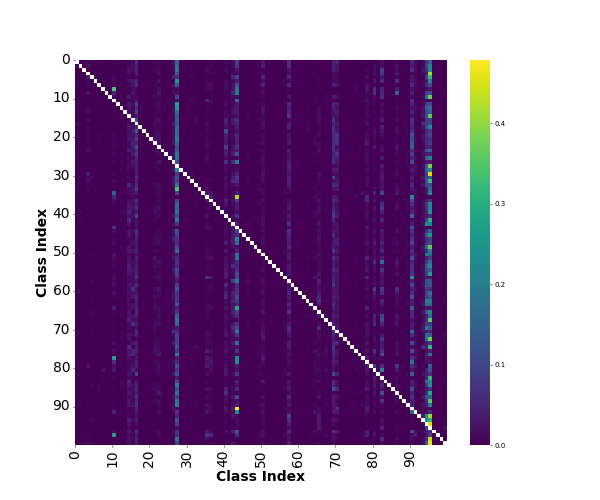}}
\subfloat[][SwinV2 (5)]{\includegraphics[width=0.25\textwidth]{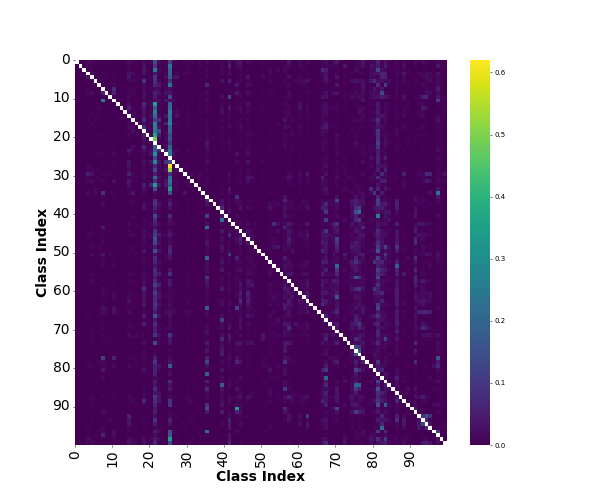}}
\subfloat[][SwinV2 (25)]{\includegraphics[width=0.25\textwidth]{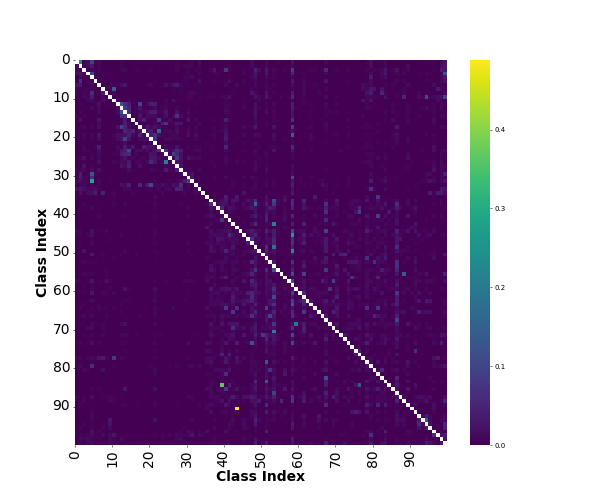}}
\subfloat[][SwinV2 (200)]{\includegraphics[width=0.25\textwidth]{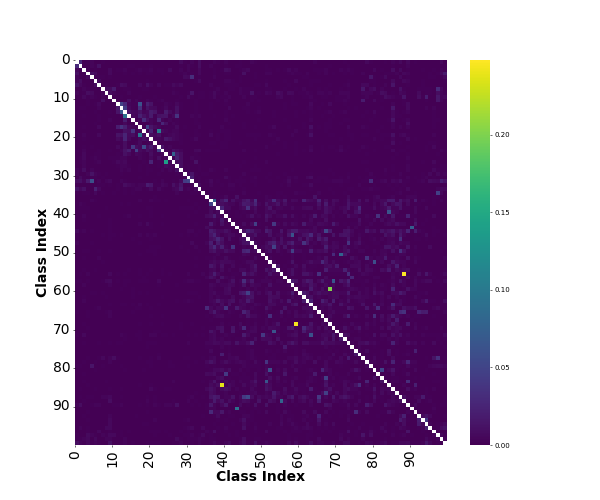}}
\caption{Mini-ImageNet: CCSMs of ResNet18 and SwinV2 (epoch number).  }
\label{fig:confusion_matrices_resnet_swin_mini}
\end{figure}
\FloatBarrier

\begin{figure}[h]
\centering
\subfloat[][ConvNeXt (1)]{\includegraphics[width=0.25\textwidth]{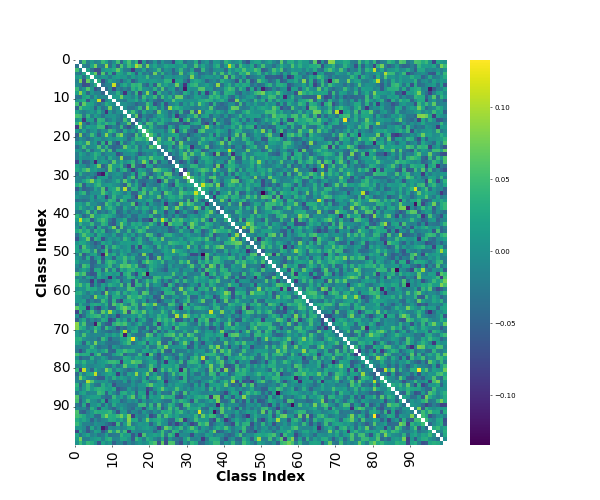}}
\subfloat[][ConvNeXt (5)]{\includegraphics[width=0.25\textwidth]{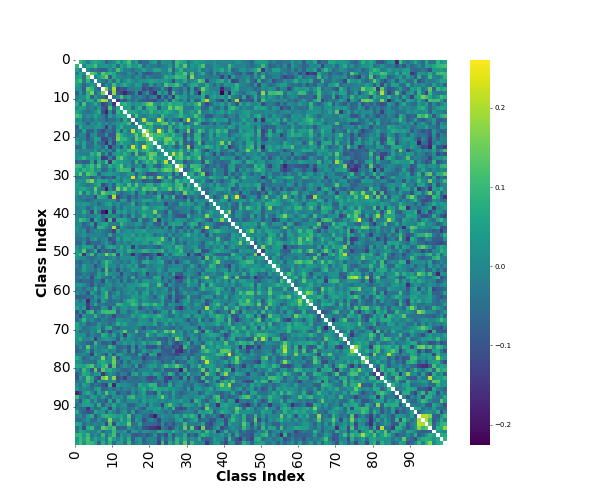}}
\subfloat[][ConvNeXt (25)]{\includegraphics[width=0.25\textwidth]{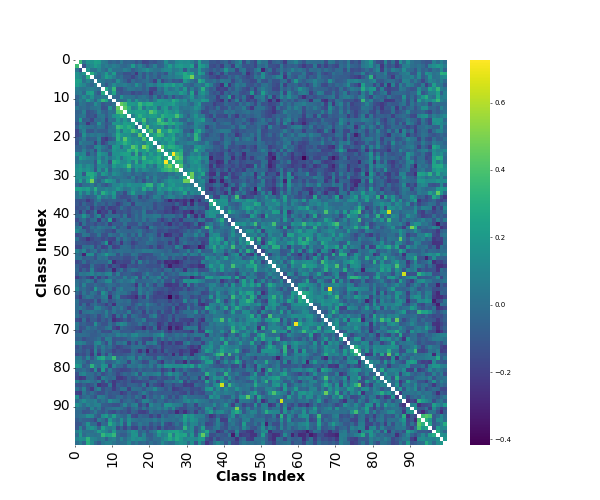}}
\subfloat[][ConvNeXt (200)]{\includegraphics[width=0.25\textwidth]{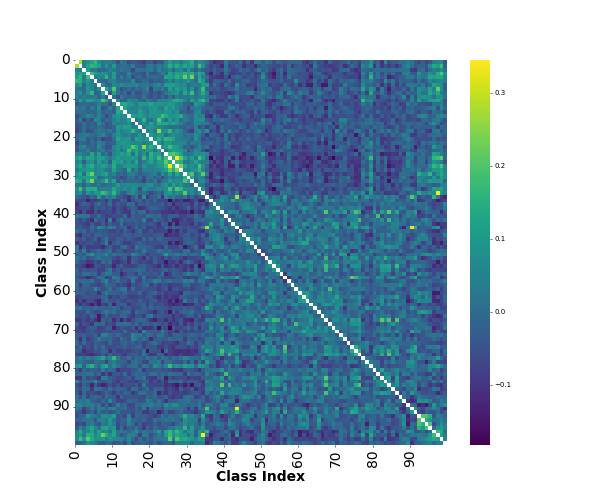}}
\\
\subfloat[][ViTB (1)]{\includegraphics[width=0.25\textwidth]{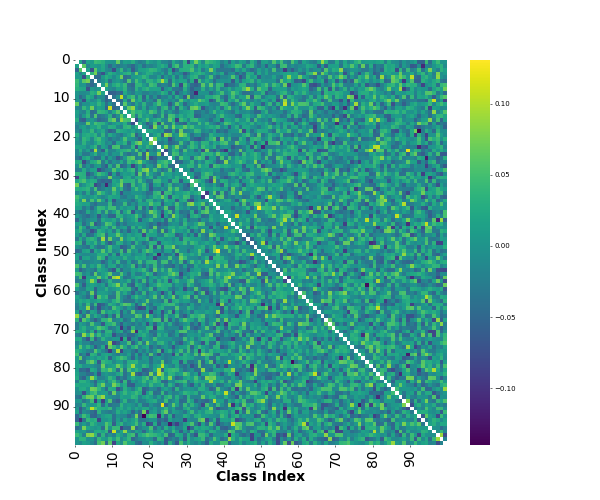}}
\subfloat[][ViTB (5)]{\includegraphics[width=0.25\textwidth]{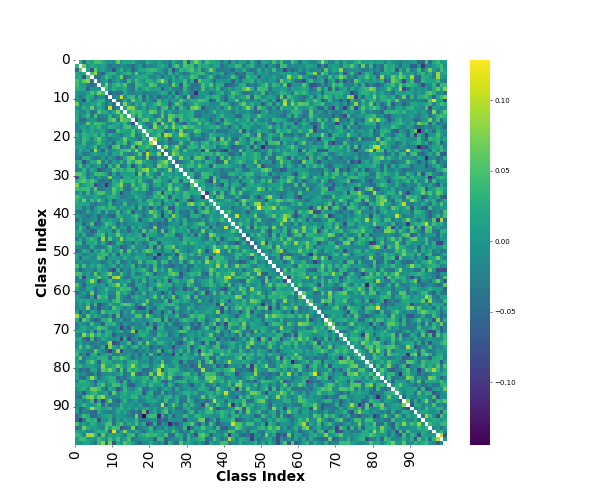}}
\subfloat[][ViTB (25)]{\includegraphics[width=0.25\textwidth]{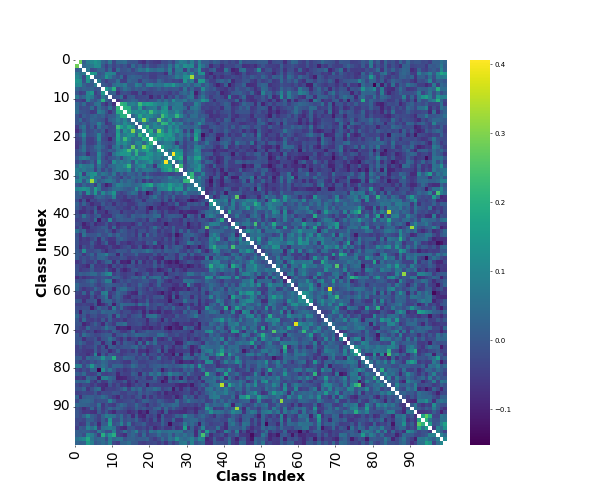}}
\subfloat[][ViTB (200)]{\includegraphics[width=0.25\textwidth]{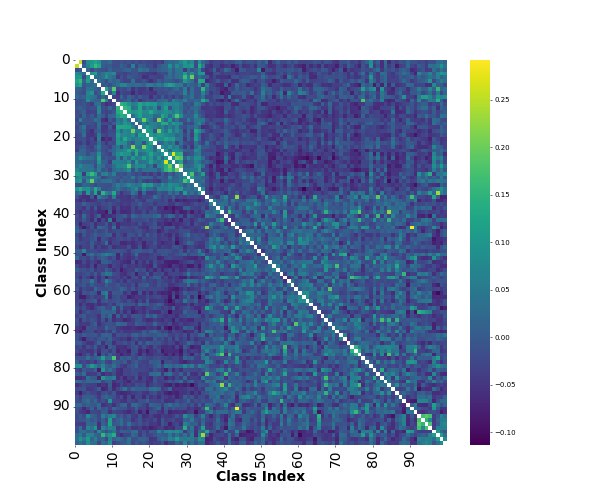}}
\\
\subfloat[][MobileNetV2 (1)]{\includegraphics[width=0.25\textwidth]{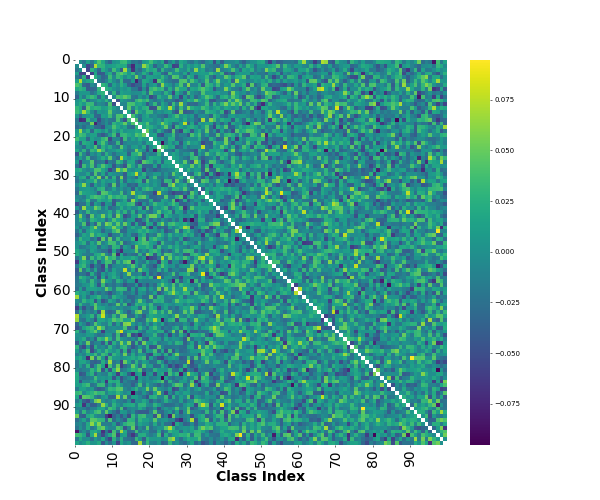}}
\subfloat[][MobileNetV2 (5)]{\includegraphics[width=0.25\textwidth]{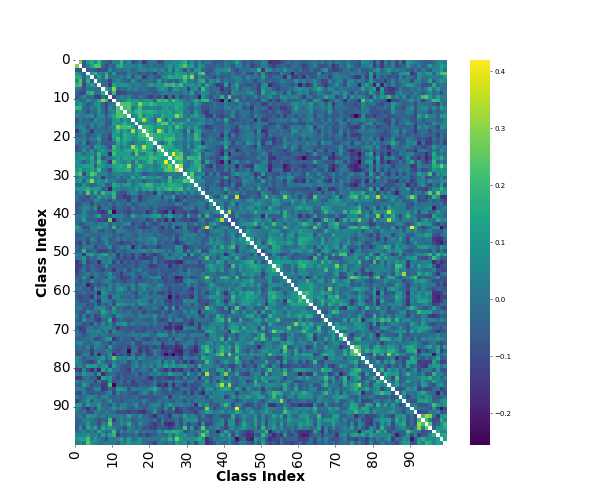}}
\subfloat[][MobileNetV2 (25)]{\includegraphics[width=0.25\textwidth]{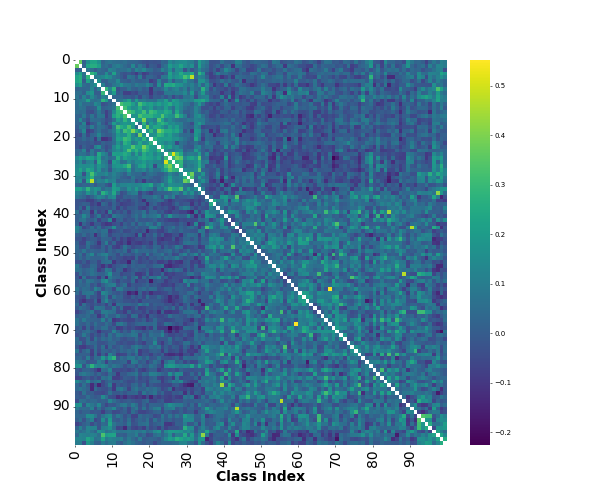}}
\subfloat[][MobileNetV2 (200)]{\includegraphics[width=0.25\textwidth]{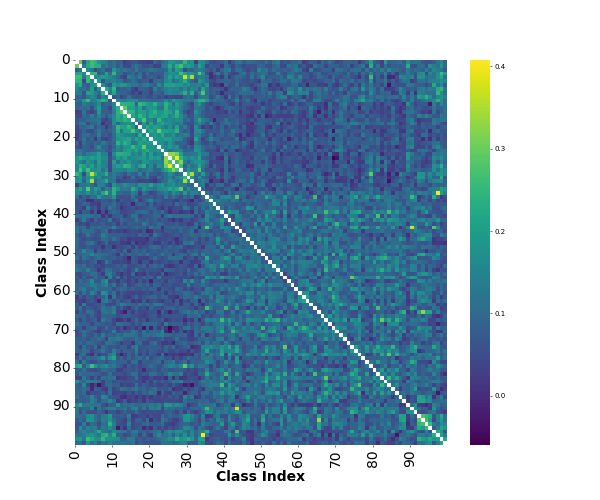}}
\\
\subfloat[][MaxViTT (1)]{\includegraphics[width=0.25\textwidth]{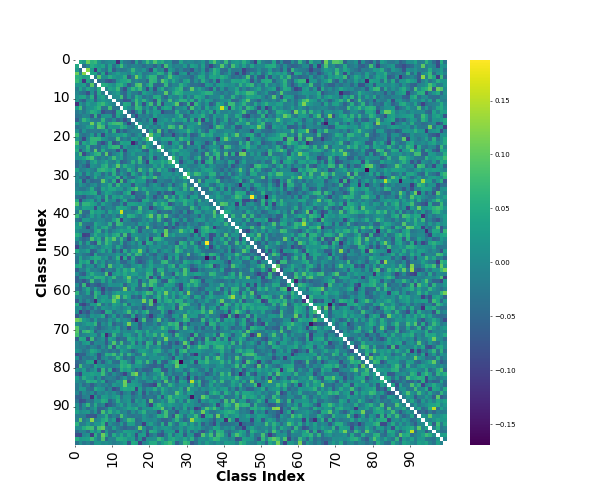}}
\subfloat[][MaxViTT (5)]{\includegraphics[width=0.25\textwidth]{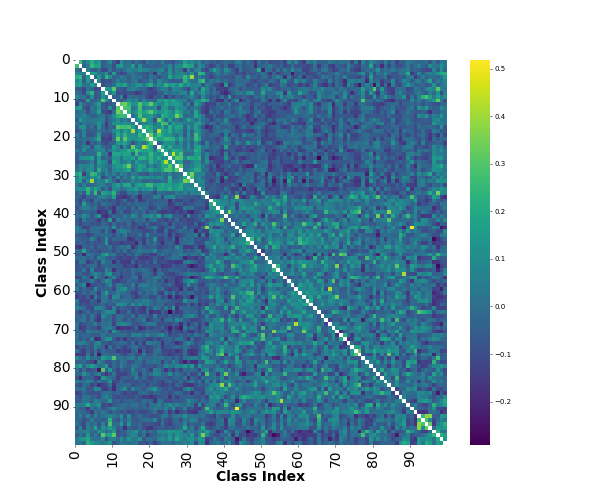}}
\subfloat[][MaxViTT (25)]{\includegraphics[width=0.25\textwidth]{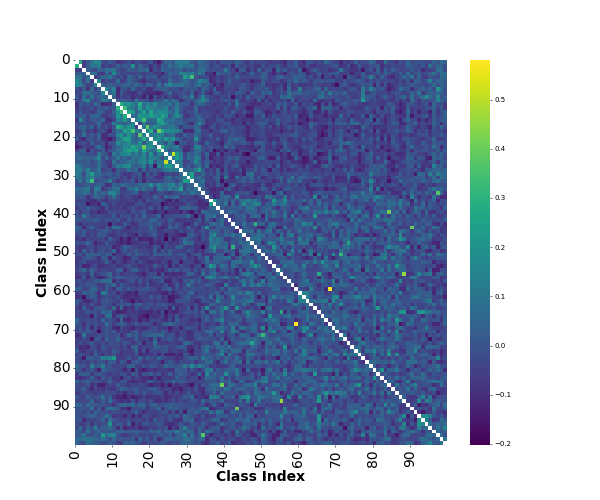}}
\subfloat[][MaxViTT (200)]{\includegraphics[width=0.25\textwidth]{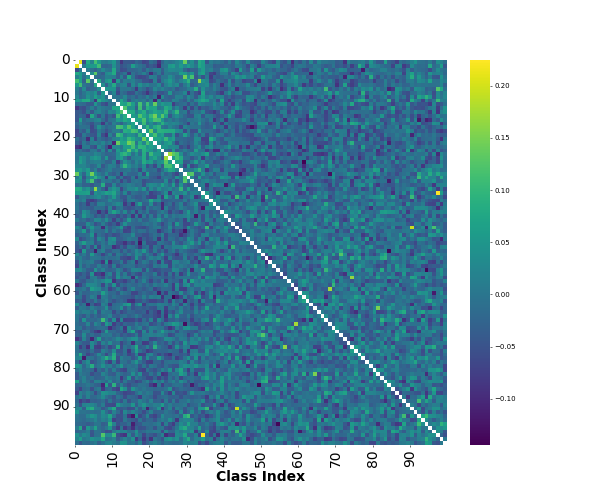}}
\caption{Mini-ImageNet: Network Class Similarity Matrices - Network-based similarity of the remaining models for at different epochs (number - in brackets). Networks develop the hierarchical similarity perception in the early epochs (ResNet earlier than Swin). While the example ViT eliminates less significant similarities in the later epochs, more semantically unrelated categories emerge as more similar for ResNet18 (visible as off-diagonal 'noise').  }
\label{fig:ncsms_mini_conv_vit}
\end{figure}

In Fig. \ref{fig:ncsms_resnet_swin_mini}, we present 4 Network Class Similarity Matrices for ResNet18 and SwinV2T (1th, 5th, 25th, 200th epoch). In Fig. \ref{fig:ncsms_mini_conv_vit}, we present the Network Class Similarity Matrices for the remaining models. Similarly to the qualitative results obtained in the main body of the paper for the ResNet18 and the SwinV2T models, the matrices show that a clear hierarchical similarity structure is developed faster for CNNs than for ViTs. It is also visible that in later epochs of training, ConvNeXt, similarly to ResNet18 discovers more similarities between the classes that do not belong to the main semantic groups (the 'off-diagonal noise') than the ViT model. It is also visible that ConvNeXt, similarly to ViTs (from which it incorporates some architectural features), needs more time to develop a clear similarity structure than standard CNNs. MaXViT (a hybrid model), on the other hand, needs less epochs than other ViTs to develop such a structure, further suggesting that the introduction of techniques from one architecture to another results in the intermingling of behavioral features of both architectures.

\begin{figure}[h]
\centering
\subfloat[][ConvNeXt (1)]{\includegraphics[width=0.25\textwidth]{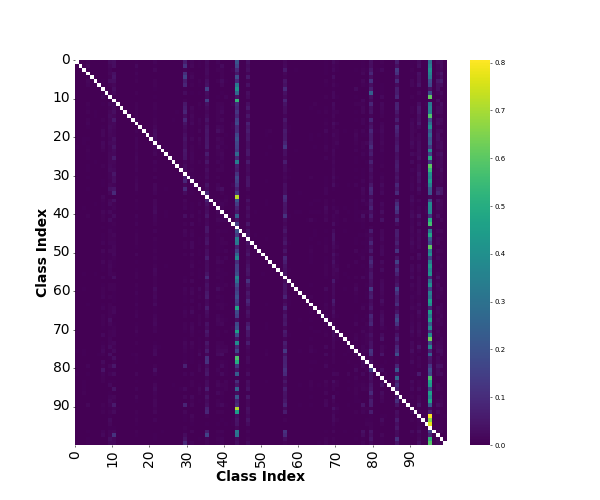}}
\subfloat[][ConvNeXt (5)]{\includegraphics[width=0.25\textwidth]{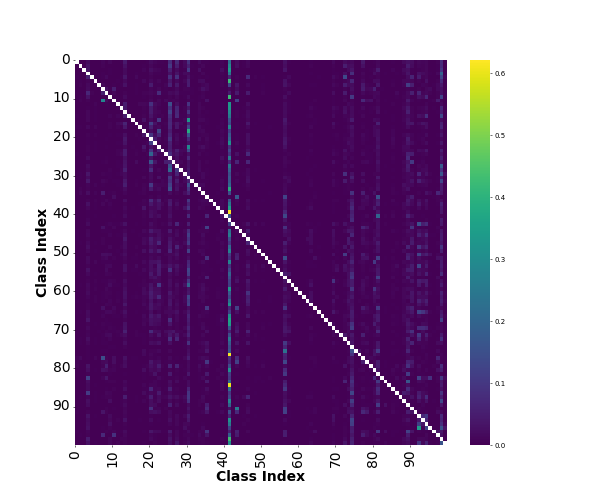}}
\subfloat[][ConvNeXt (25)]{\includegraphics[width=0.25\textwidth]{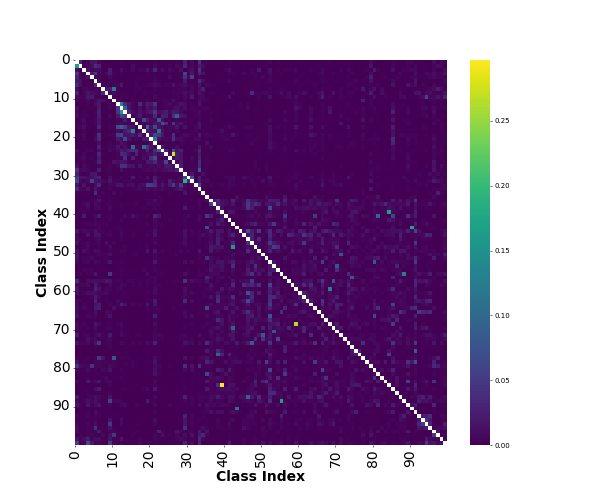}}
\subfloat[][ConvNeXt (200)]{\includegraphics[width=0.25\textwidth]{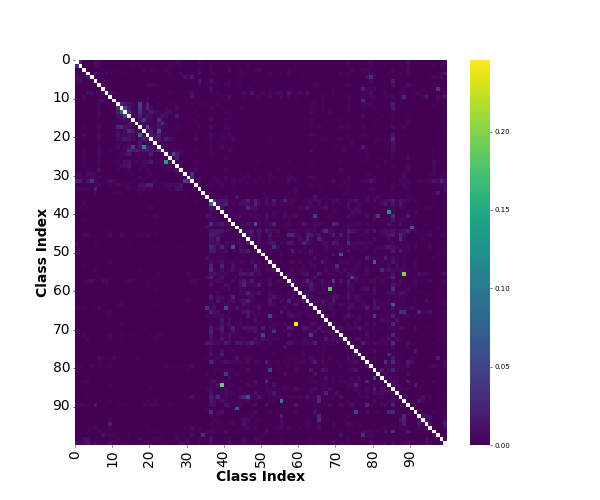}}
\\
\subfloat[][ViTB (1)]{\includegraphics[width=0.25\textwidth]{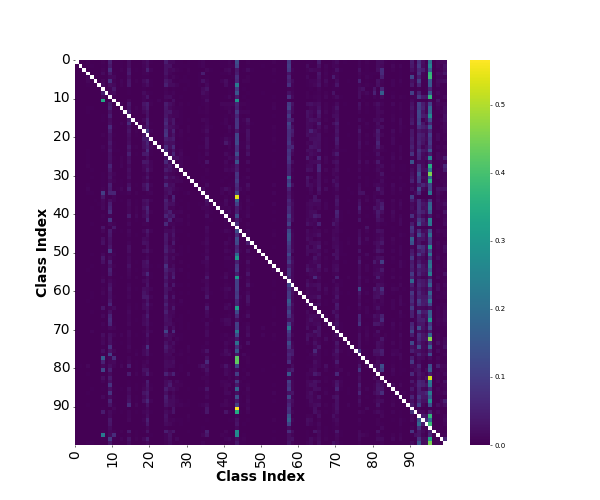}}
\subfloat[][ViTB (5)]{\includegraphics[width=0.25\textwidth]{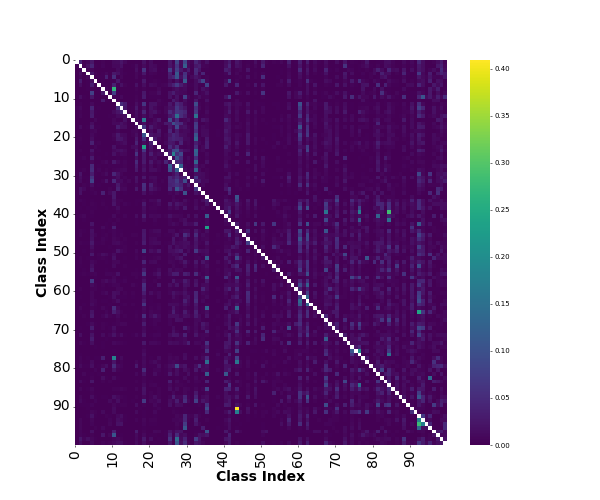}}
\subfloat[][ViTB (25)]{\includegraphics[width=0.25\textwidth]{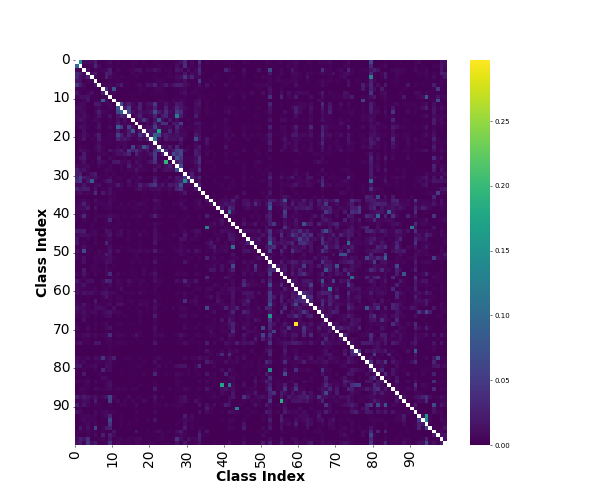}}
\subfloat[][ViTB (200)]{\includegraphics[width=0.25\textwidth]{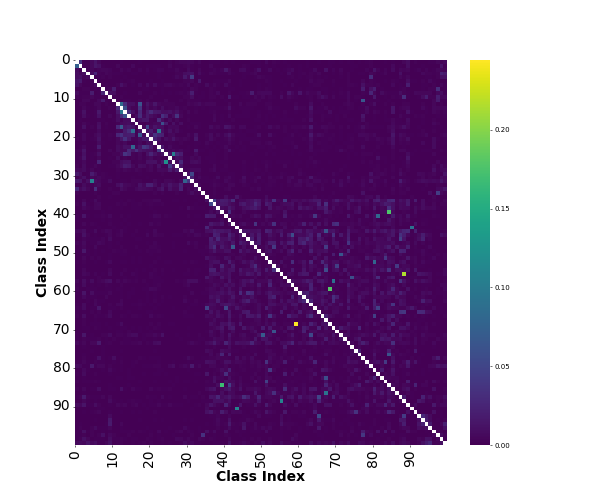}}
\\
\subfloat[][MobileNetV2 (1)]{\includegraphics[width=0.25\textwidth]{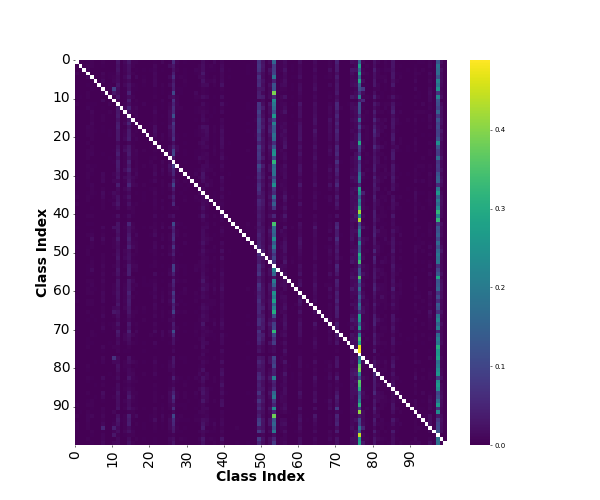}}
\subfloat[][MobileNetV2 (5)]{\includegraphics[width=0.25\textwidth]{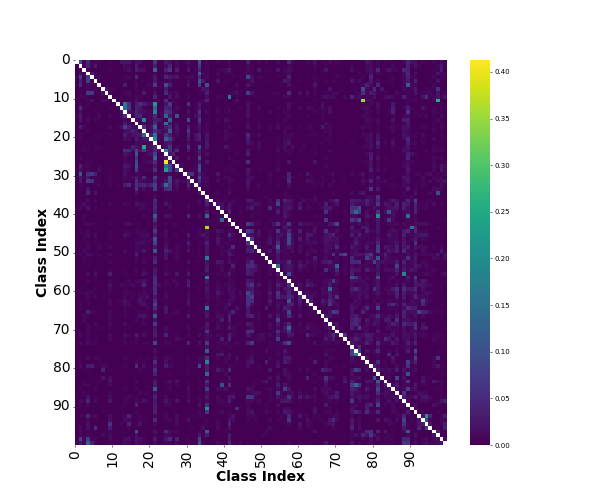}}
\subfloat[][MobileNetV2 (25)]{\includegraphics[width=0.25\textwidth]{FIGS/ImageNet_matrices/MobileNetV24_Confusion.png}}
\subfloat[][MobileNetV2 (200)]{\includegraphics[width=0.25\textwidth]{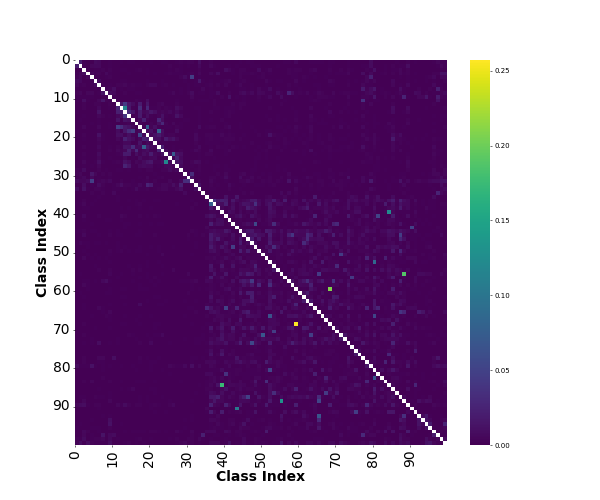}}
\\
\subfloat[][MaxViTT (1)]{\includegraphics[width=0.25\textwidth]{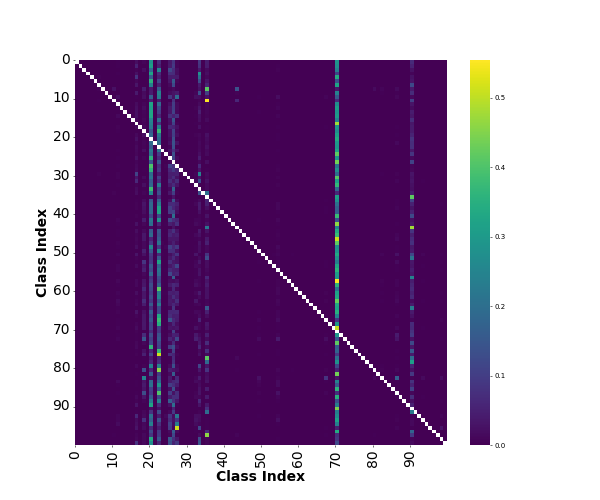}}
\subfloat[][MaxViTT (5)]{\includegraphics[width=0.25\textwidth]{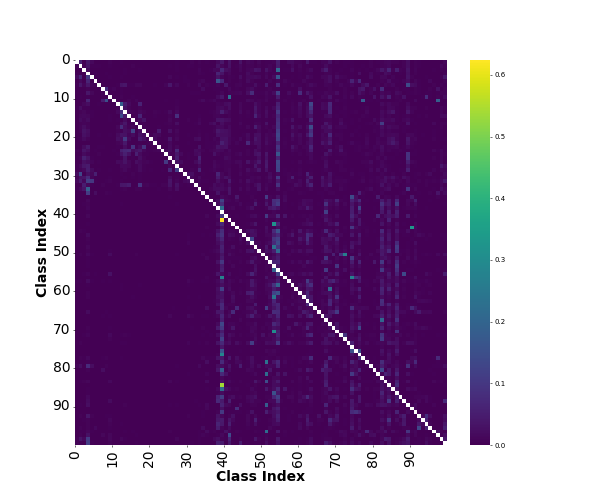}}
\subfloat[][MaxViTT (25)]{\includegraphics[width=0.25\textwidth]{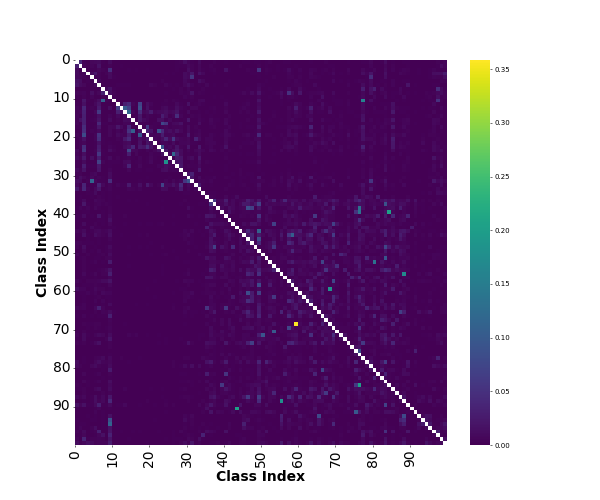}}
\subfloat[][MaxViTT (200)]{\includegraphics[width=0.25\textwidth]{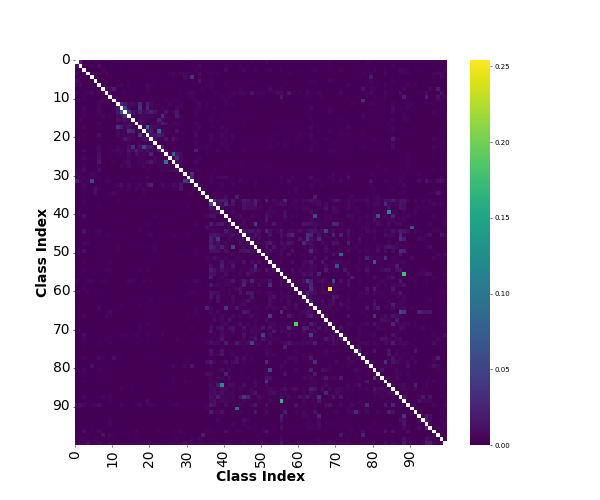}}
\caption{Mini-ImageNet: Confusion-based similarity of the remaining models at different epochs (number - in brackets). At the beginning, both networks targets only a few distinct classes as a confusion result. They initially cover the whole space, and then smaller and smaller groups of hierarchy, making the mistakes more distributed and as a result - clearly showing the hierarchy (SwinV2 needs more epochs than ResNet18 to achieve this). }
\label{fig:ccsms_mini_conv_vit}
\end{figure}

In Fig. \ref{fig:confusion_matrices_resnet_swin_mini}, we present 4 Confusion-based Class Similarity Matrices for ResNet18 and SwinV2T (1th, 5th, 25th, 200th epoch). In Fig. \ref{fig:ccsms_mini_conv_vit}, we present the Confusion-based Class Similarity Matrices for the remaining models (ConvNeXt, ViTB, MaxViTT, MobileNetV2). Similarly to the qualitative results obtained in the main body of the paper for the ResNet18 and the SwinV2T models, the matrices show that the confusion patterns after app. 25 epochs of training reveal a hierarchical similarity structure. Again, faster for  CNNs/the hybrid model than for ViTs. Especially for the categories from the animals basic-level category, it can be observed that the mistakes are made mainly within narrower semantic categories in the later epochs of training.

\clearpage

\section{Experiments with CIFAR100}\label{appendix_cifar}


To increase the generalizability of our findings, we conducted experiments analogous to those performed on Mini-ImageNet on the CIFAR100 dataset. In order to focus on the impact of the training data on the behavior of the network and exclude other factors, we decided to use the same hyperparameters of the models as for Mini-ImageNet (reproducibility: the configuration files of the models used in the experiments can be found in our GitHub repository - supplementary materials and Zenodo during the revision stage). We also train our models for 400 epochs and inspect the models during the training procedure with the implementation of our Deep Similarity Inspector Framework and save the result for presentation. In this appendix, we present analogical results for CIFAR100 to those obtained on Mini-ImageNet and discuss them shortly.

\subsection{How does the network’s similarity perception change throughout the training process for CNNs and ViTs? Is it in line with semantic similarity?}

\begin{figure}[h]
\begin{center}
\subfloat[][Mean Weights Similarity]{\includegraphics[width=0.49\textwidth]{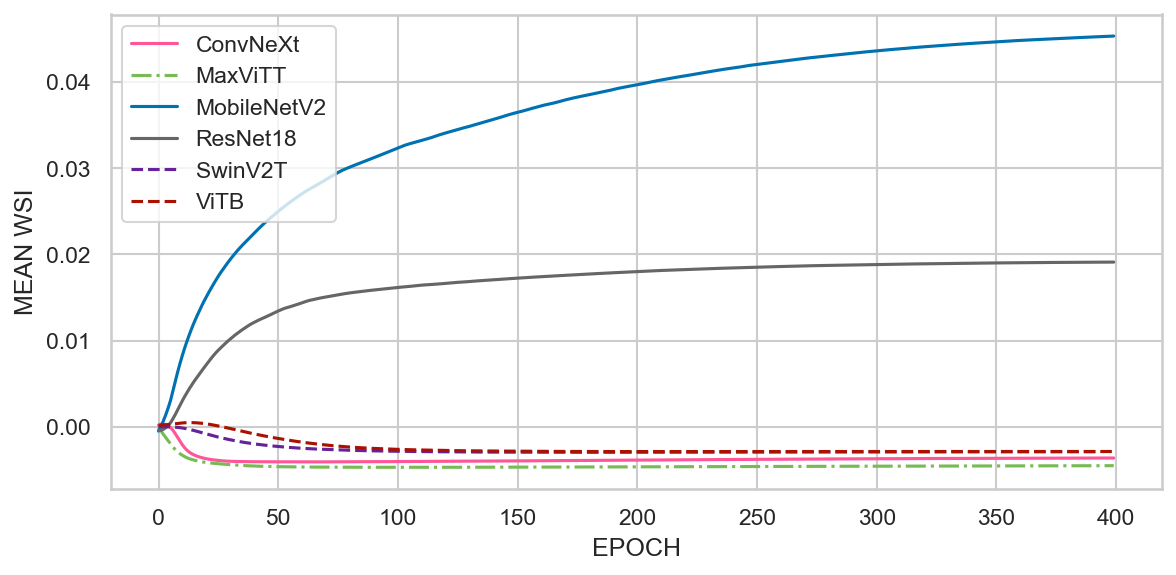}}
\subfloat[][Max Weights Similarity]{\includegraphics[width=0.49\textwidth]{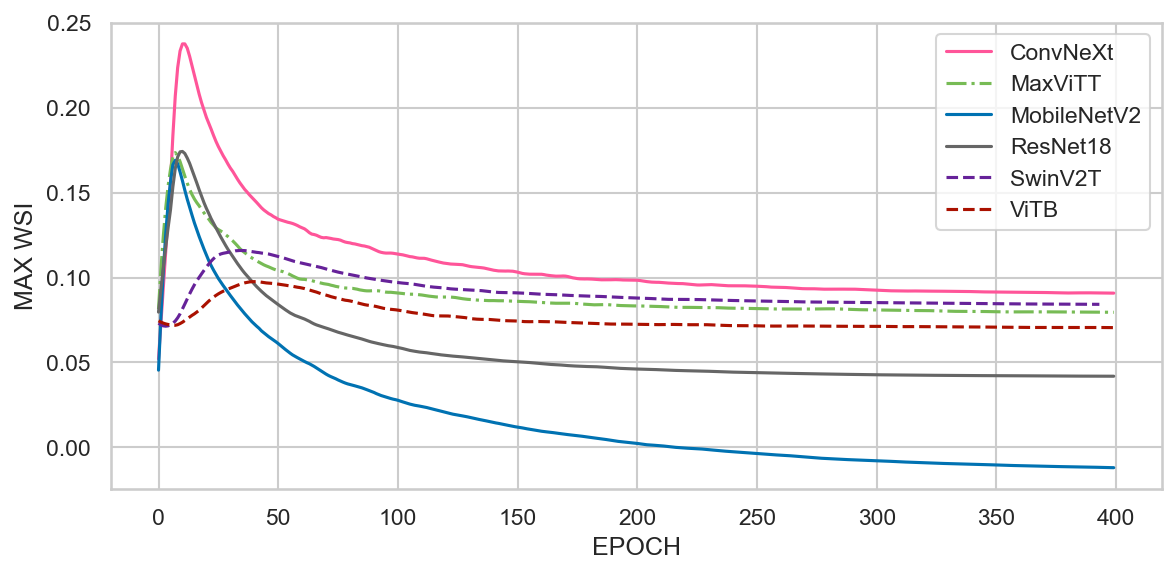}}
\\
\subfloat[][Min Weights Similarity]{\includegraphics[width=0.49\textwidth]{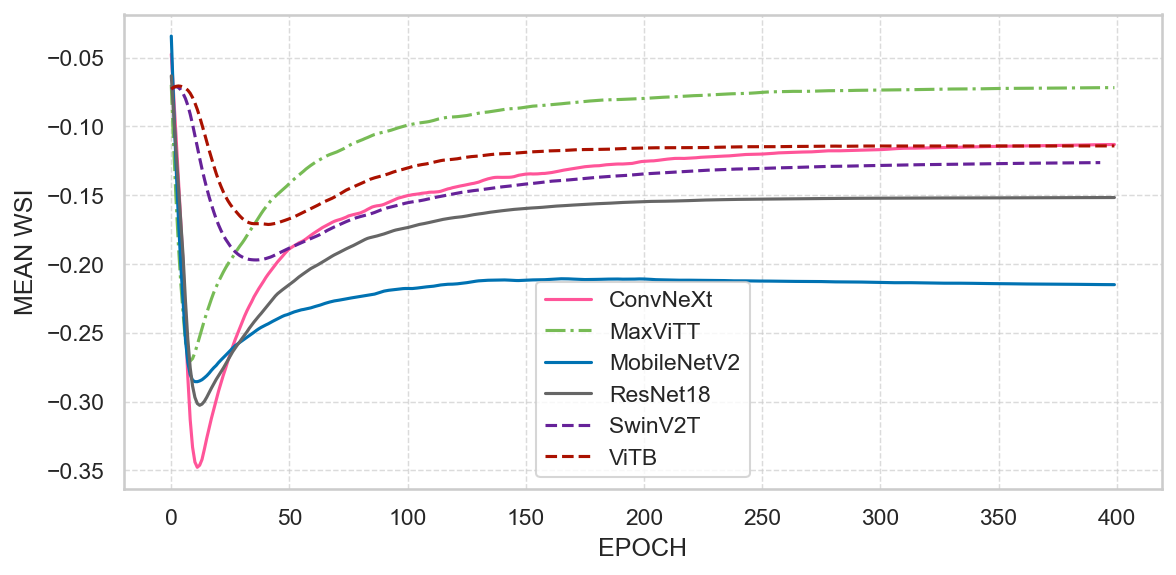}}
\end{center}
\caption{CIFAR100: Weights Similarity Index (WSI) Curves. The min and max variants maintain an approximately inverse relationship. The variants also show similarities within the network families (ViTs, CNNs) in terms of the changes in the perception of the most/least similar categories.}
\label{fig:wsi_cifar100}
\end{figure}

In Fig. \ref{fig:wsi_cifar100}, we present the results of different variants of the Weights Similarity Index (WSI) Curves obtained for the CIFAR100-trained models. The results are in line with the results obtained for Mini-ImageNet. It shows that the chosen network architecture and its hyperparameters impact the behavior of the network more than the chosen dataset. Again, for the majority of networks, Mean WSI drops with training for ViTs, a hybrid model and a ViT-inspired CNN. Standard CNNs, on the other hand, are characterized with Mean WSI increases, but with values still close to 0. Also, the Max/Min WSI variants behave in the same manner: the examined CNNs are characterized by a steep increase/decrease followed by a steep decrease/increase respectively, while the ViTs behave more steadily with their changes in similarity perception of templates. The hybrid model and the ViT-inspired CNN firstly behave similarly to CNNs, to get closer to the ViT behavior (via values) in the later stages of training.

\begin{figure}[h]
\begin{center}
\subfloat[][Cosine Similarity]{\includegraphics[width=0.49\textwidth]{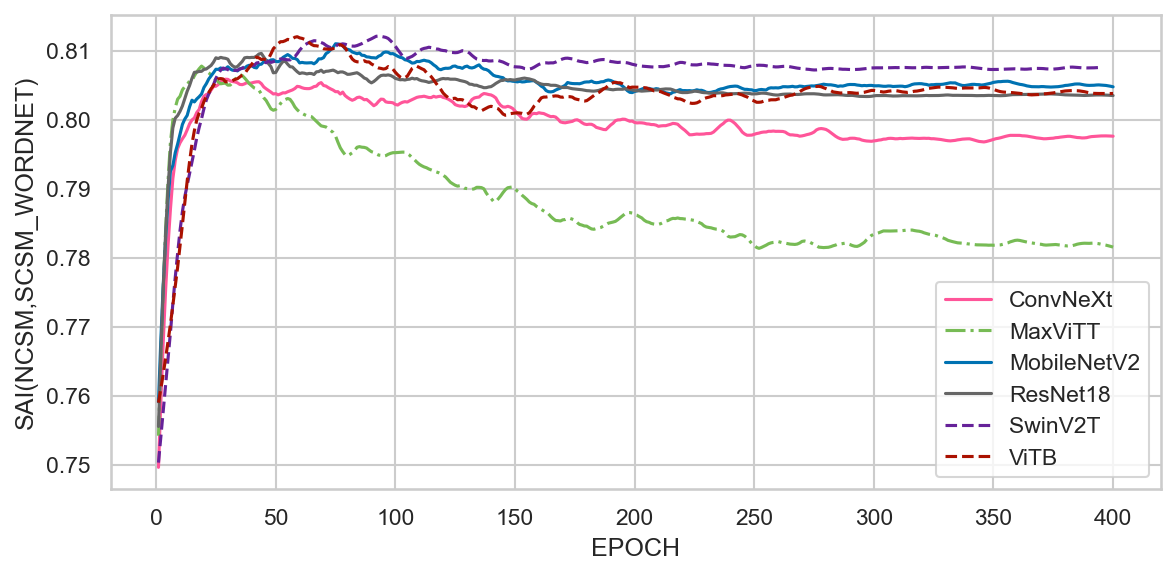}}
\subfloat[][Structural Similarity]{\includegraphics[width=0.49\textwidth]{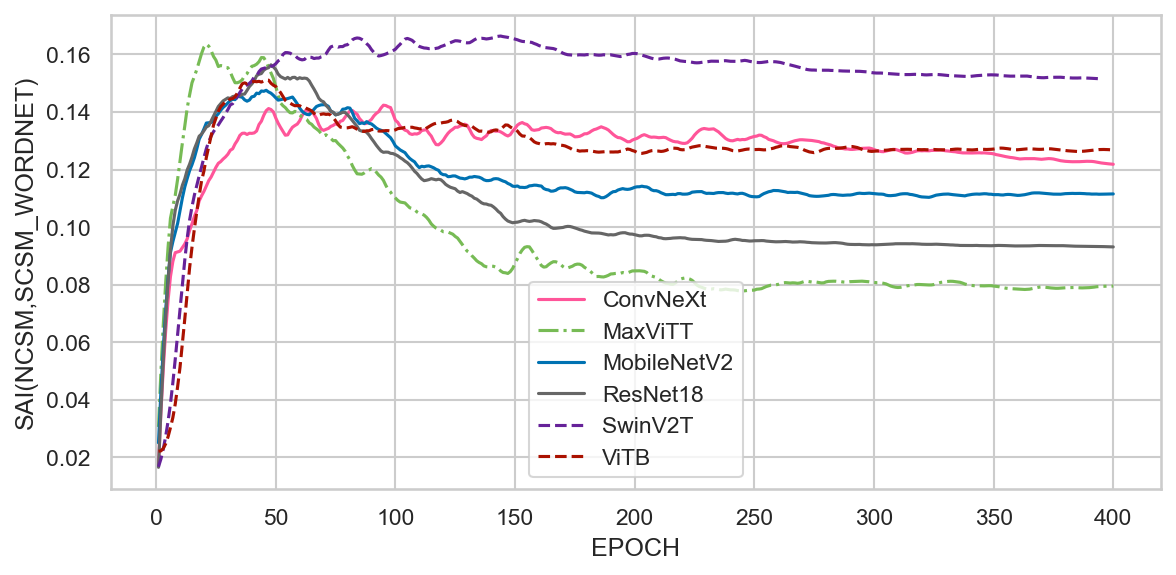}}
\\
\subfloat[][Mean Squared Error]{\includegraphics[width=0.49\textwidth]{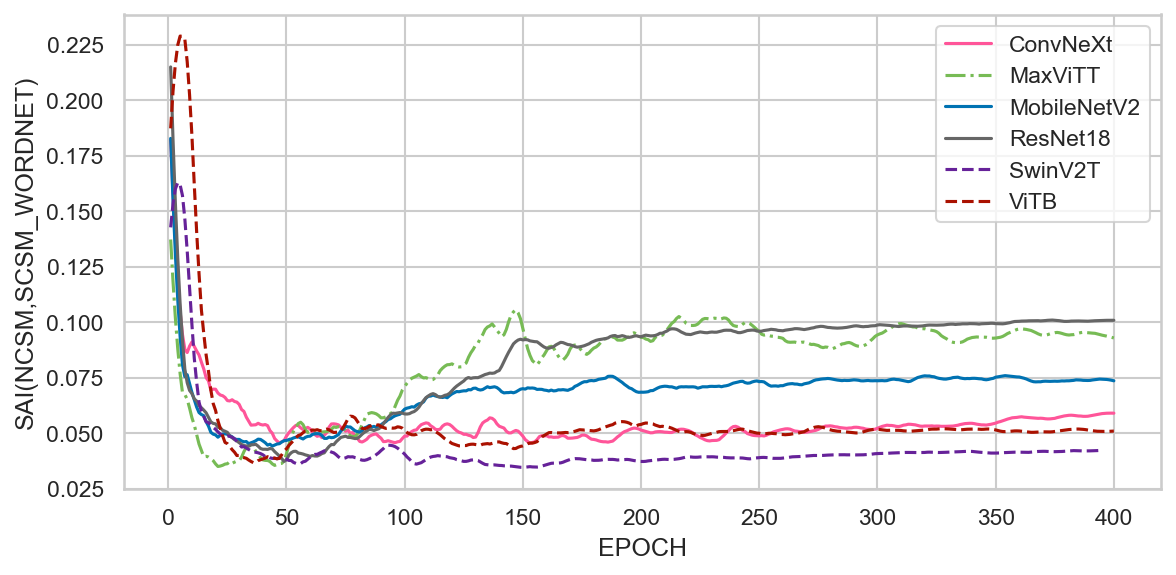}}
\subfloat[][Mean Absolute Error]{\includegraphics[width=0.49\textwidth]{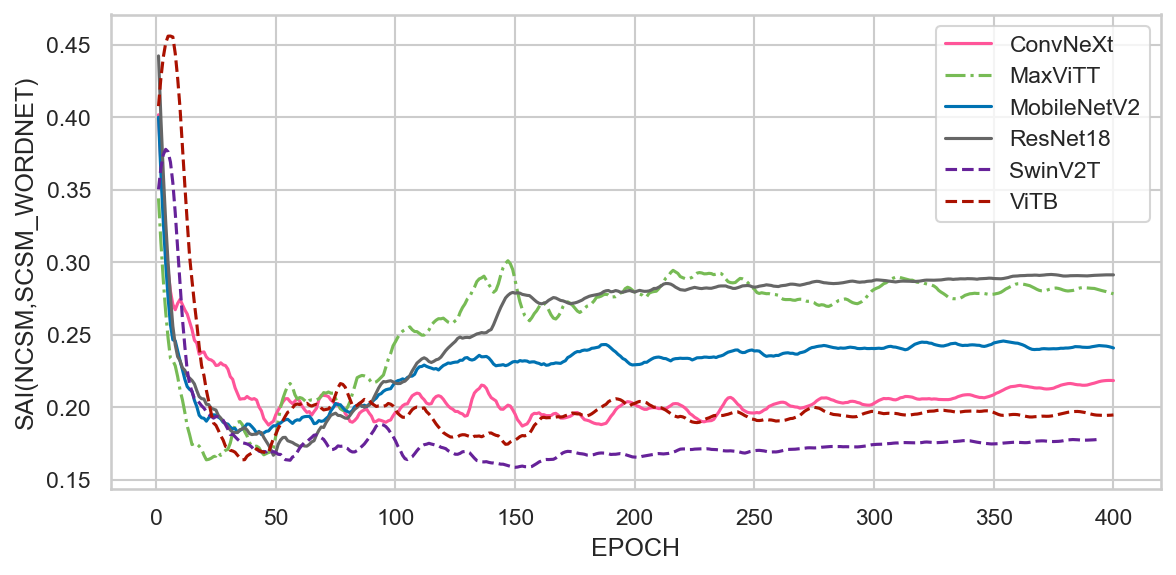}}
\end{center}
   \caption{CIFAR100: Similarity Alignment Index Curves for Network and WordNet similarity perception - \textbf{SAI(NCSM, SCSM)} for all possible similarity/distance measures. Both measures show that networks quickly develop a similarity perception that largely aligns with semantic relations. Excluding some minor drops, this alignment persists as training continues. }
\label{fig:sai_ncsm_scsm_cifar}
\end{figure}

The results of different variants of Similarity Alignment Index Curves for Network and WordNet similarity perception - \textbf{SAI(NCSM, SCSM)} presented in Fig. \ref{fig:sai_ncsm_scsm_cifar} are also in line with those obtained for Mini-ImageNet. The alignment grows quickly in the first app. 25 epochs, to later drop slightly and stabilize. The drop is the most visible for MaxViTT and structural similarity/distance measures.

\begin{figure}
\centering
\includegraphics[width=0.85\textwidth]{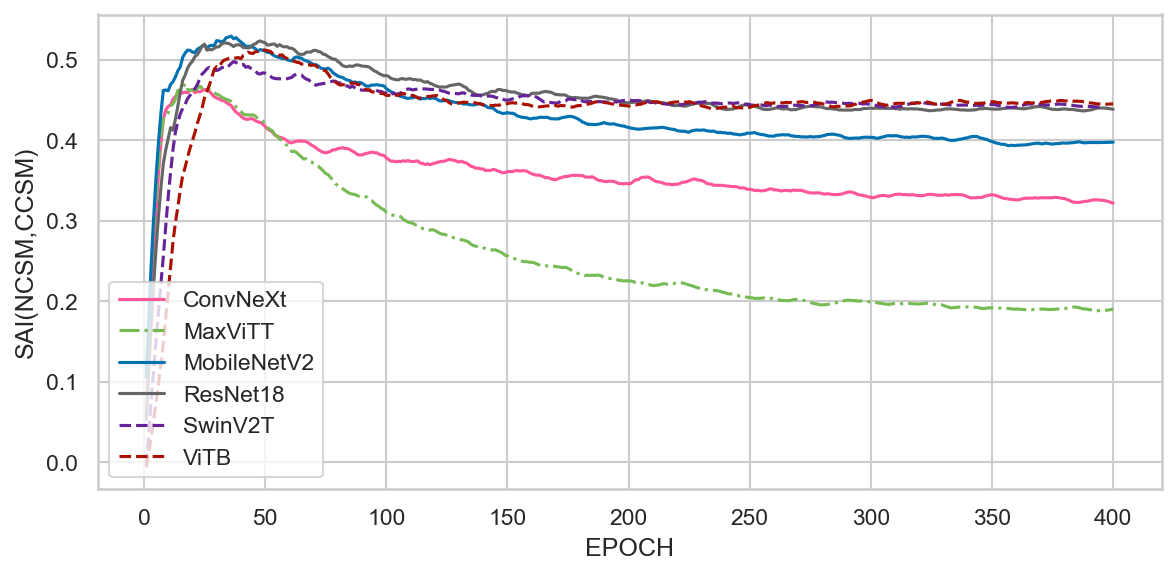}
\caption{CIFAR100: Similarity Alignment Index Curve between the Confusion-based similarity and the Network-based similarity - \textbf{SAI(NCSM, CCSM)}. The index rapidly grows in the very first epochs of training, reaches its maximum, then drops slightly with time. }
\label{fig:sai_ncsm_ccsm_cifar100}
\end{figure}

\begin{figure}
\centering
\includegraphics[width=0.85\textwidth]{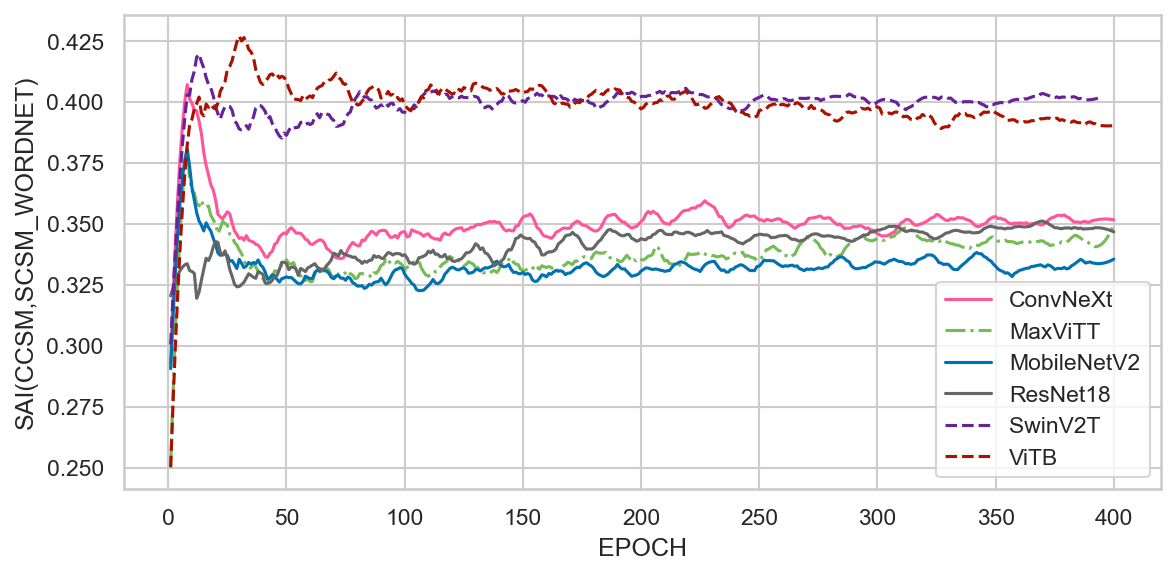}
\caption{CIFAR100: Similarity Alignment Index Curve for the Confusion-based similarity and the WordNet-based similarity - \textbf{SAI(CCSM, SCSM)}. The index rapidly grows in the very first epochs of training, reaches its maximum, then drops slightly with time and stabilizes.}
\label{fig:sai_ccsm_scsm_cifar100}
\end{figure}

\begin{figure}
\centering
\subfloat[][ResNet18 (1)]{\includegraphics[width=0.23\textwidth]{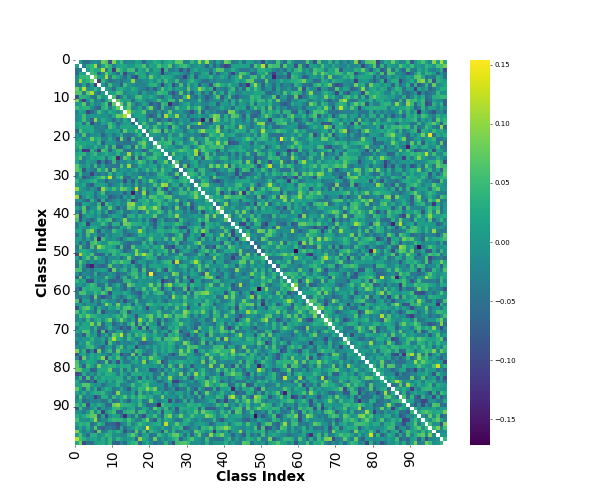}}
\subfloat[][ResNet18 (5)]{\includegraphics[width=0.23\textwidth]{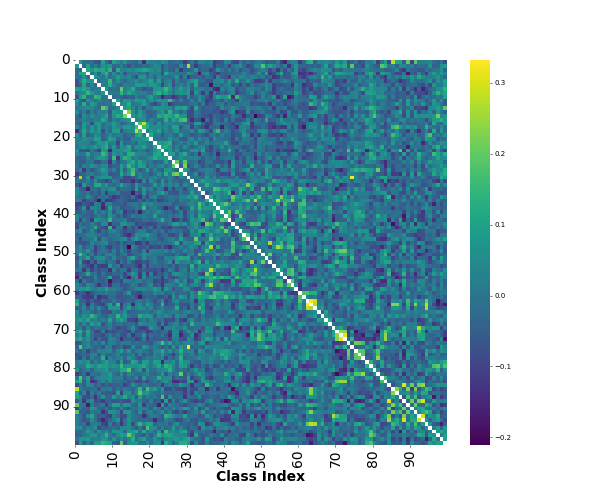}}
\subfloat[][ResNet18 (25)]{\includegraphics[width=0.23\textwidth]{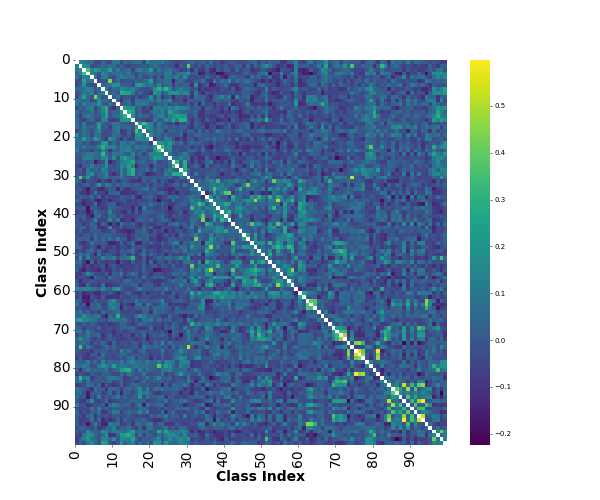}}
\subfloat[][ResNet18 (200)]{\includegraphics[width=0.23\textwidth]{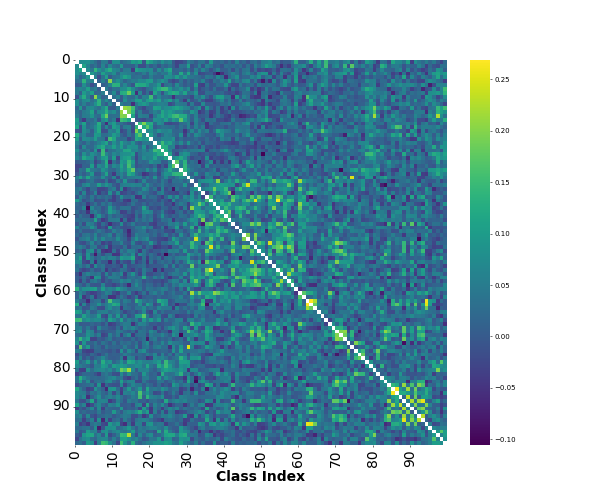}}
\\[-1em]
\subfloat[][SwinV2 (1)]{\includegraphics[width=0.23\textwidth]{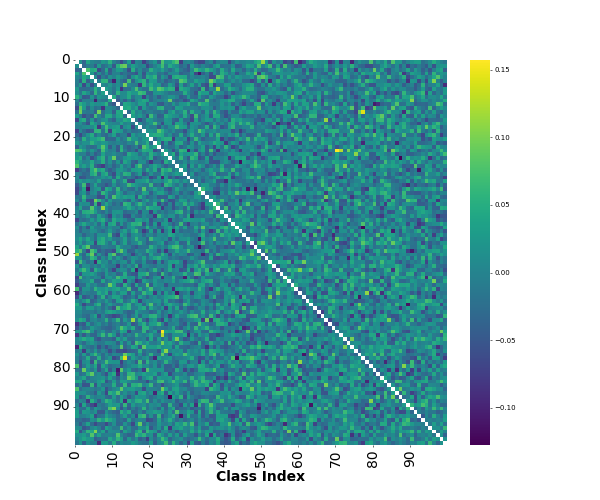}}
\subfloat[][SwinV2 (5)]{\includegraphics[width=0.23\textwidth]{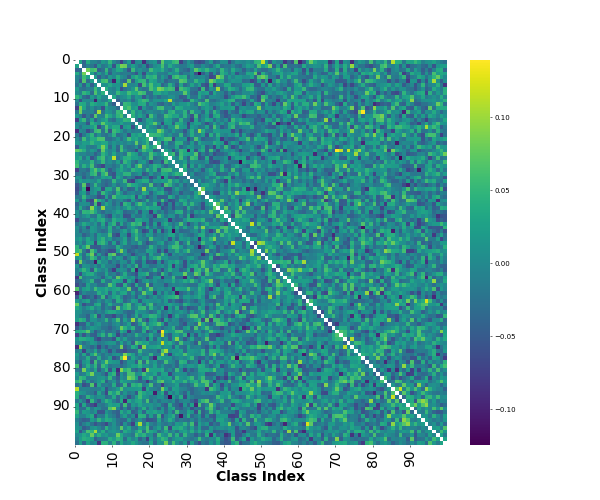}}
\subfloat[][SwinV2 (25)]{\includegraphics[width=0.23\textwidth]{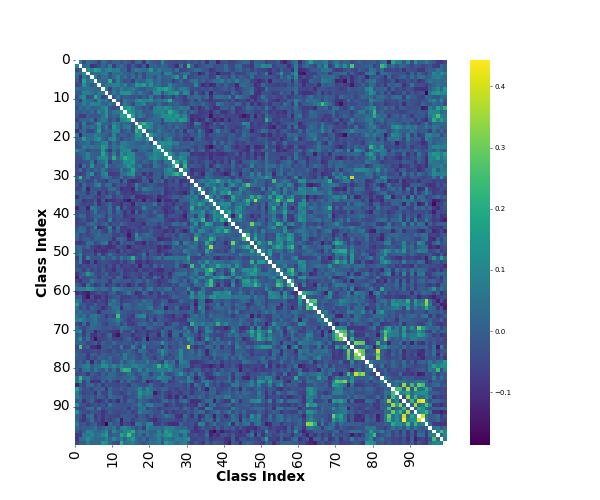}}
\subfloat[][SwinV2 (200)]{\includegraphics[width=0.23\textwidth]{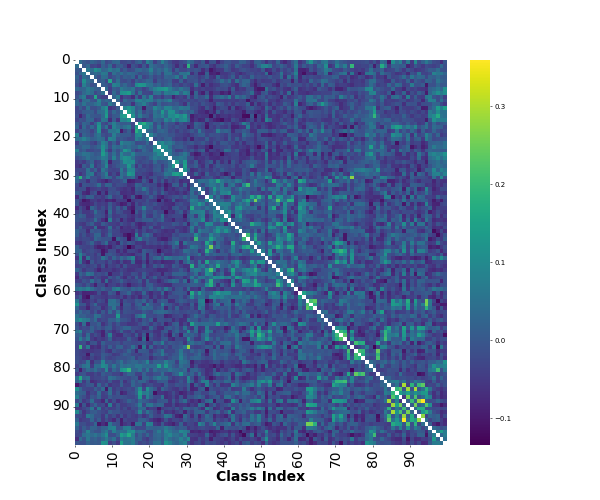}}
\\[-1em]
\subfloat[][ConvNeXt (1)]{\includegraphics[width=0.23\textwidth]{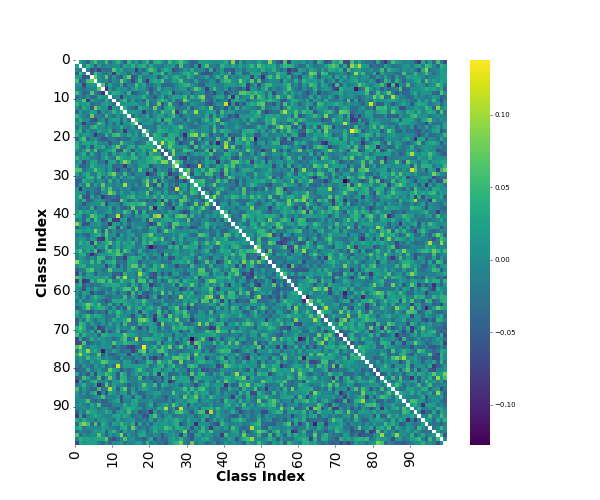}}
\subfloat[][ConvNeXt (5)]{\includegraphics[width=0.23\textwidth]{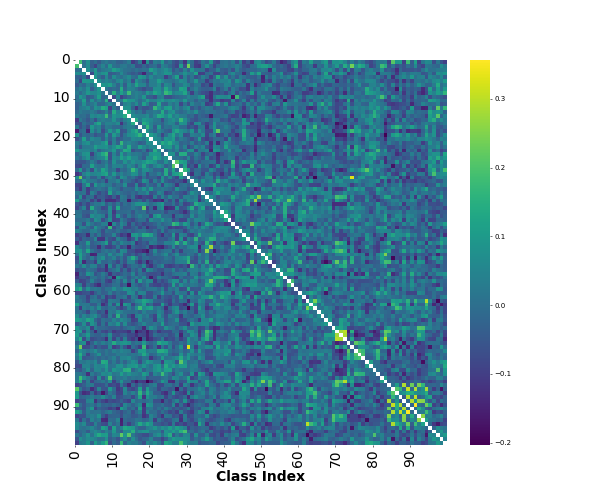}}
\subfloat[][ConvNeXt (25)]{\includegraphics[width=0.23\textwidth]{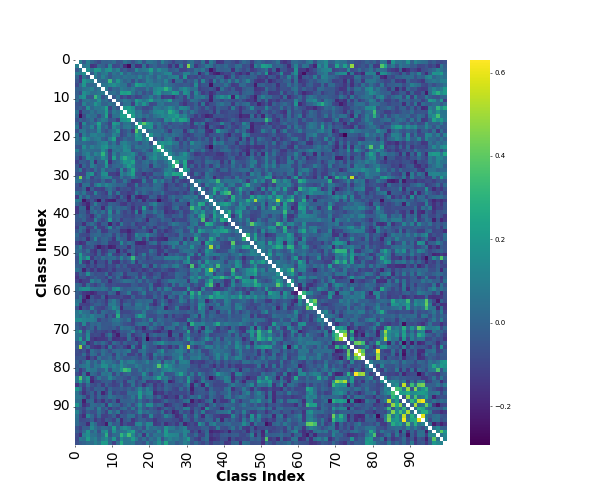}}
\subfloat[][ConvNeXt (200)]{\includegraphics[width=0.23\textwidth]{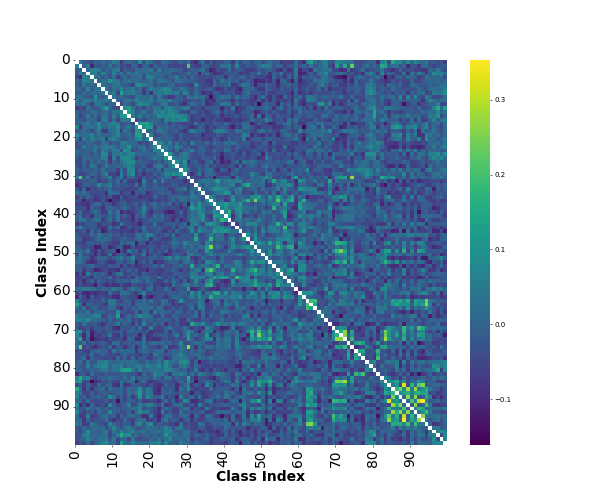}}
\\[-1em]
\subfloat[][ViTB (1)]{\includegraphics[width=0.23\textwidth]{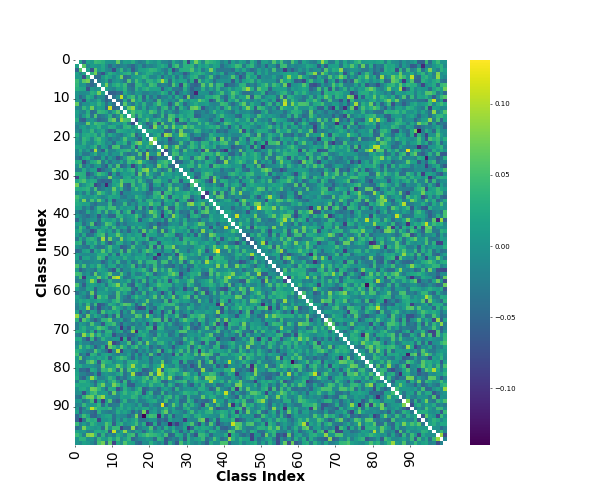}}
\subfloat[][ViTB (5)]{\includegraphics[width=0.23\textwidth]{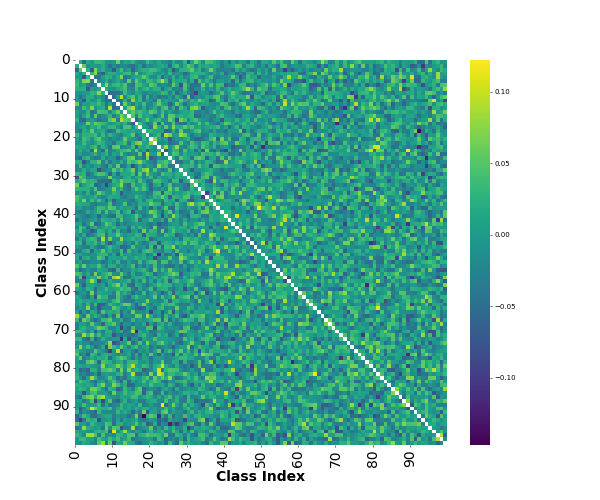}}
\subfloat[][ViTB (25)]{\includegraphics[width=0.23\textwidth]{FIGS/CIFAR100_matrices/ViTB4.png}}
\subfloat[][ViTB (200)]{\includegraphics[width=0.23\textwidth]{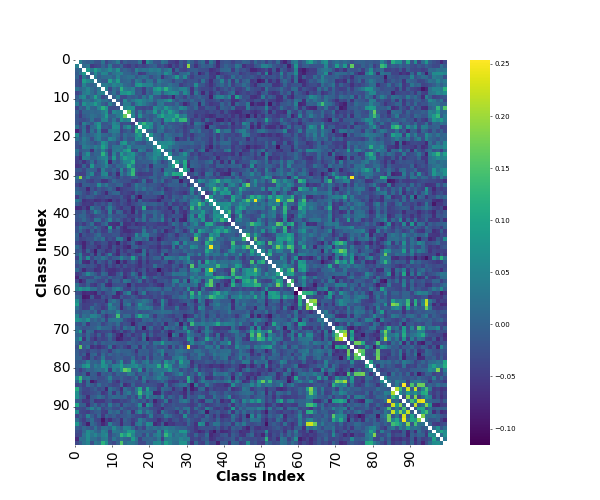}}
\\[-1em]
\subfloat[][MobileNetV2 (1)]{\includegraphics[width=0.23\textwidth]{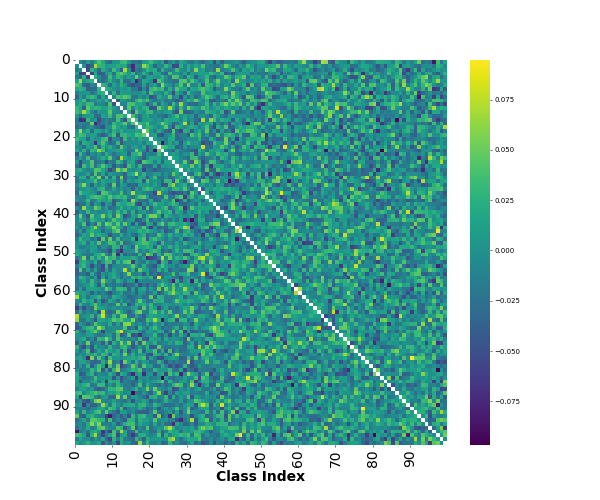}}
\subfloat[][MobileNetV2 (5)]{\includegraphics[width=0.23\textwidth]{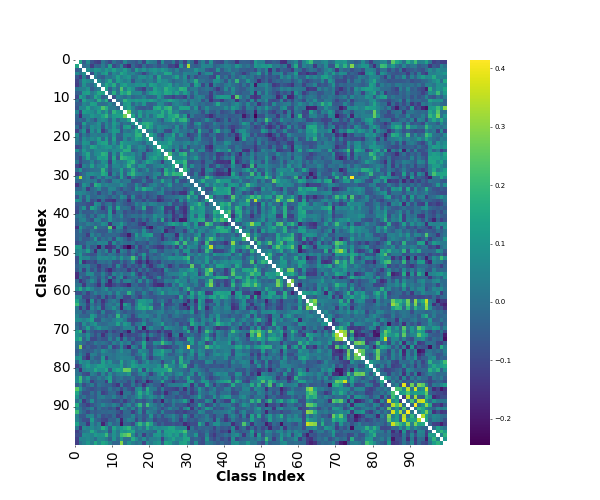}}
\subfloat[][MobileNetV2 (25)]{\includegraphics[width=0.23\textwidth]{FIGS/CIFAR100_matrices/MobileNetV24.png}}
\subfloat[][MobileNetV2 (200)]{\includegraphics[width=0.23\textwidth]{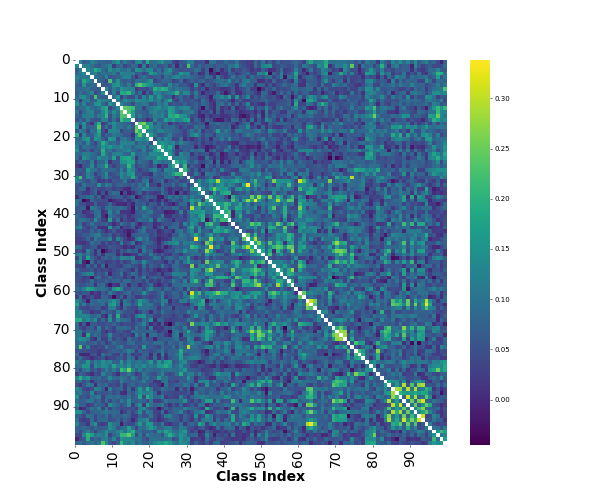}}
\\[-1em]
\subfloat[][MaxViTT (1)]{\includegraphics[width=0.23\textwidth]{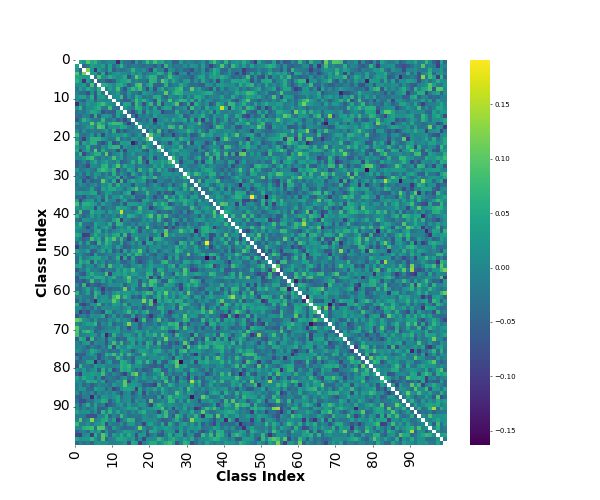}}
\subfloat[][MaxViTT (5)]{\includegraphics[width=0.23\textwidth]{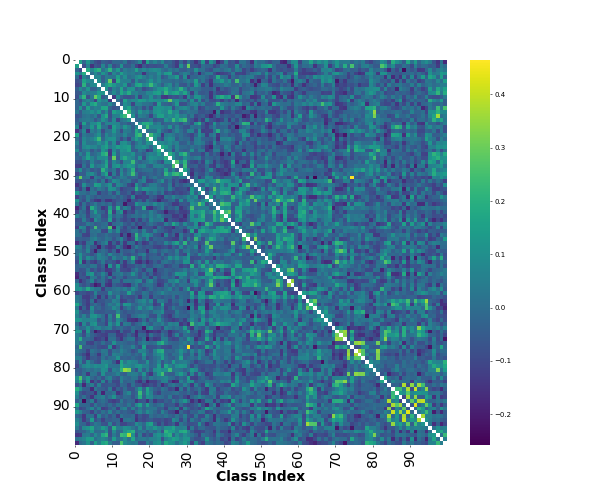}}
\subfloat[][MaxViTT (25)]{\includegraphics[width=0.23\textwidth]{FIGS/CIFAR100_matrices/MaxViTT4.png}}
\subfloat[][MaxViTT (200)]{\includegraphics[width=0.23\textwidth]{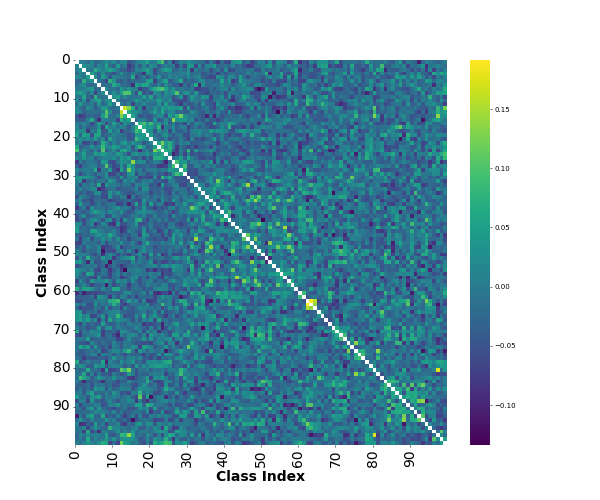}}
\caption{CIFAR100: NCSMs of all models for at different epochs (number - in brackets). }
\label{fig:ncsm_cifar100}
\end{figure}

The qualitative analysis with Network Class Similarity Matrices presented in Fig. \ref{fig:ncsm_cifar100} also shows similar results to the ones obtained for Mini-ImageNet. The examined CNNs quicker reveal a clear hierarchical similarity structure than ViTs (also at the very beginning of training - after app. 5 epochs).

\begin{figure}
\begin{center}
\subfloat[][All]{\includegraphics[width=0.5\textwidth]{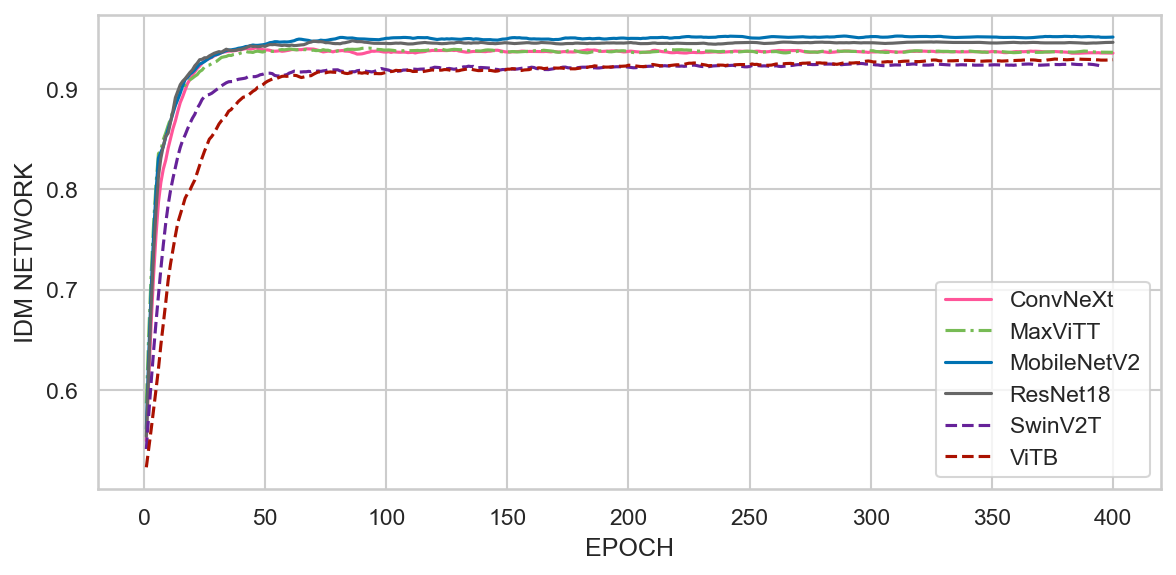}}
\subfloat[][Errors only]{\includegraphics[width=0.5\textwidth]{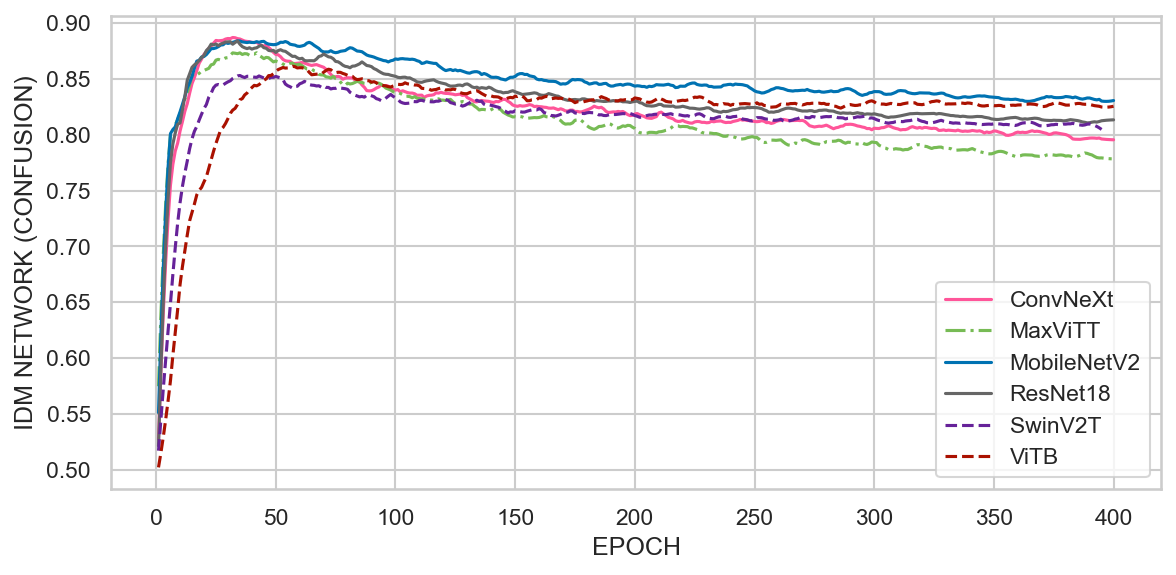}}
\end{center}
   \caption{CIFAR100: Network-based IDM. The plots show that all networks quickly start to make mistakes between categories they perceive similar. After the initial gains, IDM stabilizes. Surprisingly, the errors only variant shows that with time, the networks start to make mistakes that are perceived as less similar (balanced by the increasing accuracy in the basic variant).}
\label{fig:idm_cifar100}
\end{figure}

\begin{figure}
\begin{center}
\subfloat[][All]{\includegraphics[width=0.5\textwidth]{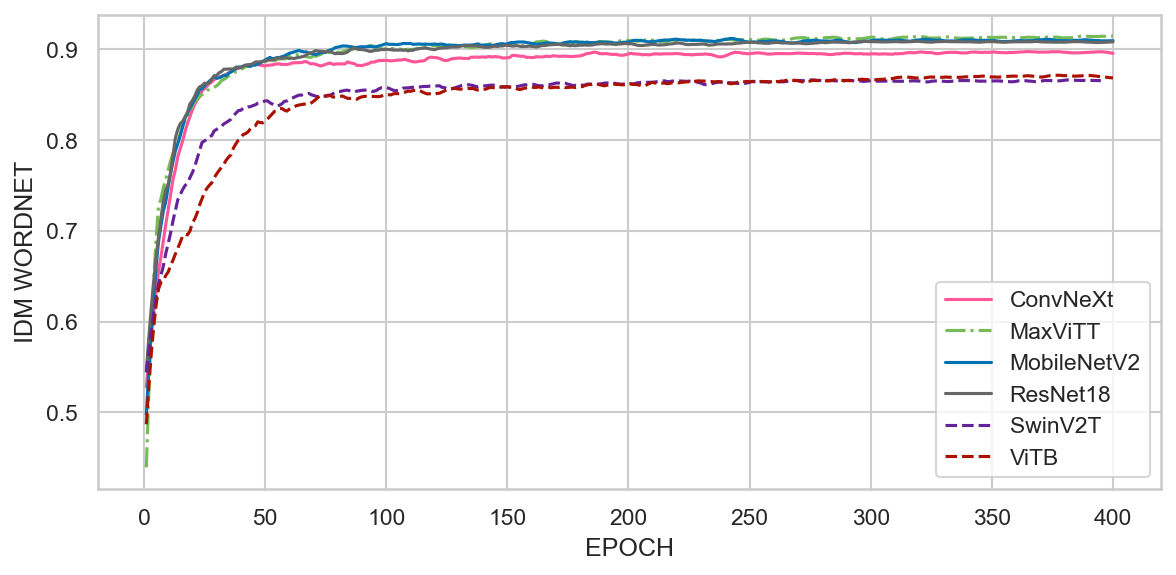}}
\subfloat[][Errors only]{\includegraphics[width=0.5\textwidth]{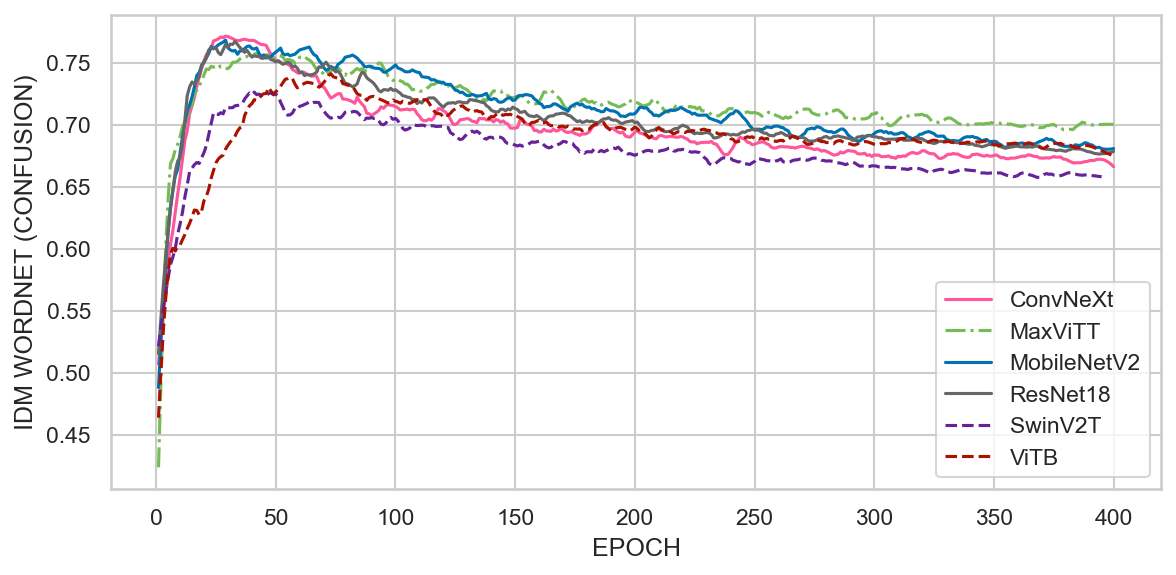}}
\end{center}
   \caption{CIFAR100: WordNet-based IDM. The plots show that all networks quickly start to make semantically-related mistakes. After the initial gains, IDM stabilizes. The errors only variant shows that with time, the networks start to make mistakes that are more distant in the WordNet hierarchy (balanced by the increasing accuracy in the basic variant).  }
\label{fig:idm_wordnet_cifar100}
\end{figure}

\subsection{Do the confusion patterns of CNNs and ViTs match their similarity perception throughout the training?}

Also in this case, the results for CIFAR100 regarding the alignment of the direct and indirect similarity perception of networks confirms the observations from the main body of the paper obtained for Mini-ImageNet. All the examined networks behave practically the same as for Mini-ImageNet with slightly higher values of the Similarity Alignment Index Curve between the Confusion-based similarity and the Network-based similarity - \textbf{SAI(NCSM, CCSM)} (see Fig. \ref{fig:sai_ncsm_ccsm_cifar100}). The hybrid model and the ViT-inspired CNN, again, can be easily distinguished from other models with their very similar to each other behavior. Similarly, the Network-based IDM Curves presented in Fig. \ref{fig:idm_cifar100} show a very close behavior to the curves obtained for Mini-ImageNet. The small difference is that in the case of CIFAR100, the examined ViTB and SwinV2T models obtained slightly lower results for the basic variant than CNNs. Also in the case of CIFAR100, it is visible that CNNs faster align their confusions with their perception of similarity than ViTs. This observation is also supported by the qualitative results based on Confusion-based Class Similarity Matrices (CCSMs) presented in Fig. \ref{fig:ccsm_cifar100}. While for the ResNet18 and ConvNeXt models, the indirect similarity perception is revealed as soon as after app. 25 epochs, more time is needed for ViTs used in the experiments. Again, it can be observed that the indirect similarity measurements via confusion matrices results in much less dense similarity matrices with only an approximate structure of similarity perception.

\subsection{Do the confusion patterns of CNNs and ViTs align with  semantic similarity throughout the training?}

\begin{figure}
\centering
\subfloat[][ResNet18 (1)]{\includegraphics[width=0.23\textwidth]{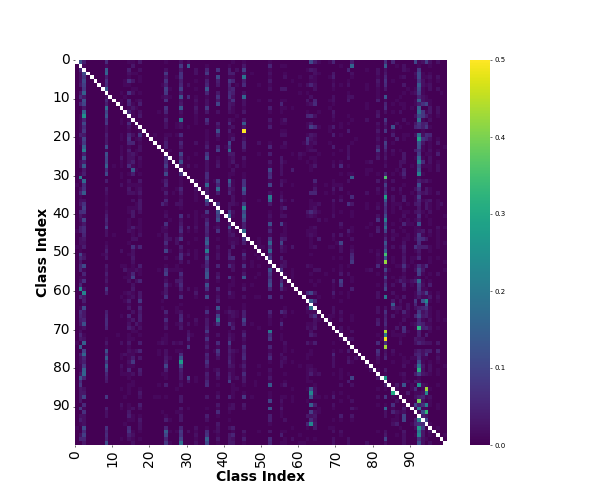}}
\subfloat[][ResNet18 (5)]{\includegraphics[width=0.23\textwidth]{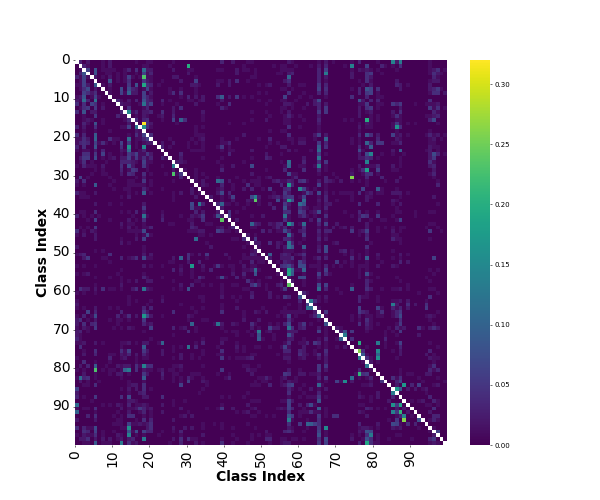}}
\subfloat[][ResNet18 (25)]{\includegraphics[width=0.23\textwidth]{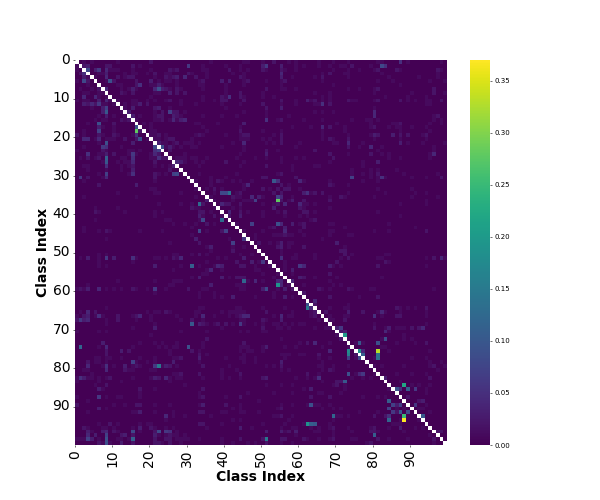}}
\subfloat[][ResNet18 (200)]{\includegraphics[width=0.23\textwidth]{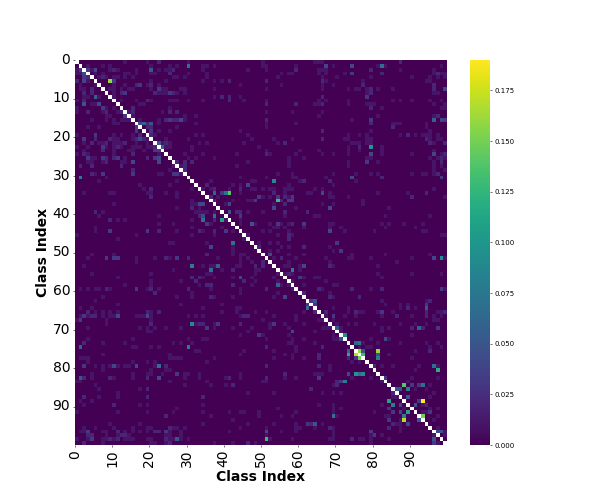}}
\\[-1em]
\subfloat[][SwinV2 (1)]{\includegraphics[width=0.23\textwidth]{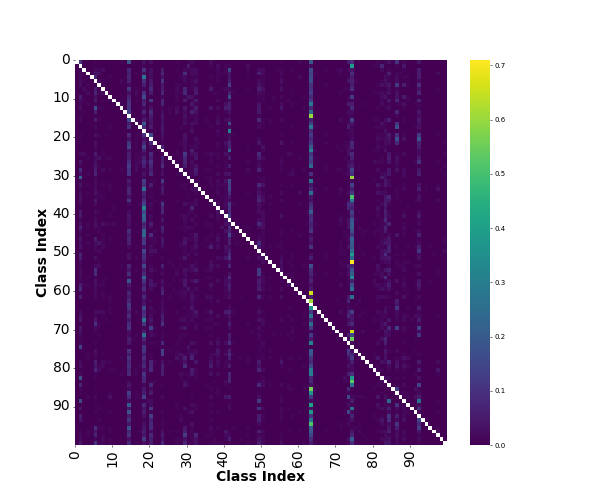}}
\subfloat[][SwinV2 (5)]{\includegraphics[width=0.23\textwidth]{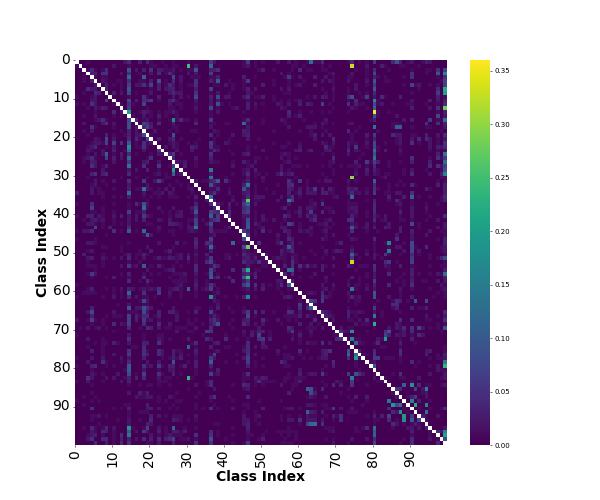}}
\subfloat[][SwinV2 (25)]{\includegraphics[width=0.23\textwidth]{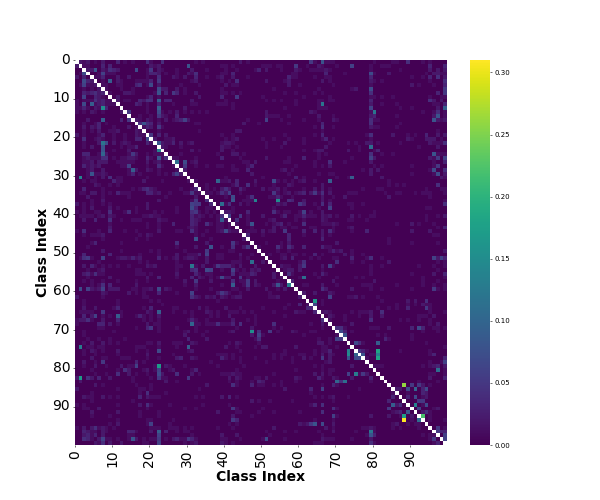}}
\subfloat[][SwinV2 (200)]{\includegraphics[width=0.23\textwidth]{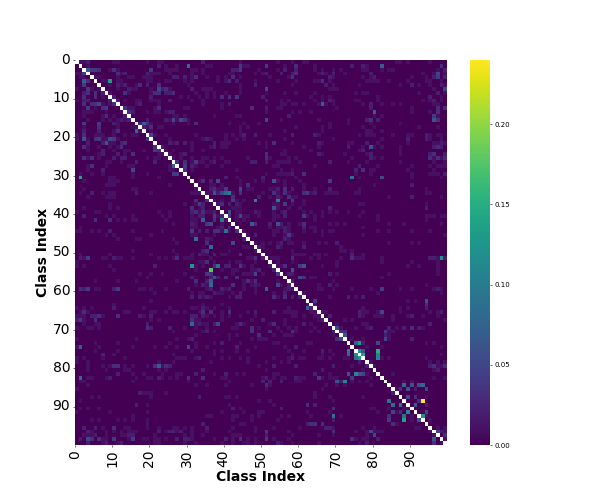}}
\\[-1em]
\subfloat[][ConvNeXt (1)]{\includegraphics[width=0.23\textwidth]{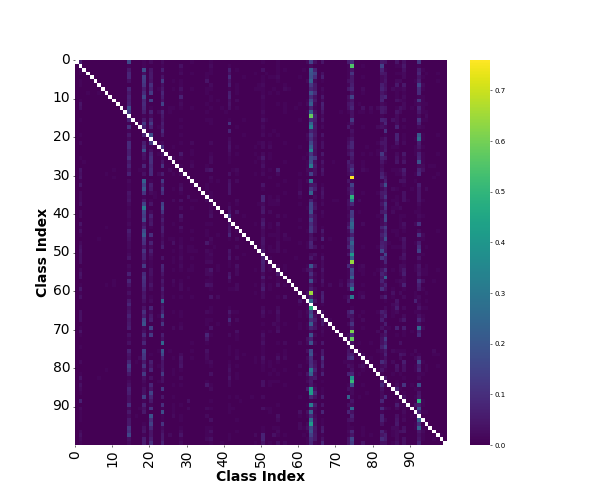}}
\subfloat[][ConvNeXt (5)]{\includegraphics[width=0.23\textwidth]{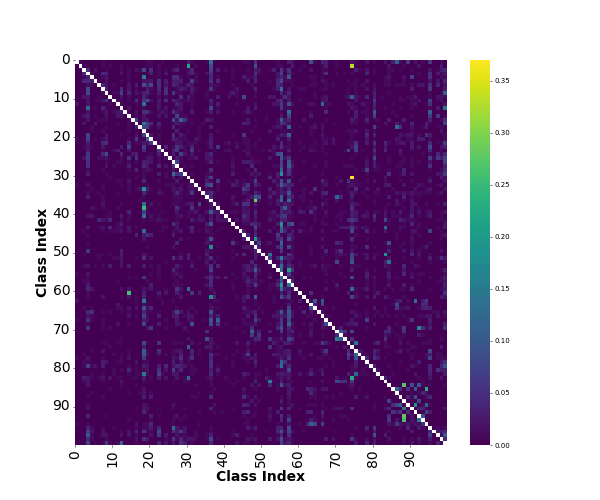}}
\subfloat[][ConvNeXt (25)]{\includegraphics[width=0.23\textwidth]{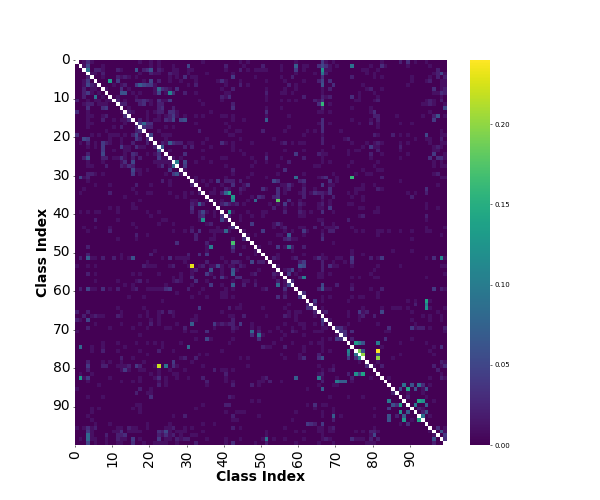}}
\subfloat[][ConvNeXt (200)]{\includegraphics[width=0.23\textwidth]{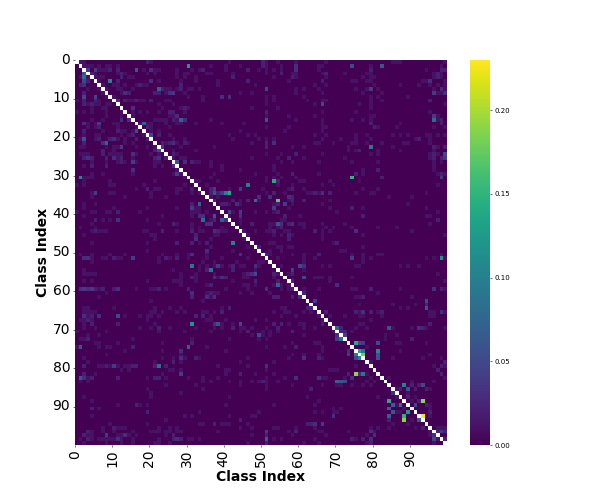}}
\\[-1em]
\subfloat[][ViTB (1)]{\includegraphics[width=0.23\textwidth]{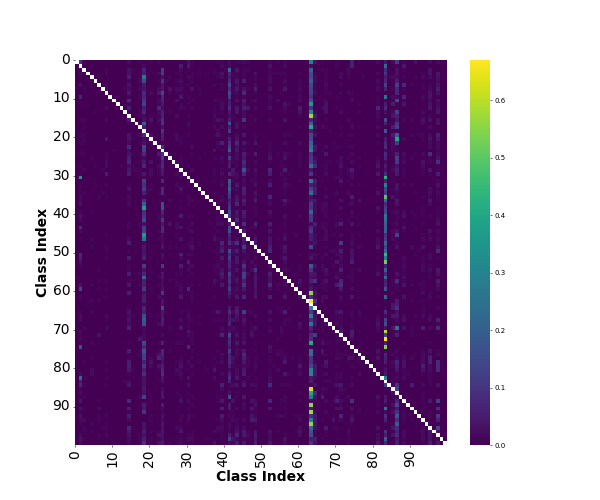}}
\subfloat[][ViTB (5)]{\includegraphics[width=0.23\textwidth]{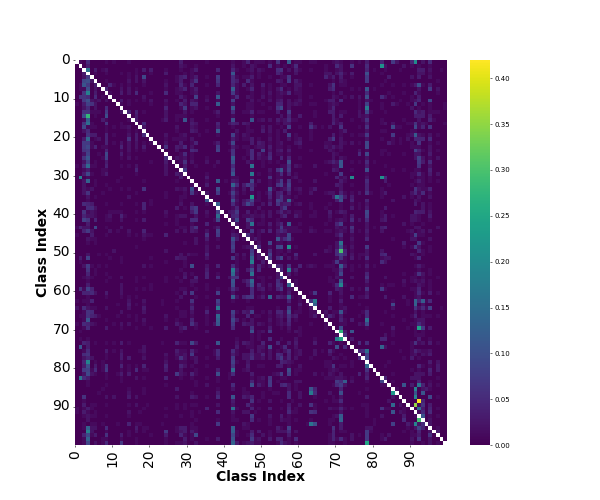}}
\subfloat[][ViTB (25)]{\includegraphics[width=0.23\textwidth]{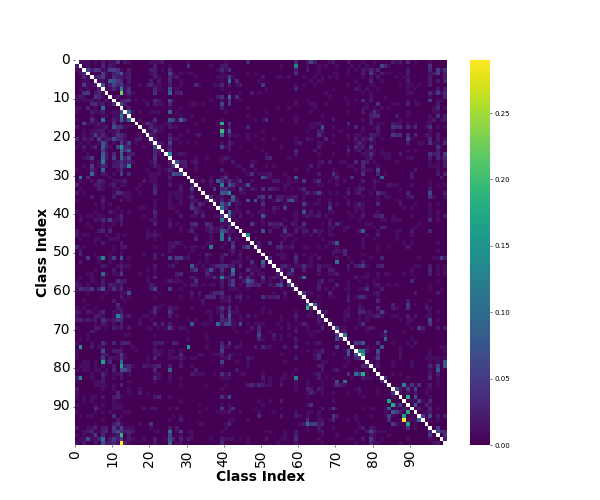}}
\subfloat[][ViTB (200)]{\includegraphics[width=0.23\textwidth]{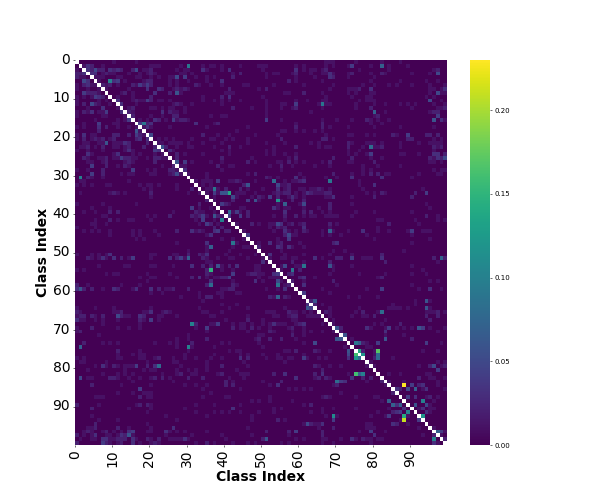}}
\\[-1em]
\subfloat[][MobileNetV2 (1)]{\includegraphics[width=0.23\textwidth]{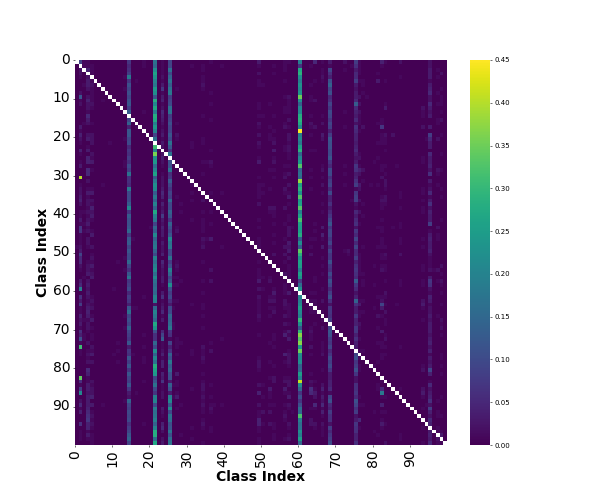}}
\subfloat[][MobileNetV2 (5)]{\includegraphics[width=0.23\textwidth]{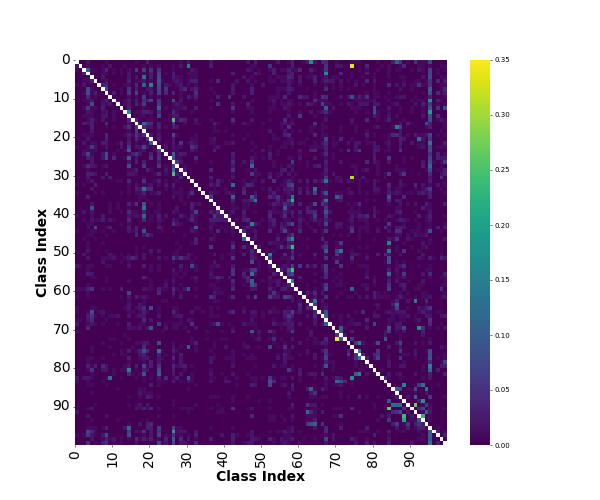}}
\subfloat[][MobileNetV2 (25)]{\includegraphics[width=0.23\textwidth]{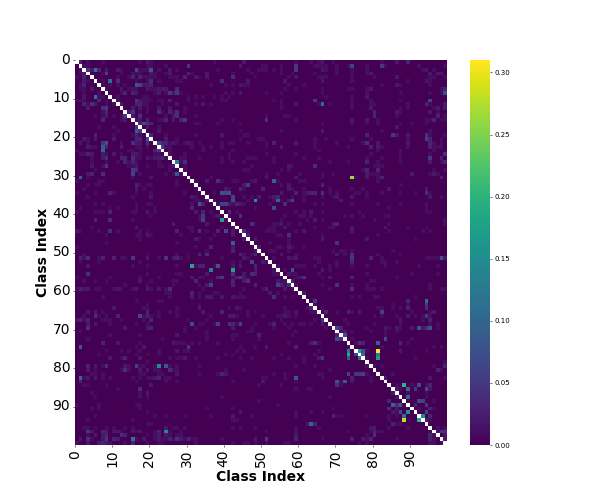}}
\subfloat[][MobileNetV2 (200)]{\includegraphics[width=0.23\textwidth]{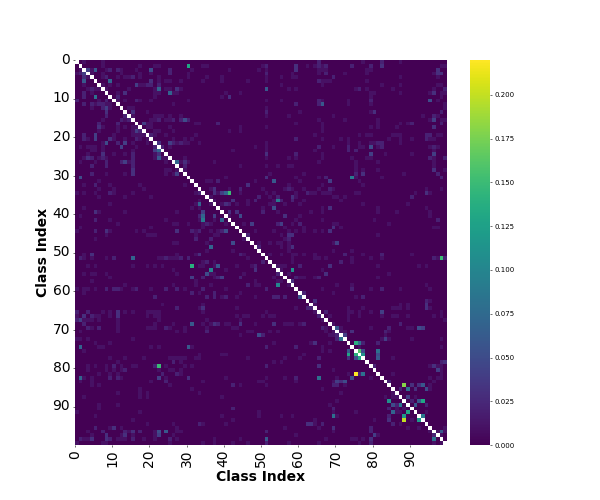}}
\\[-1em]
\subfloat[][MaxViTT (1)]{\includegraphics[width=0.23\textwidth]{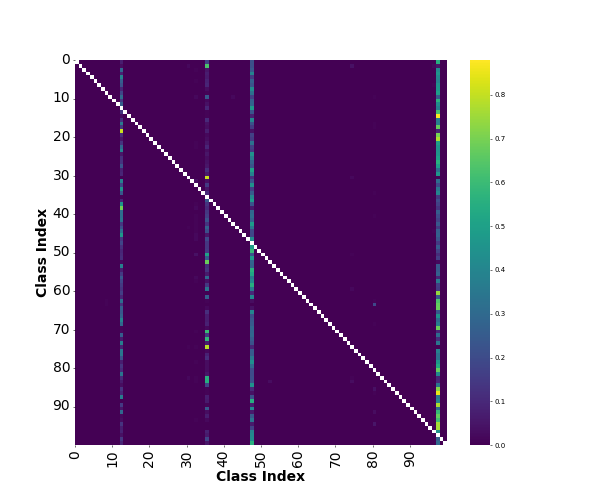}}
\subfloat[][MaxViTT (5)]{\includegraphics[width=0.23\textwidth]{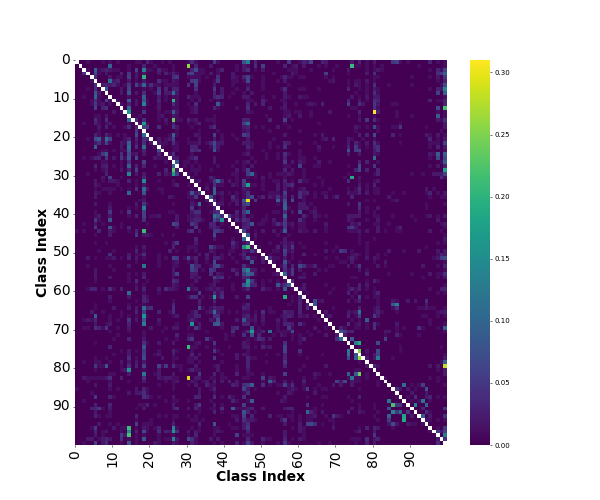}}
\subfloat[][MaxViTT (25)]{\includegraphics[width=0.23\textwidth]{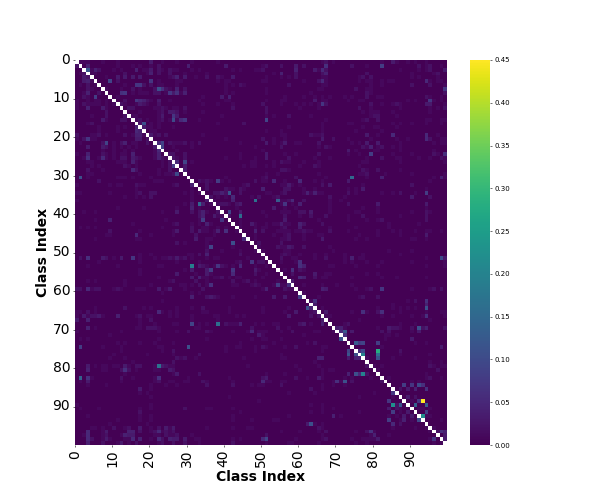}}
\subfloat[][MaxViTT (200)]{\includegraphics[width=0.23\textwidth]{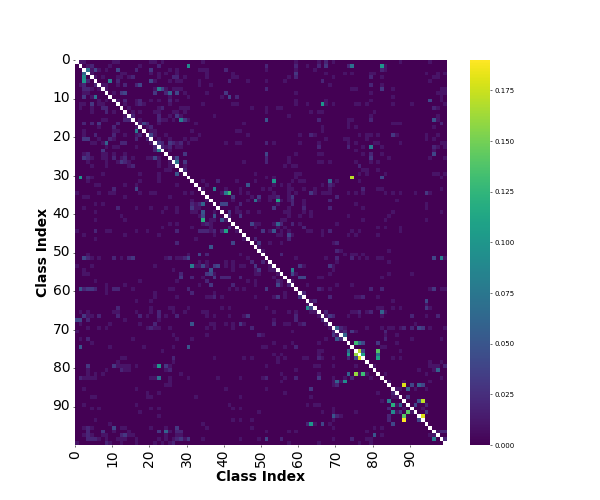}}
\caption{CIFAR100: Confusion-based similarity of all models at different epochs (number - in brackets).  }
\label{fig:ccsm_cifar100}
\end{figure}

The results obtained on CIFAR100 support our results from the main body of the paper. For this dataset, the indirect similarity patterns derived from the confusion matrices also partially align with semantic similarity. It is, first of all, visible via the visualization of the CCSM presented in Fig. \ref{fig:ccsm_cifar100} and similarity to WCSM (SCSM) obtained for CIFAR100 (especially of the last CCSMs obtained for the 200th epoch). In Fig. \ref{fig:sai_ccsm_scsm_cifar100}, we present the SAI Curve for the Confusion-based and the WordNet-based similarity - \textbf{SAI(CCSM, SCSM)}, which numerically supports this partial alignment. Again, the SAI(CCSM, SCSM) values are significantly lower than the SAI(NCSM, SCSM) values (direct, structural similarity assessment), which is caused by less dense structure of the CSM created based on the confusions. A rapid increase in the SAI is visible in the first training epochs, followed by similarity refinement and stabilization. This behavior is reflected in the plots of WordNet-based IDM in Fig. \ref{fig:idm_wordnet_cifar100}. The obtained values are slightly higher than the ones obtained for Mini-ImageNet, suggesting that for CIFAR100, networks make mistakes from a narrower semantic neighborhood of the ground truth classes.

\clearpage

\section{Does similarity perception emerge in ``bad'' net\-wor\-ks?}\label{appendix_bad}

\begin{figure}[h]
\begin{center}
\subfloat[][Testing accuracy - Mini-ImageNet]{\includegraphics[width=0.49\textwidth]{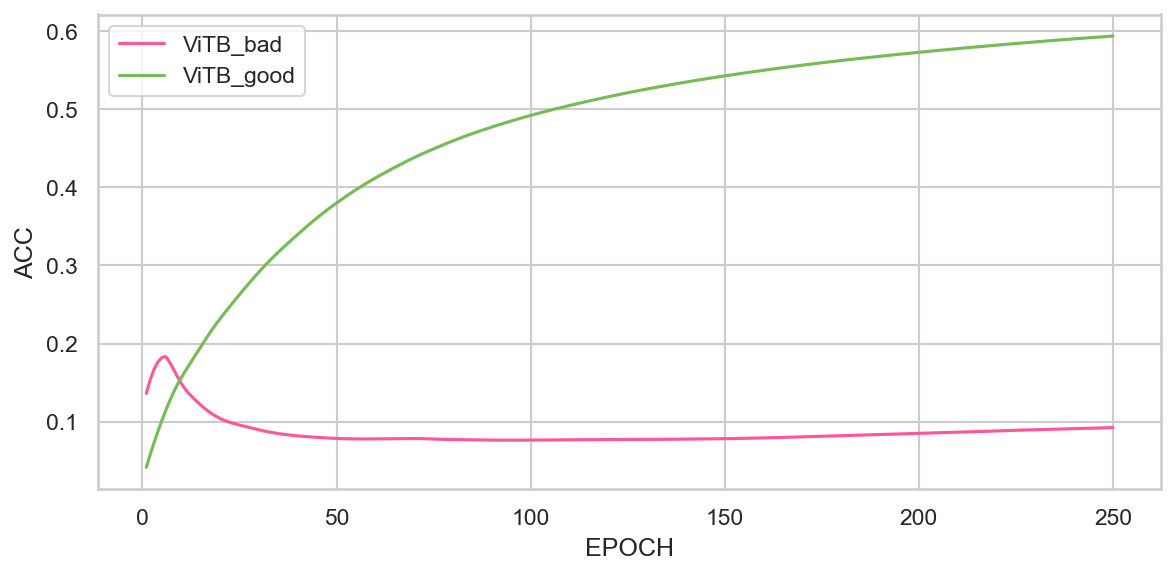}}
\subfloat[][Loss - Mini-ImageNet]{\includegraphics[width=0.49\textwidth]{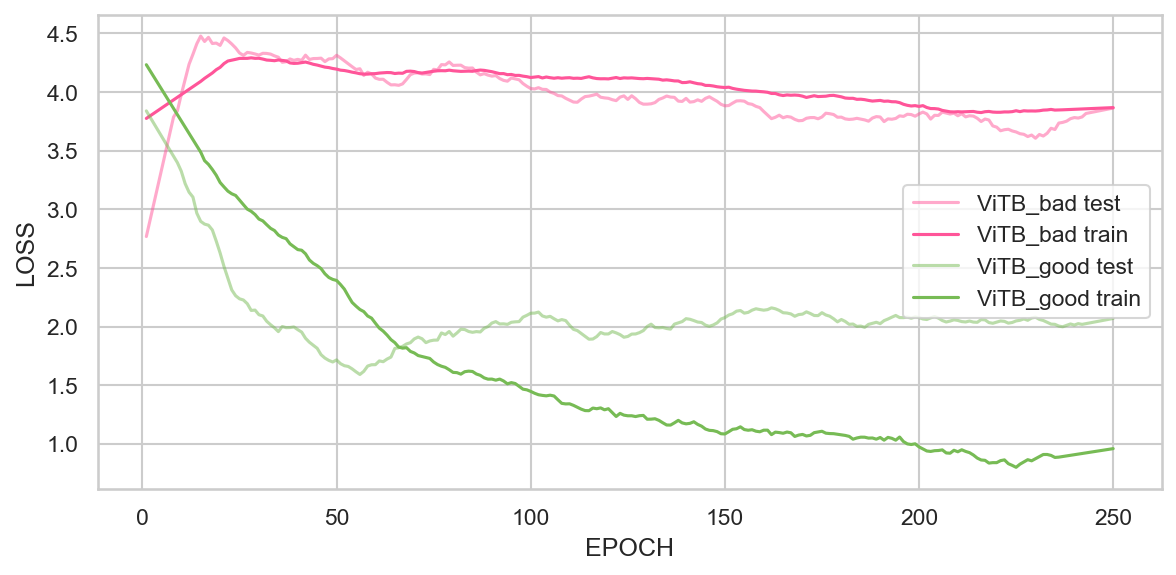}}
\end{center}
   \caption{Testing accuracy and Train/Test loss value curves for the Mini-ImageNet for a ``good'' and a ``bad'' ViT.}
\label{fig:accuracy_loss_badnet}
\end{figure}

We also decided to examine a case, in which network cannot reach an acceptable accuracy in a task that it is trained for. We name such a network a ``bad'' network, in contrast to ``good'' networks used in our experiments, that achieve relatively good accuracies. We choose the same ViT as the one used in our experiments but with significantly higher learning rate, due to which the network cannot learn effectively. It is our ``bad'' network. For a ``good'' network, we take the same ViT as in our original experiments. In Fig. \ref{fig:accuracy_loss_badnet}, one can notice that ``bad'' ViT achieves a very poor accuracy and almost no optimization is visible for it loss function.

\begin{figure}[h]
\begin{center}
\subfloat[][Mean Weights Similarity]{\includegraphics[width=0.49\textwidth]{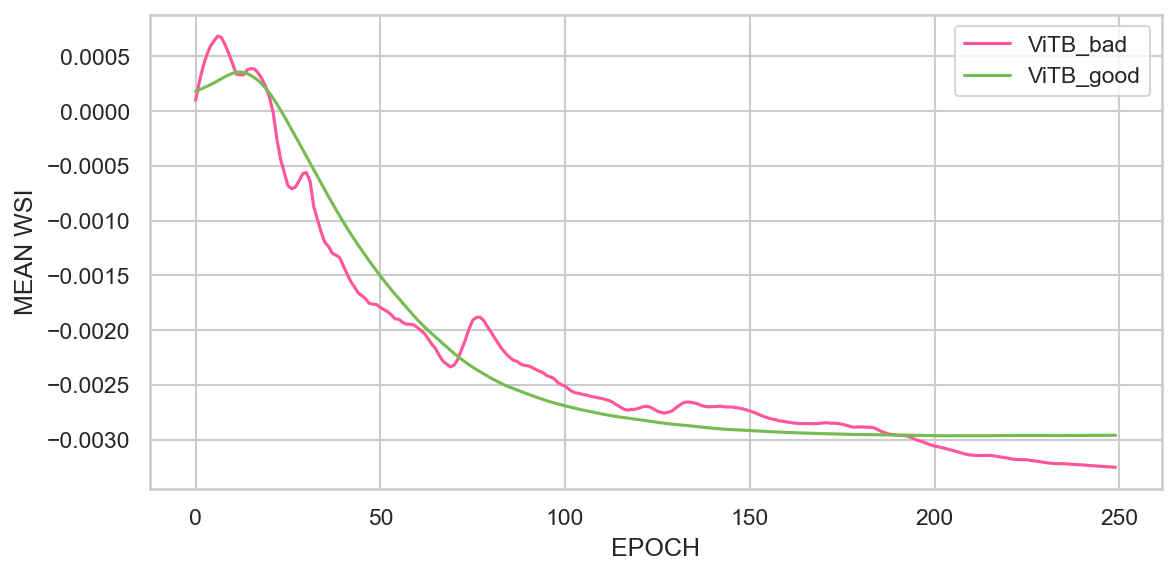}}
\subfloat[][Max Weights Similarity]{\includegraphics[width=0.49\textwidth]{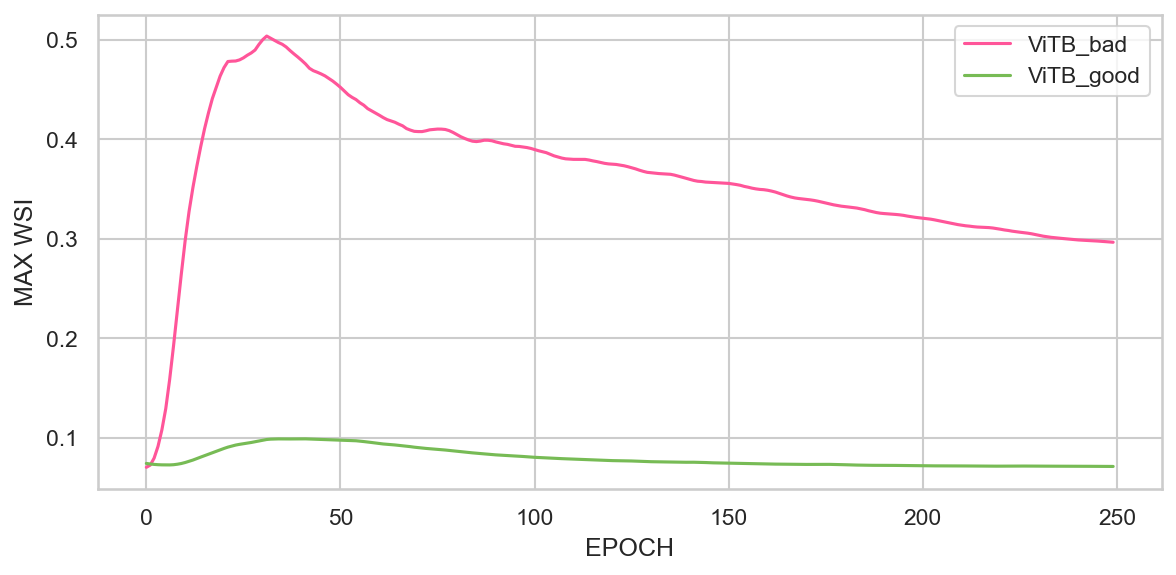}}
\\
\subfloat[][Min Weights Similarity]{\includegraphics[width=0.49\textwidth]{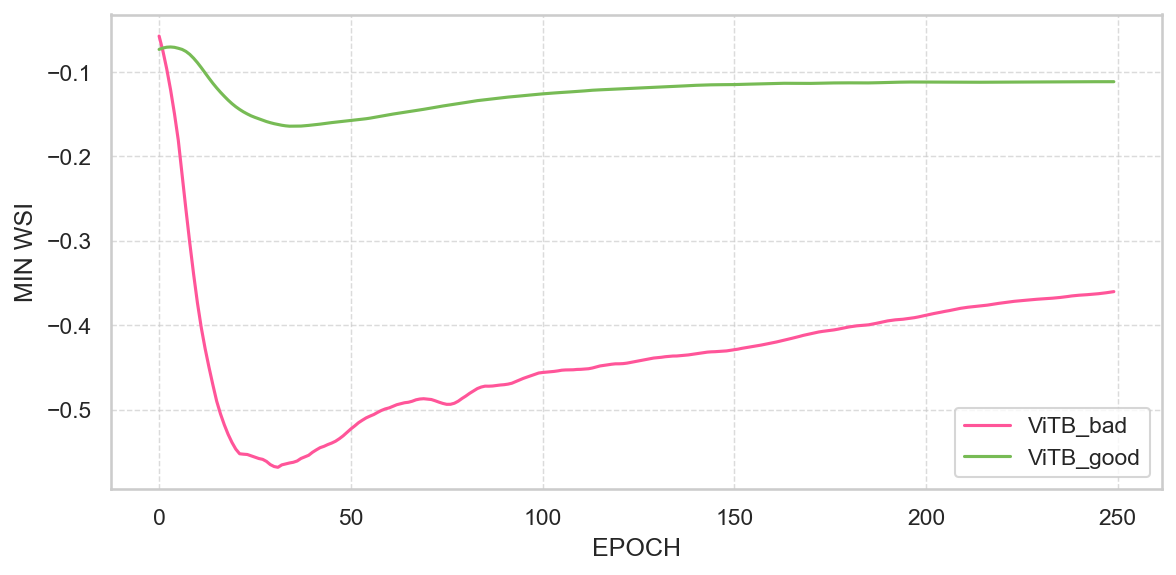}}
\end{center}
\caption{Mini-ImageNet: Weights Similarity Index (WSI) Curves for a ``good'' and a ``bad'' ViT.}
\label{fig:wsi_badnet}
\end{figure}

When it comes to the behaviour of weights and its analysis with different WSI variants (see Fig. \ref{fig:wsi_badnet}), surprisingly, the curves of a ``bad'' network are quite similar to the ones of a ``good'' network, especially the Mean variant. For the Mean variant, a main difference is that the plot is not that smooth as the one of a ``good'' one, but they mostly overlap. The Min and Max variants are similar in terms of the curves' shape, however they can be perceived as the "scaled" versions (they obtain significantly smaller/larger values respectively).

\begin{figure}[h]
\begin{center}
\subfloat[][SAI(NCSM, SCSM)]{\includegraphics[width=0.49\textwidth]{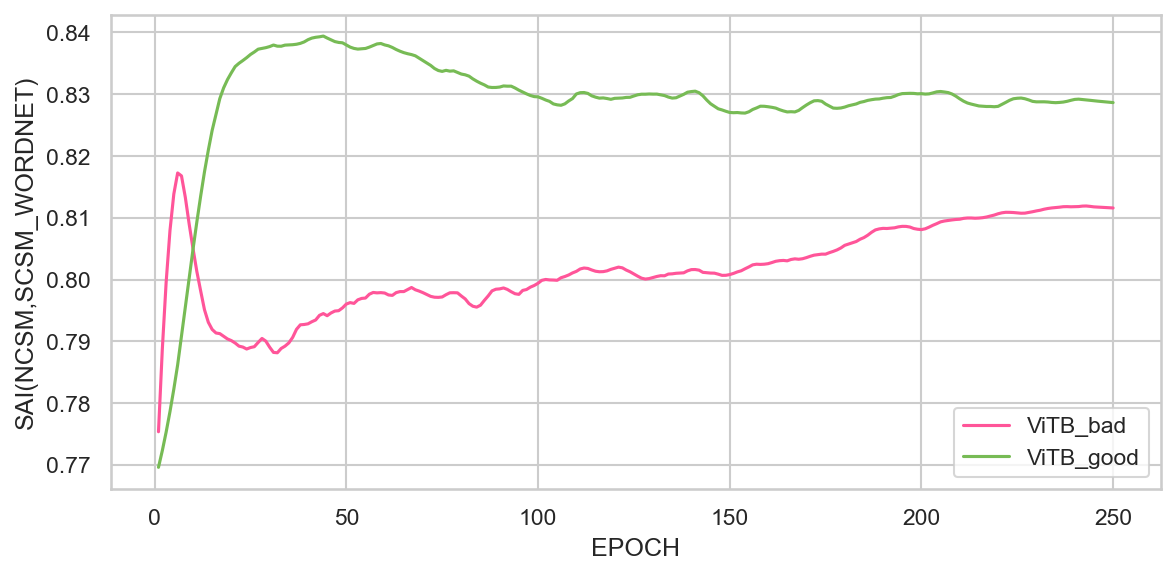}}
\subfloat[][SAI(NCSM, CCSM)]{\includegraphics[width=0.49\textwidth]{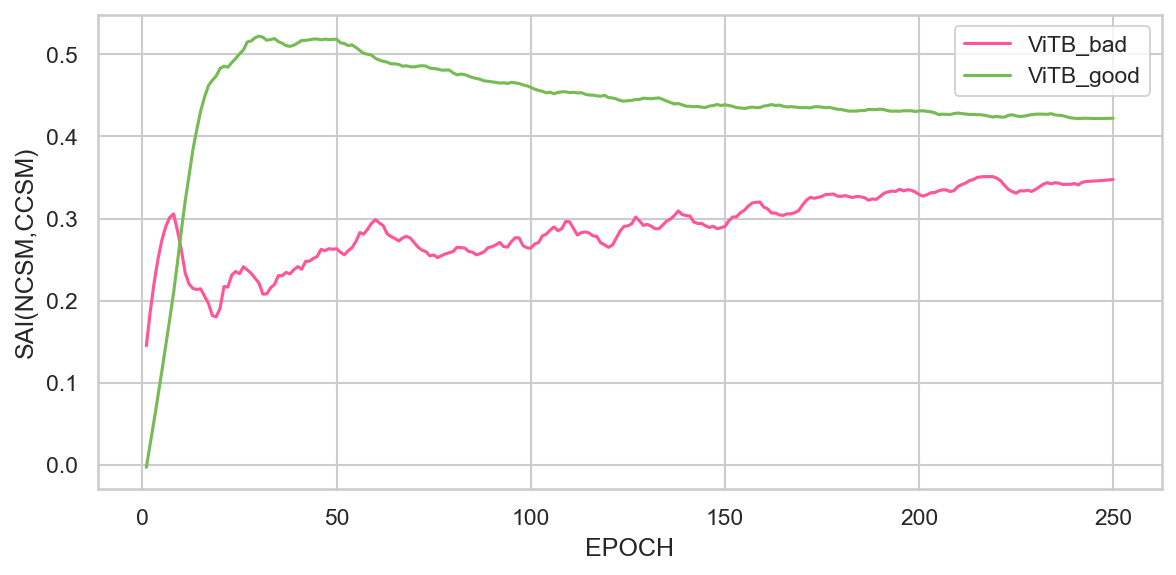}}
\\
\subfloat[][SAI(CCSM, SCSM)]{\includegraphics[width=0.49\textwidth]{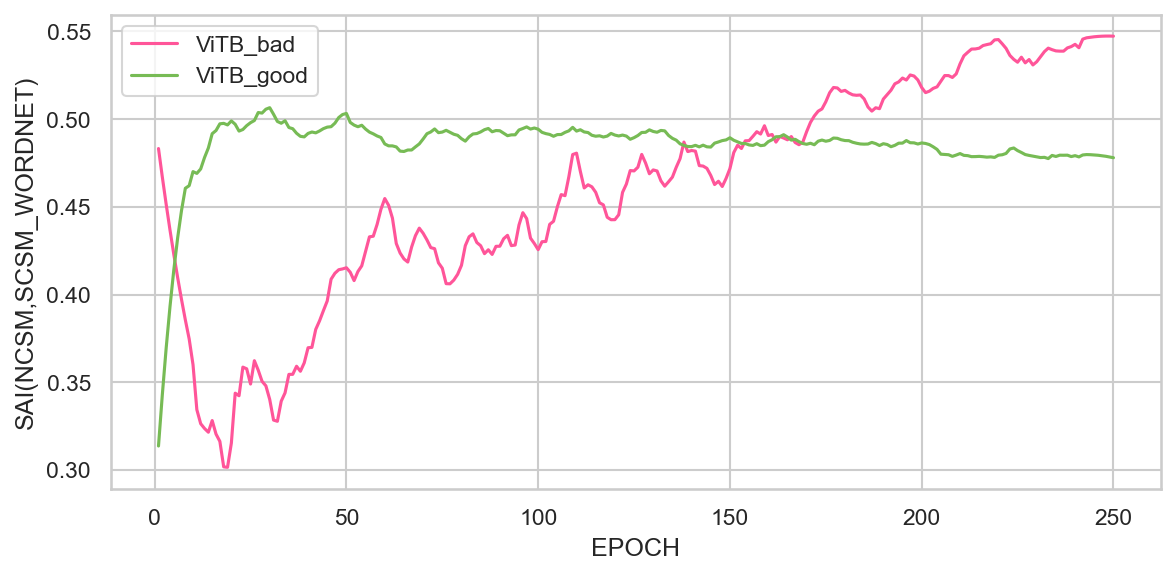}}
\end{center}
\caption{Mini-ImageNet: All SAI Curves for a ``good'' and a ``bad'' ViT.}
\label{fig:sai_badnet}
\end{figure}

In Fig. \ref{fig:sai_badnet}, we provide the SAI Curves obtained for all SAI variants for the ``good''-``bad'' network pair. It is visible, that all ``bad'' network's SAI curves behave significantly different than their ``good'' network counterparts. First of all, as expected, SAI(NCSM, SCSM) and SAI(NCSM, CCSM) obtain significantly lower values. Also, the shape of the curve is different. Instead of a harmonic (increase-refinement-stabilization) curves, one can notice a sudden increase (with lower maximum peak value than for a ``good'' network), a sudden drop and a slow increase until the end of training. This final phase shows that even "bad" networks tend to partially improve their similarity perception and incorporate it into their operation. It further suggests that the similarity emergence is due to a highly correlated structure of the real-world objects and that it is natural for a categorization system to discover these correlations via similarities. Surprisingly, for the SAI(NCSM, SCSM), values obtained at the end of the training are higher than the ones obtained for the ``good'' network. It may be caused, by the fact aforementioned in the main body of the paper, that in the later epochs, ``good'' networks make significantly less mistakes than ``bad'' networks, therefore their CCSMs are very sparse. That is why, to measure the relationships between CCSMs and other CSMs, it is good to also include in the analysis the exploration of different IDM variants (which focus on a more functional and local approach to errors), which we do in the next part of this section.

\begin{figure}[h]
\begin{center}
\subfloat[][Network]{\includegraphics[width=0.49\textwidth]{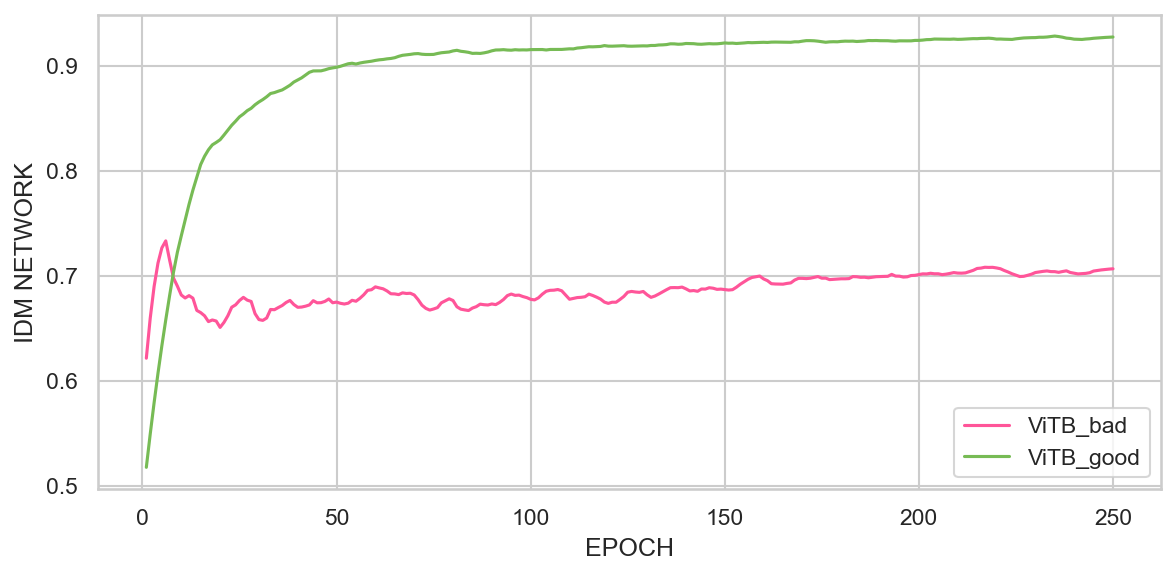}}
\subfloat[][Network, errors only]{\includegraphics[width=0.49\textwidth]{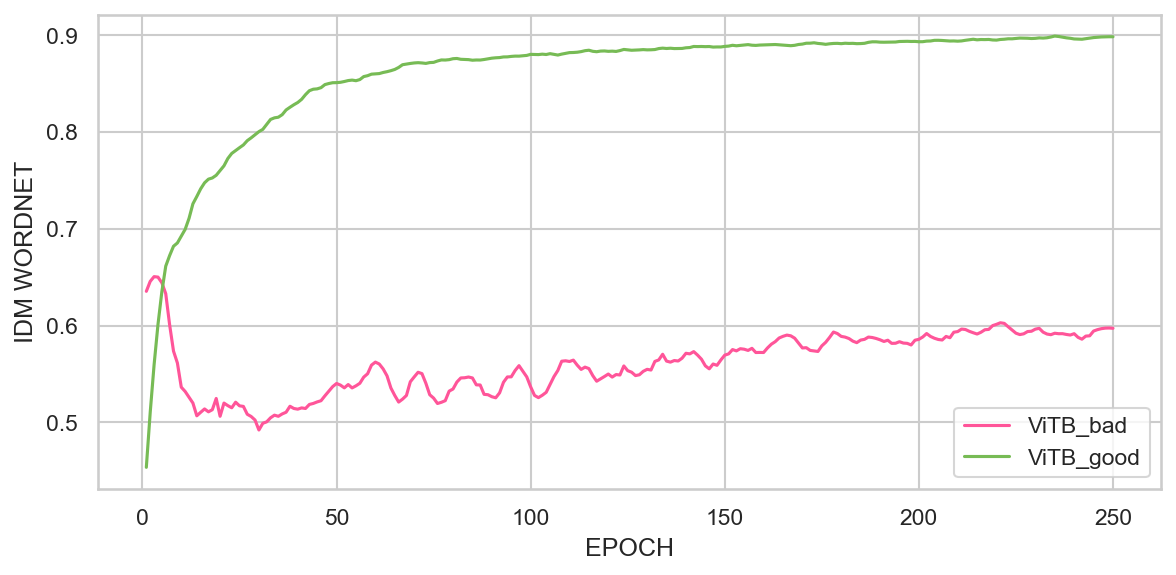}}
\\
\subfloat[][WordNet]{\includegraphics[width=0.49\textwidth]{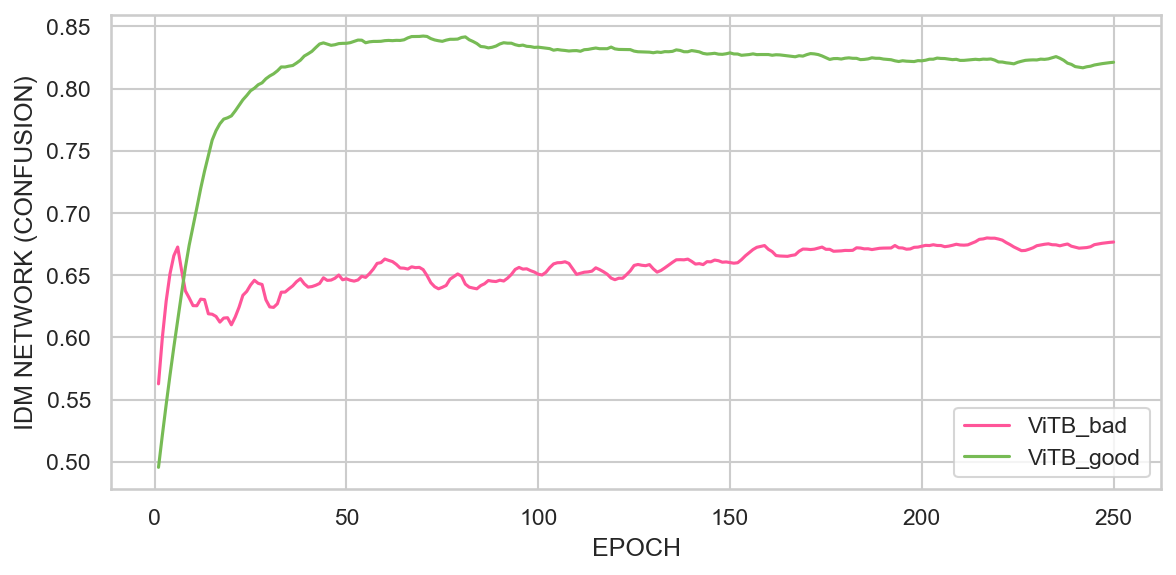}}
\subfloat[][WordNet, errors only]{\includegraphics[width=0.49\textwidth]{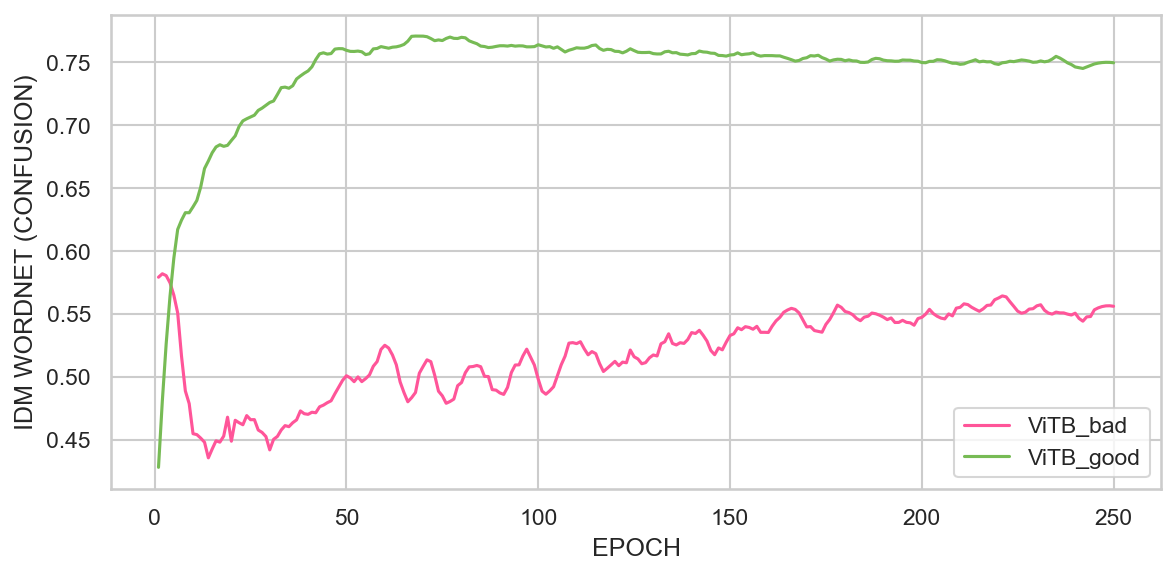}}
\end{center}
\caption{Mini-ImageNet: All DM Curves for a ``good'' and a ``bad'' ViT.}
\label{fig:dm_badnet}
\end{figure}

\begin{figure}[h]
\centering
\subfloat[][NCSM (1)]{\includegraphics[width=0.25\textwidth]{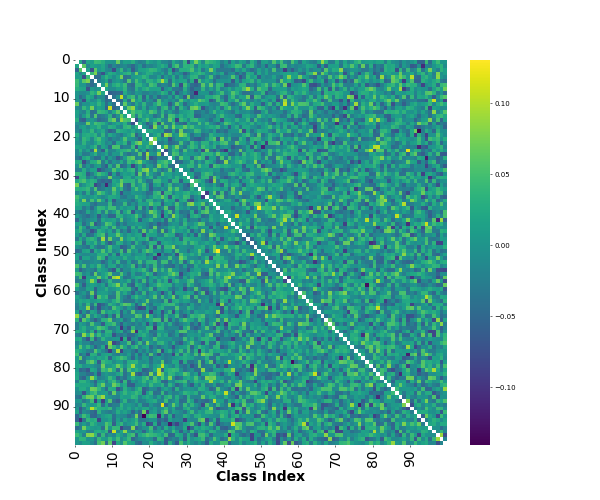}}
\subfloat[][NCSM (5)]{\includegraphics[width=0.25\textwidth]{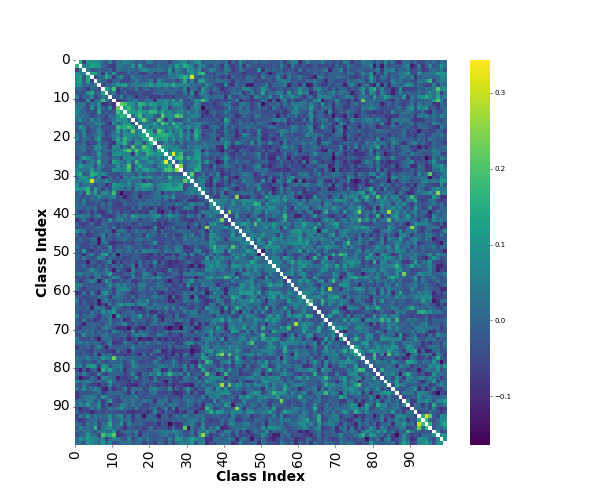}}
\subfloat[][NCSM (25)]{\includegraphics[width=0.25\textwidth]{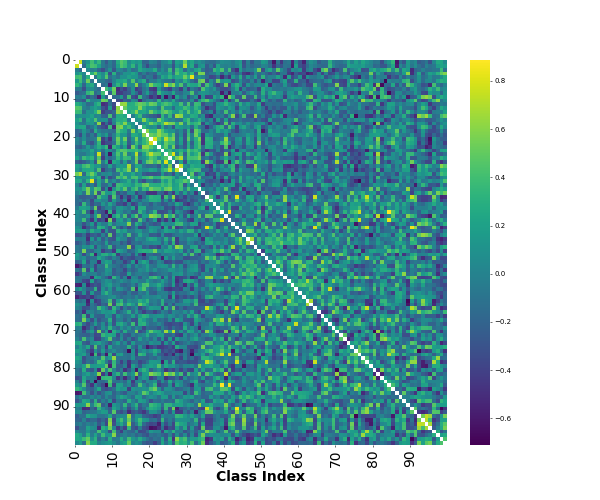}}
\subfloat[][NCSM (200)]{\includegraphics[width=0.25\textwidth]{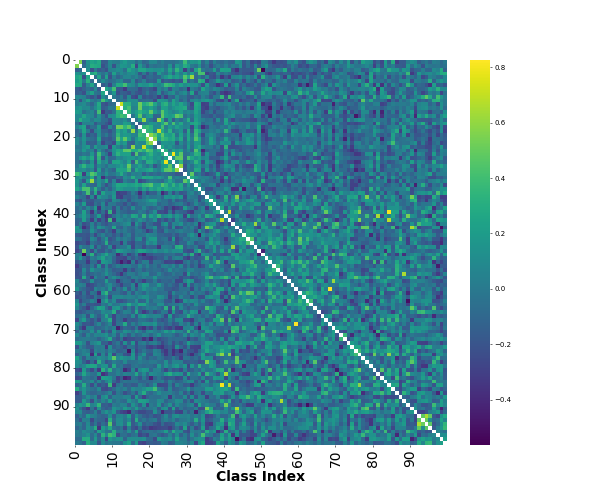}}
\\
\subfloat[][CCSM (1)]{\includegraphics[width=0.25\textwidth]{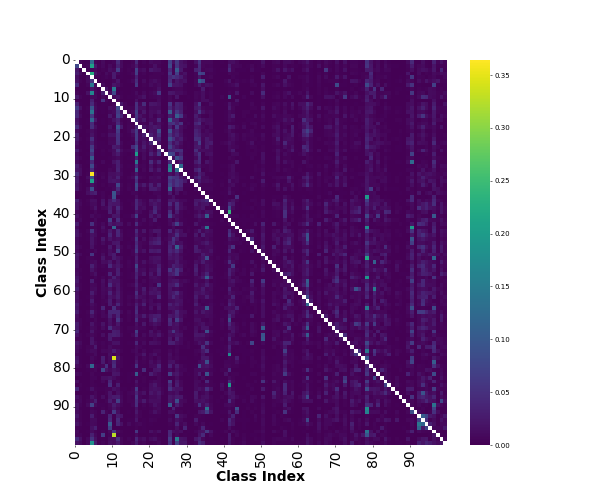}}
\subfloat[][CCSM (5)]{\includegraphics[width=0.25\textwidth]{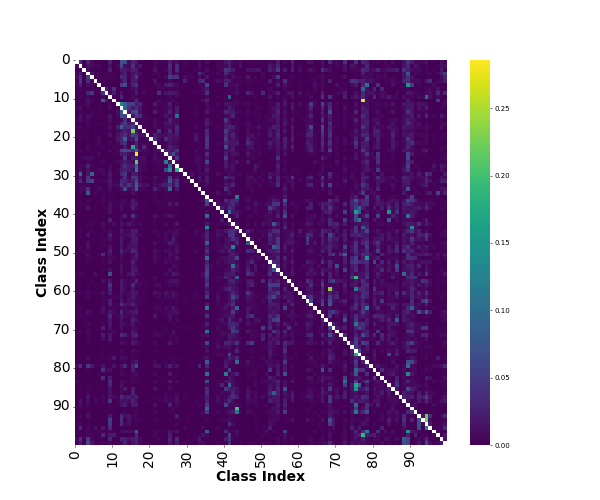}}
\subfloat[][CCSM (25)]{\includegraphics[width=0.25\textwidth]{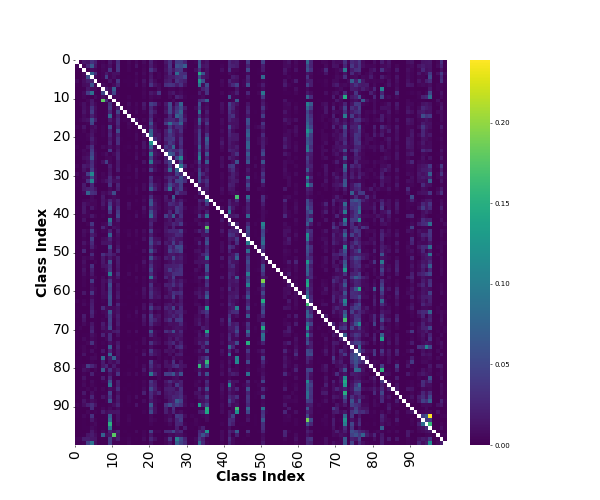}}
\subfloat[][CCSM (200)]{\includegraphics[width=0.25\textwidth]{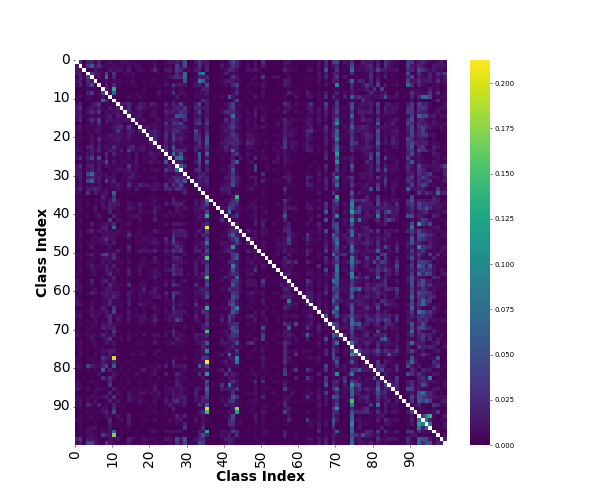}}
\caption{Mini-ImageNet: Network Class Similarity Matrices and Confusion-based Class Similarity Matrices for a ``bad'' ViTB for at different epochs (number in brackets). }
\label{fig:badnet_ncsm_ccsm}
\end{figure}

Now, let us focus on the quality of a ``bad'' network's predictions and more in-detail analysis of its mistakes. In Fig. \ref{fig:dm_badnet}, we present all DM variants (both the network-based and the WordNet-based ones). It is visible here, that the quality of the ``bad'' network's prediction is much lower than the one of a ``good'' network, although some improvement with time can be noticed (which i barely visible in the test accuracy plot), showing that the similarity perception optimization does introduce some performance improvements in the latter training stages (the \textit{error refinement phenomenon} is visible also for ``bad'' networks, but happens later in the training). This improvement (although not visible in the accuracy plot) is also visible as the decrease in the loss plot, therefore the IDM plots can be used as an explanation of this decrease. The errors-only variant shows, that although the improvements undoubtedly happen, the mistakes still are on average placed around the half of the class space (both for the network-based and the WordNet-based variants). It indicates high randomness of the mistakes being made. It is also visible in CCSMs obtained for our ``bad'' network in Fig. \ref{fig:badnet_ncsm_ccsm}. While for the 5th epoch, a hierarchy is visible, it is much less prominent for the later epochs (although some hierarchical groups are slightly visible, \emph{e.g.} close to the animal classes).

To summarize, also for bad networks, some similarity phases can be defined, however they are significantly different than for ``good'' networks. Here, we define 2 most prominent phases (we discard some initial gains of the network at the beginning of the training, as they are most probably related to the initial accuracy/loss improvement of the network). These phases are: (1) \textit{initial similarity drop} and (2) \textit{stable similarity growth}.

Our last finding can be further developed into a method of assessing the progress of the network's training and monitoring its potential overfitting (or as in ``good'' and ``bad'' networks example, as an early indicator of overall model performance). With using our metrics, different phases of the network training can be distinguished as the training progresses, and can help with early stopping, checkpointing or managing learning rate during training. It can also be further developed as a loss function component for added regularization -- both of the use cases will be considered by us in our future works.

\clearpage

\section{Is similarity perception emergence tied solely to object recognition and large datasets?}\label{appendix_otherdatasets}

After the initial experiments with the object recognition model trained on Mini-ImageNet (the main body of the paper) and CIFAR100, we decided to performs some qualitative experiments with models trained on different datasets and tasks to see whether our results have a potential to generalize to (1) smaller/larger datasets than the medium-sized datasets used in the study and (2) models for object detection/scene segmentation. For the purpose of (1), we generated additional NCSMs for example models trained on ImageNet-1k (a larger dataset with 1000 classes) and CIFAR10 (a smaller dataset with only 10 classes, at a higher level of abstraction than CIFAR100 and Mini-ImageNet).

\begin{figure}[h]
\centering
\subfloat[][mini-ImageNet - ConvNeXt-T]{\includegraphics[width=0.45\textwidth]{FIGS/ImageNet_matrices/ConvNeXt199.png}}
\subfloat[][ImageNet-1k - ConvNeXt-S]{\includegraphics[width=0.45\textwidth]{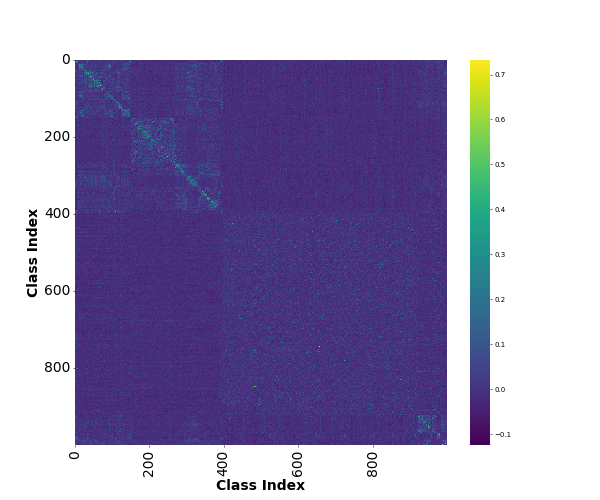}}
\caption{Network Class Similarity Matrices obtained for 2 ImageNet versions. }
\label{fig:imagenet}
\end{figure}

In the case of ImageNet-1k, we used a trained ConvNeXt-S model from \url{https://huggingface.co/facebook/convnext-small-224}. We present its NCSM in Fig. \ref{fig:imagenet} along with the NCSM obtained for ConvNeXt-T from our experiments (at epoch 200). It is visible that a similar characteristic structure of the semantic categories has been developed by two models. While the ranges of the matrices values differ, the overall structure stays the same, suggesting that our results generalize also to larger datasets.

In the case of CIFAR10, we trained an example small network (a simple, sequential model) that obtained app. 85\% accuracy on the test set (we provide the structure of this network and the seed used for weights initialization in our GitHub repository for reproducibility - supplementary materials and Zenodo during the revision stage). Although CIFAR10 has a significantly shallower semantic hierarchy of concepts than CIFAR100 and Mini-ImageNet, also in this case clear semantic groups can be distinguished (in the left upper corners - animals: 'bird', 'frog', 'dog', 'cat', 'horse', 'deer' and the 2nd group - vehicles: 'aircraft', 'car', 'boat', 'truck'). This also supports the theories from the cognitive psychology than the similarities in the world are revealing and the visual structure of the world is highly correlated \cite{bib:rosch1978cognition,bib:medin1993respects}.

\begin{figure}[h]
\centering
\subfloat[][CIFAR100]{\includegraphics[width=0.45\textwidth]{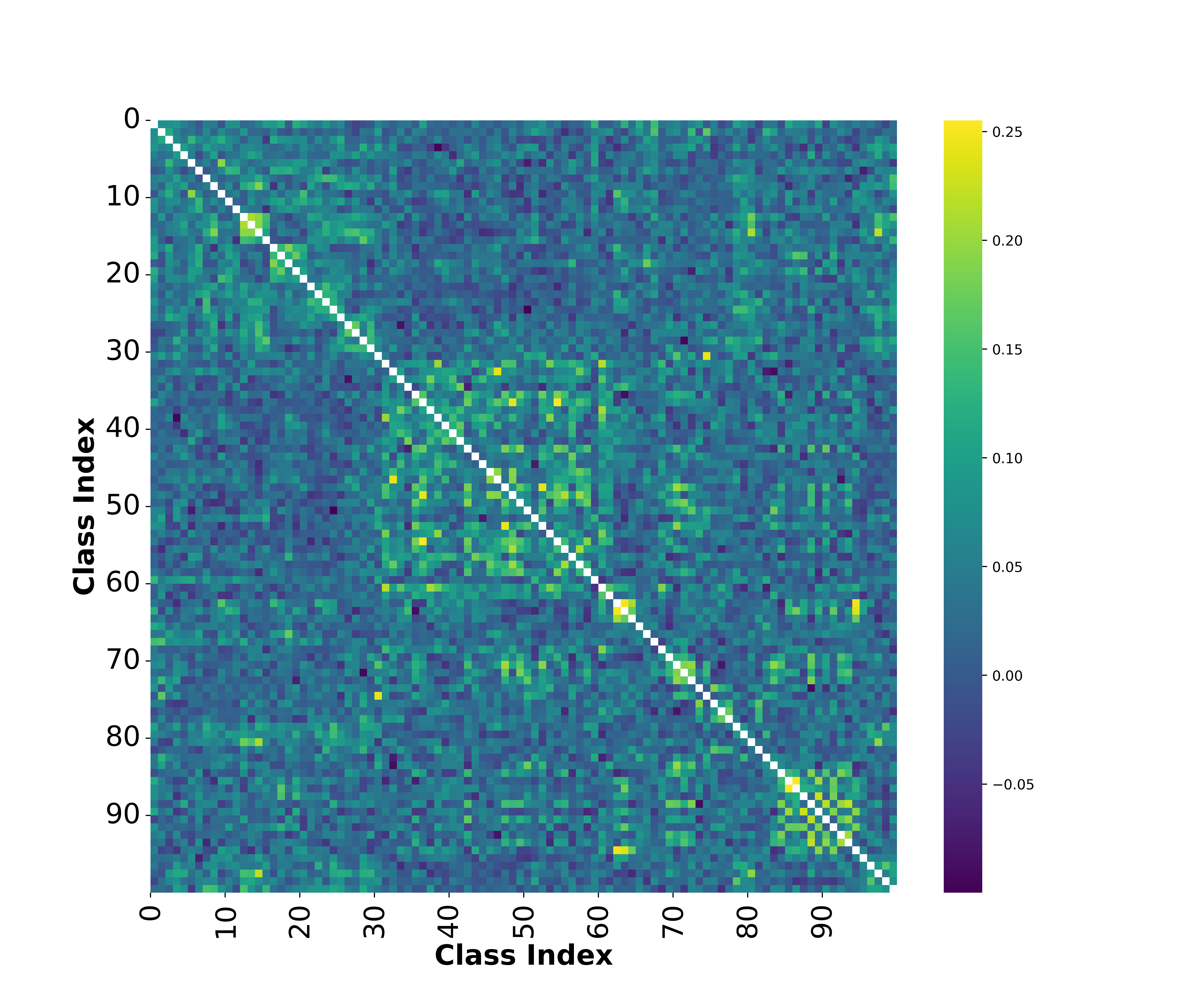}}
\subfloat[][CIFAR10]{\includegraphics[width=0.45\textwidth]{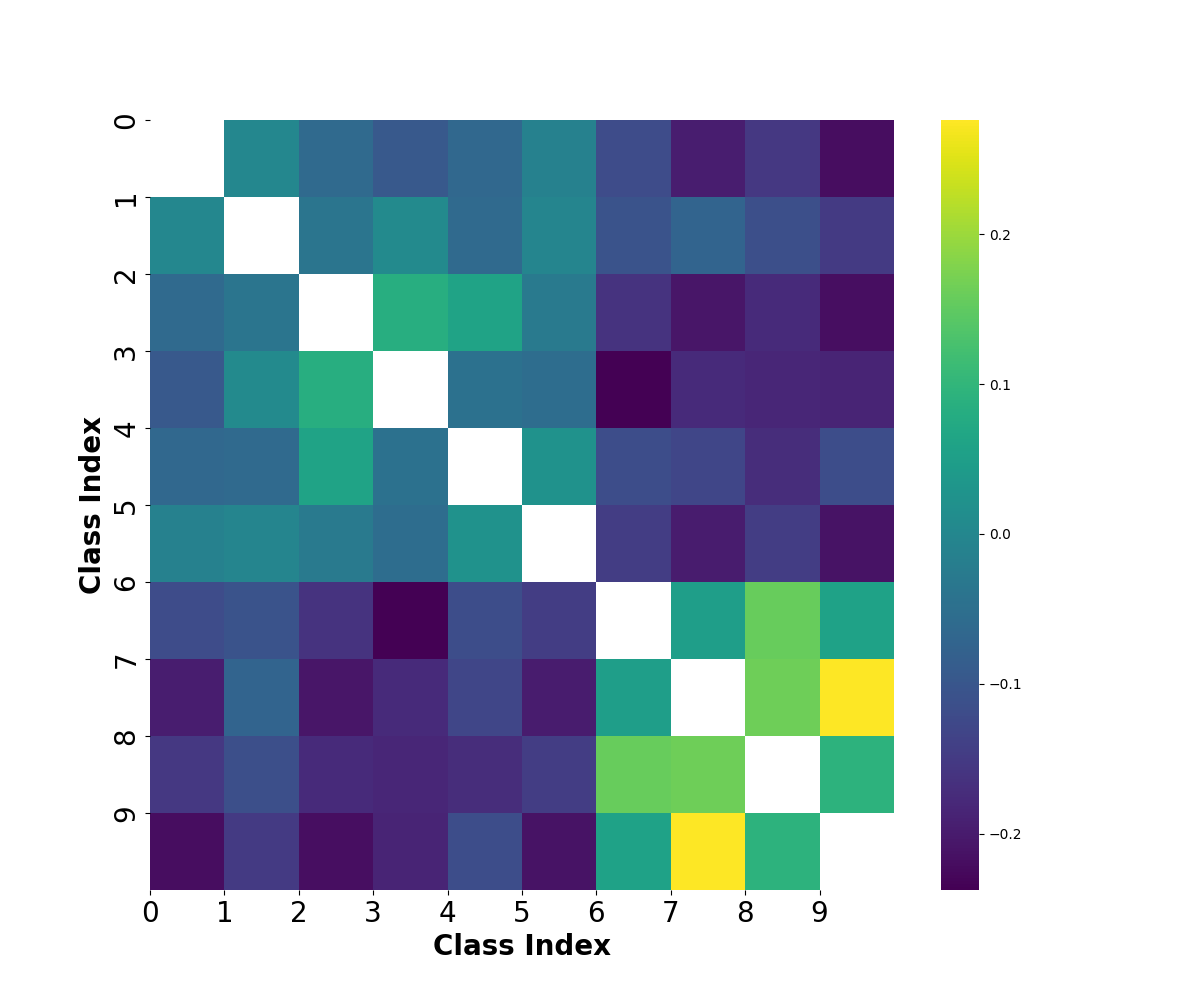}}
\caption{Network Class Similarity Matrices obtained for 2 available versions of CIFAR. }
\label{fig:histograms}
\end{figure}

\begin{figure}[htp]
\centering
\subfloat[][WordNet\\]{\includegraphics[width=0.33\textwidth]{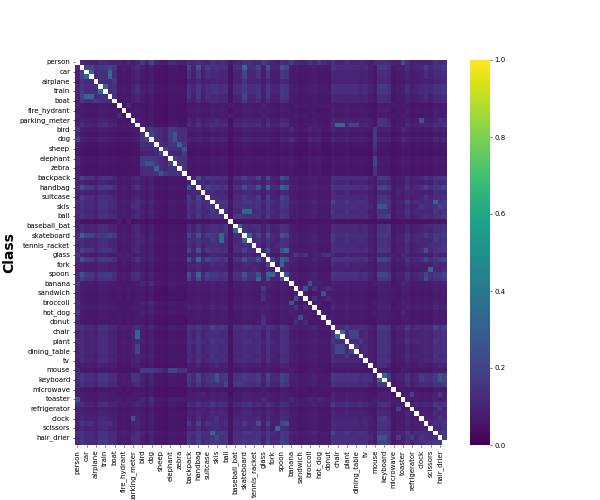}}
\subfloat[][YOLOS-T\\Object Detection]{\includegraphics[width=0.33\textwidth]{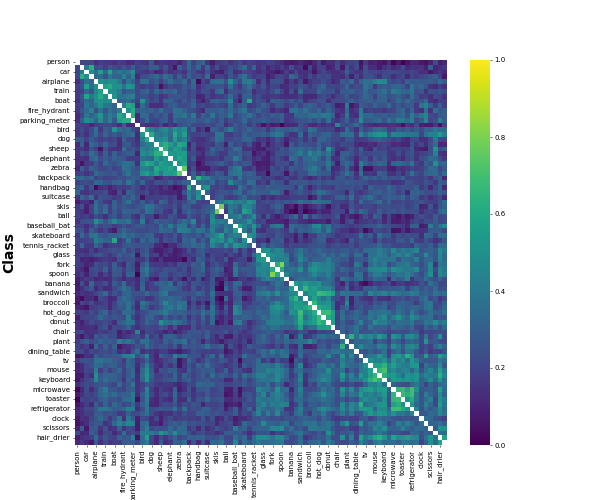}}
\subfloat[][DETR\\Object Detection]{\includegraphics[width=0.33\textwidth]{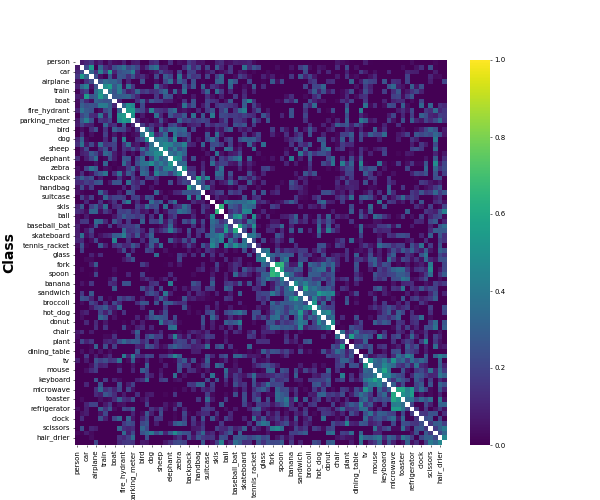}}
\\
\subfloat[][DETR\\Panoptic Segmentation]{\includegraphics[width=0.33\textwidth]{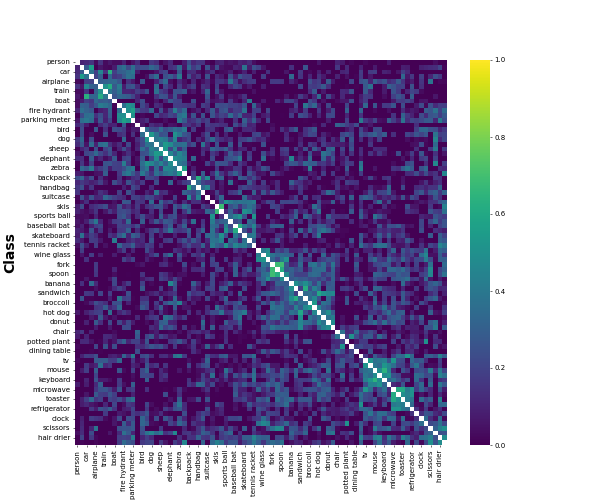}}
\subfloat[][MaskFormer\\Instance Segmentation]{\includegraphics[width=0.33\textwidth]{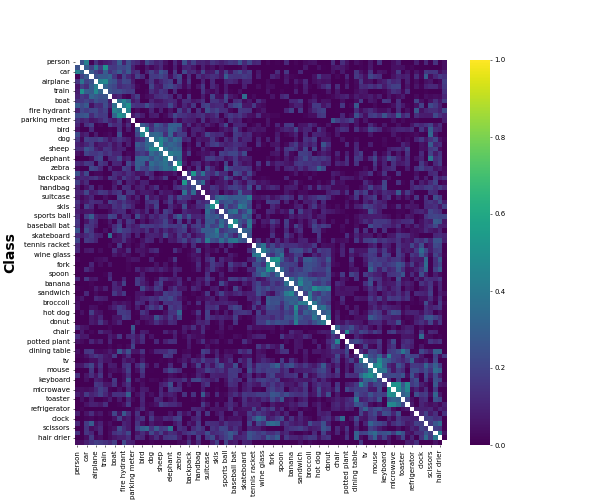}}
\subfloat[][MaskFormer\\Panoptic Segmentation]{\includegraphics[width=0.33\textwidth]{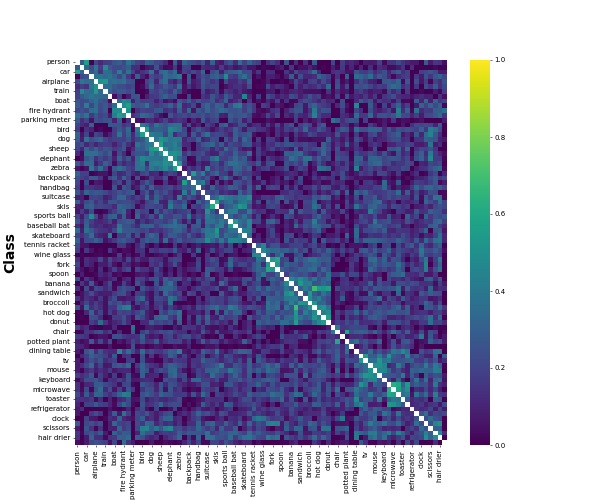}}
   \caption{Network Class Similarity Matrices obtained for COCO classes and different object detection and segmentation networks. In all cases a clear hierarchy is visible. Relations, nevertheless, are richer that the WordNet-based ones. \emph{e.g.} they rely a lot on context - see how 'fire hydrant' and 'road sign' are placed together in the first square along different vehicles.}
\label{fig:other_nets}
\end{figure}

We also performed a qualitative experiment with networks trained on different tasks on COCO 2017 dataset. We use the following available networks from HuggingFace: DEtection TRansformer (DETR) model with ResNet-50 backbone for object detection (\url{https://huggingface.co/facebook/detr-resnet-50}), YOLOS-t for object detection (\url{https://huggingface.co/hustvl/yolos-tiny}), MaskFormer model for COCO instance segmentation (\url{https://huggingface.co/facebook/maskformer-swin-tiny-coco}), Mask2Former model for COCO panoptic segmentation (\url{https://huggingface.co/facebook/mask2former-swin-base-coco-panoptic}) and DETR for COCO 2017 panoptic segmentation (\url{https://huggingface.co/facebook/detr-resnet-50-panoptic}). We provide the NCSMs generated with our framework in Fig. \ref{fig:other_nets} (it is worth noting that this method is compatible with all models that include a standard label classifier, therefore also the one present in the classifier of the object detection/segmentation networks). We also provide the WCSM for COCO in this figure. The results show that all networks used in this experiment managed to develop a clear hierarchical structure of similarity. In many parts (boxes) is is very similar to the structure of the generated WCSMs, but when we analyze the NCSMs closer, it is visible that networks rely on more semantic relations than the ones reflected by WordNet (\emph{e.g.} look how 'fire hydrant' and 'road sign' are placed together in the first square along with different vehicle types, suggesting the importance of context and co-occurence of objects in the process of similarity perception development).

\newpage

\section{Closer inspection on the WSI curves of CNNs}\label{appendix_wsi_cnn}

\begin{figure}[h]
\begin{center}
\subfloat[][Mean WSI]{\includegraphics[width=0.49\textwidth]{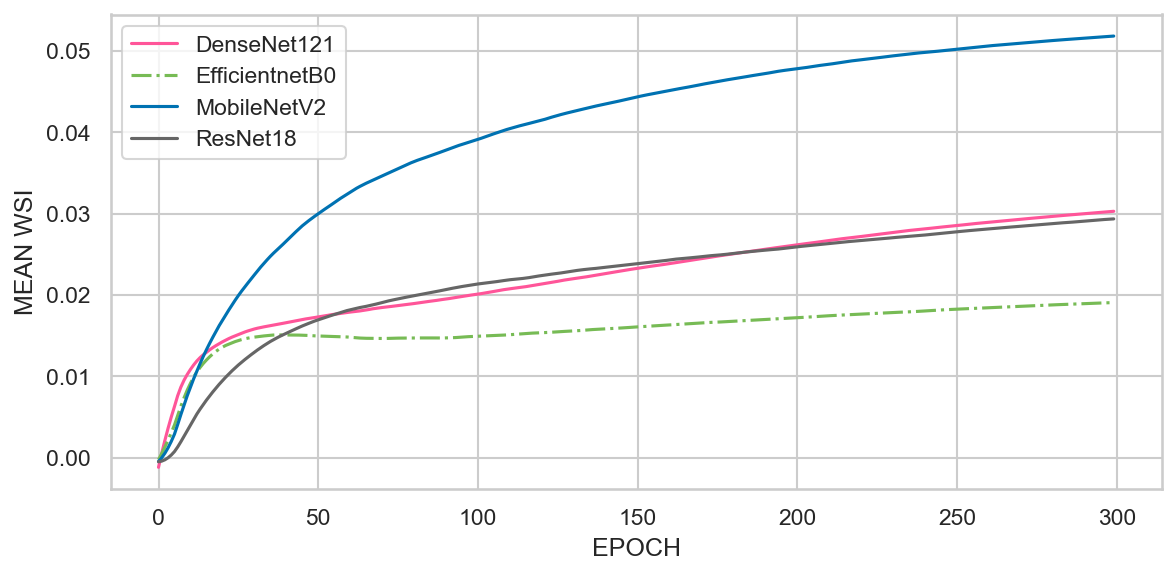}}
\subfloat[][Max WSI]{\includegraphics[width=0.49\textwidth]{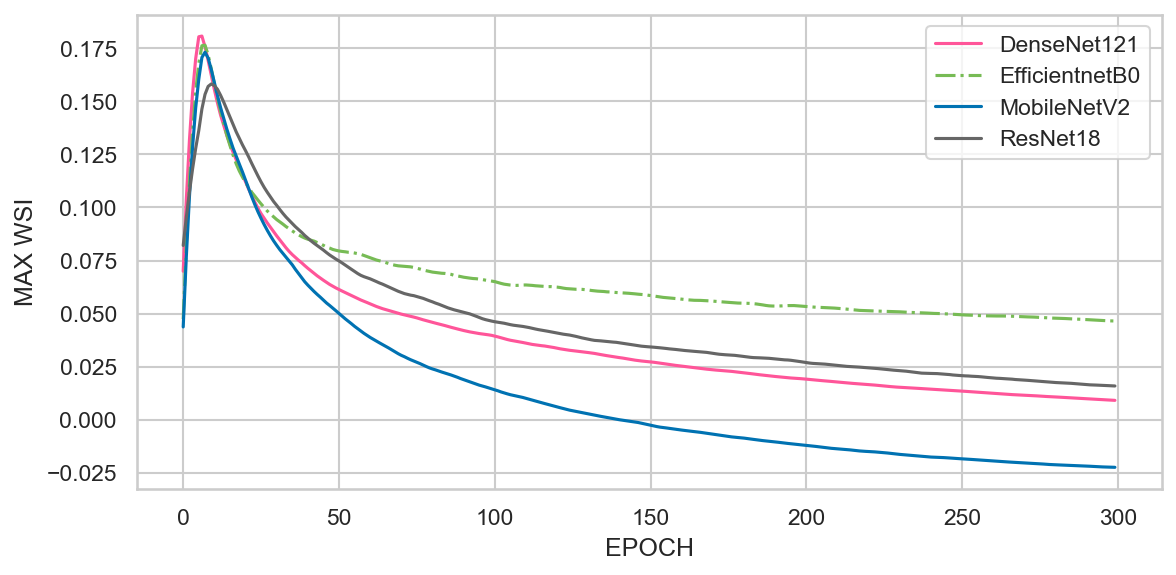}}
\\
\subfloat[][Min WSI]{\includegraphics[width=0.49\textwidth]{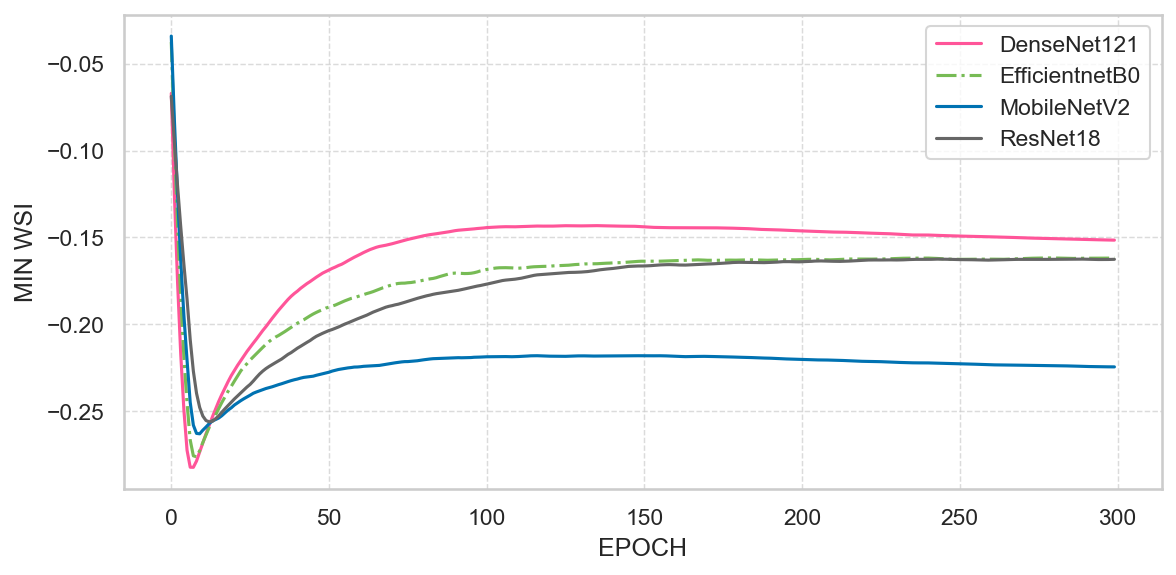}}
\end{center}
\caption{Closer inspection on the WSI curves for different CNNs on Mini-ImageNet dataset. }
\label{fig:cnn_wsi}
\end{figure}

In Figure \ref{fig:WSI_Mini} in Section \ref{sec:sai_sem} of the main paper we observed a clearly visible difference in the mean Weight Similarity Index (mean WSI, describing the mean similarity of weight templates within a network) curves for ResNet18 and MobileNetV2 and other models (ViTs, hybrids), which was surprising. It prompted us to better explore this phenomenon. As both of these models represent convolutional neural networks, our hypothesis was that such a mean WSI curve is characteristic for CNNs. To better prove it, we chose two additional CNNs, namely DenseNet121 and EfficientNetB0. We trained them on mini-ImageNet. In Figure \ref{fig:cnn_wsi}, we present the results obtained for these models and different WSI variants. The results support our hypothesis - also these new models result in a similar Mean WSI curve, which further highlights the impact of neural network architecture on the similarity perception.

\section{Closer inspection of network's mistakes during training}\label{appendix_mistakes}

\begin{figure}[h]
\begin{center}
\subfloat[][Number of mistakes for different epochs]{\includegraphics[width=0.485\textwidth]{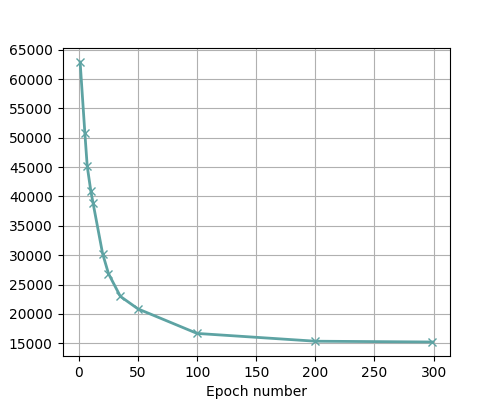}}
\subfloat[][Histogram of the testing templates similarities]{\includegraphics[width=0.49\textwidth]{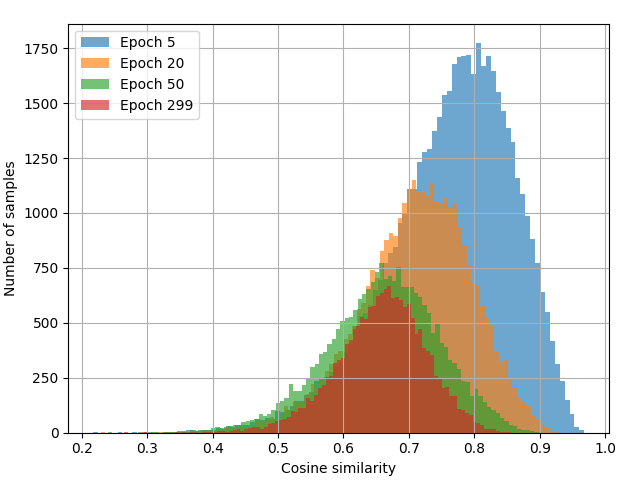}}
\end{center}
\caption{Analysis of the changing number of mistakes at different epochs. }
\label{fig:number_mistakes}
\end{figure}

In Section \ref{sec:idm_confusion} of the paper, we analyzed whether the confusion patterns of CNNs and ViTs match their similarity perception throughout the training. For this purpose, we used, among others, our proposed method, the IDM curves. The results showed us an interesting phenomenon: the errors-only IDM revealed that after reaching the peak, its values slightly drop. It indicates that the network is making errors between categories it perceives as less similar. We hypothesized that the reason can be that the network's accuracy is already high then, and more 'obvious' mistakes have been eliminated. The network is now tackling more challenging and less typical samples, or even potential noise from mislabeled data. We now take a closer look on the network's mistakes. First of all, we examine the difference in the similarity perception of correctly and incorrectly classified samples by an example network used in our analysis - MobileNetV2. To do this, we first extract the templates for all samples  from the training and testing datasets. After averaging, we obtain 2 matrices storing the dataset-level templates. We use them as a reference. Then, we create the average template matrices, but separately for the correctly and incorrectly classified images. For the two splits of the dataset, we obtain 4 matrices of templates in total. We then compare the average templates for the correctly and incorrectly classified samples with the overall template via cosine similarity (for the whole dataset). Note that this comparison is done for particular classes, so we compare with each other the templates of the same class. We present the results in Table \ref{tab:cos_templates}. The results show that in the later epochs, the samples that are misclassified are more distant (less similar) in the feature space from those classified correctly. This also suggests that the network makes more informed mistakes at this point, and that the drop in the IDM is due to less typical/difficult examples, or even mislabeling issues. In Fig. \ref{fig:number_mistakes}, we also included the plot of the number of mistakes as a function of the epoch number. The plot shows that the number of mistakes decreases in a logarithmic fashion (which is expected, as the network learns its task). As the number of errors decreases, the remaining errors caused by the less typical/difficult etc. samples become more prominent within the set of incorrectly classified samples (also visible in the second plot of \ref{fig:number_mistakes}, presenting the histograms of cosine similarity values between the incorrectly classified samples and their templates for 4 example epochs). That is why a slight decrease in errors-only IDM occurs, but overall (as IDM is still high) mistakes are still driven by the similarity perception of the network. These observations align with our earlier findings. To examine this even more comprehensively (image-level), we take MobileNetV2 and manually analyze the images, for which the network made mistakes to show in a practical and straight-forward way what samples cause errors.

{\footnotesize
\begin{table}\caption{Average cosine similarity between averaged templates obtained for the correctly and incorrectly classified samples after different training epochs.}\label{tab:cos_templates}
\resizebox{\textwidth}{!}{\footnotesize
\begin{tabular}{lrrrrrrrrrr}
\toprule
Data split & \multicolumn{10}{c}{Epoch} \\
\cmidrule(lr){2-11}
 & 7 & 10 & 12 & 20 & 25 & 35 & 50 & 100 & 200 & 299 \\
\midrule
Train-True  & 0.978 &\textbf{ 0.981} & \textbf{0.984} & \textbf{0.991} & \textbf{0.994} & \textbf{0.996} & \textbf{0.998} & \textbf{1.} & \textbf{1.} & \textbf{1.} \\
Train-False  & \textbf{0.984} &\textbf{ 0.981} & 0.980 & 0.972 & 0.964 & 0.951 & 0.950 & 0.930 & 0.893 & 0.859 \\
\midrule
Test-True  & 0.977 & 0.980 & \textbf{0.982} & \textbf{0.989} & \textbf{0.993} &\textbf{ 0.994} & \textbf{0.995} & \textbf{0.997} & \textbf{0.997} & \textbf{0.997} \\
Test-False  & \textbf{0.983 }& \textbf{0.983} & 0.981 & 0.973 & 0.966 & 0.957 & 0.955 & 0.949 & 0.952 & 0.954 \\
\bottomrule
\end{tabular}}
\end{table}
}

\begin{figure}[h]
\centering
\subfloat[][\raggedright golden \\ retriever \\ $\downarrow$ \\ malamute ]{\includegraphics[height=0.11\textheight]{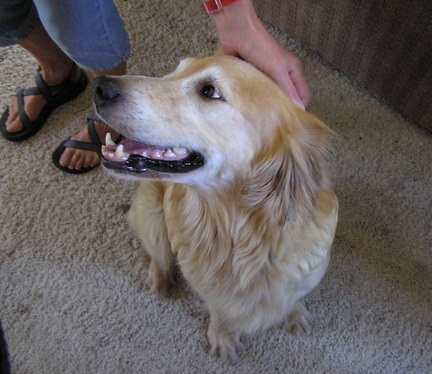}}
\subfloat[][\raggedright electric \\ guitar\\ $\downarrow$ \\ triceratops ]{\includegraphics[height=0.11\textheight]{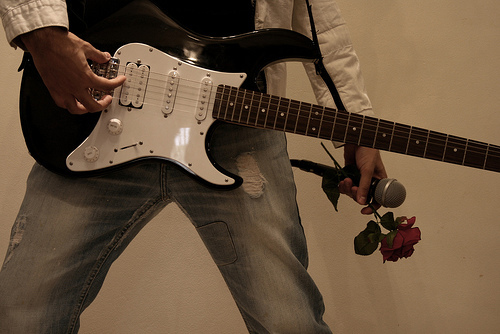}}
\subfloat[][\raggedright missile \\ $\downarrow$ \\ dalmatian ]{\includegraphics[height=0.11\textheight]{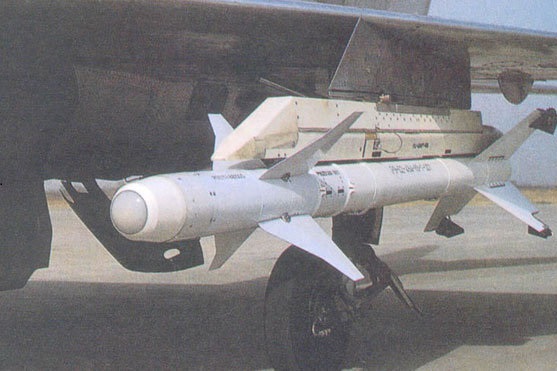}}
\subfloat[][\raggedright tobacco \\ shop \\ $\downarrow$ \\ school \\ bus ]{\includegraphics[height=0.11\textheight]{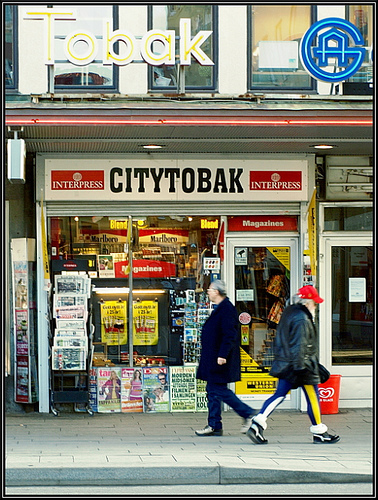}}
\subfloat[][\raggedright upright \\ $\downarrow$ \\ green \\ mamba]{\includegraphics[height=0.11\textheight]{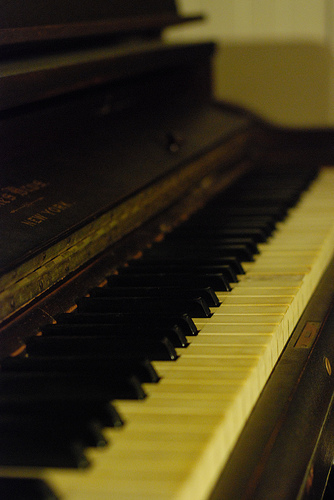}}
\caption{Inspection of MobileNetV2's mistakes on the test set at epoch \textbf{5} and Mini-ImageNet. (a) a rather typical image, mistaken with a similar class (another breed), (b), (c), (d), (e) a very clear, typical image, yet mistaken with a very dissimilar class.  }
\label{fig:examples_mistakes_5}
\end{figure}

\begin{figure}[h]
\centering
\subfloat[][\raggedright golden \\ retriever \\ $\downarrow$ \\ saluki ]{\includegraphics[height=0.11\textheight]{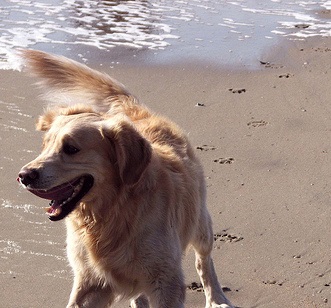}}
\subfloat[][\raggedright tibetan \\ mastiff \\ $\downarrow$ \\ newfoundland ]{\includegraphics[height=0.11\textheight]{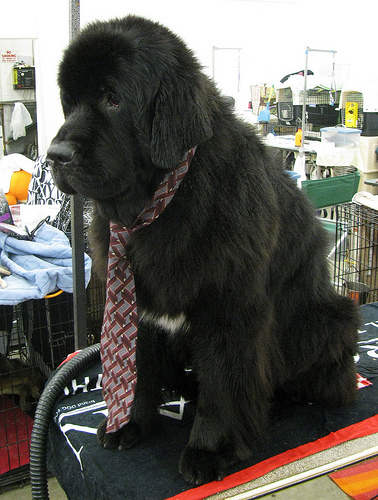}}
\subfloat[][\raggedright parallel \\ bars \\ $\downarrow$ \\ horizontal \\ bar ]{\includegraphics[height=0.11\textheight]{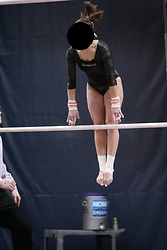}}
\subfloat[][\raggedright stage \\ $\downarrow$ \\ electric  guitar ]{\includegraphics[height=0.11\textheight]{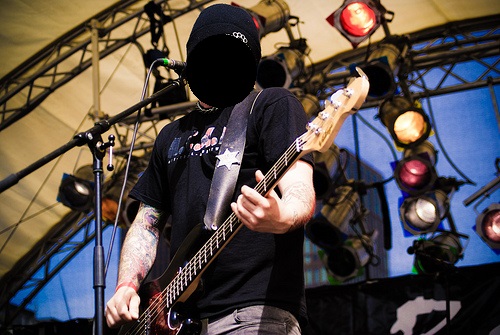}}
\subfloat[][\raggedright wok \\ $\downarrow$ \\ frying \\ pan]{\includegraphics[height=0.11\textheight]{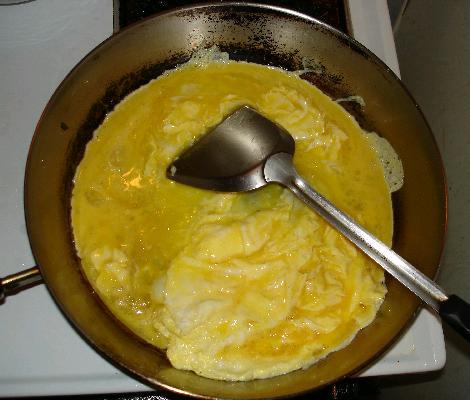}}
\caption{Inspection of MobileNetV2's mistakes on the test set at epoch \textbf{50} and Mini-ImageNet. (a) A rather typical image of a dog mistaken with a similar class (another, similar breed of dog), (b) A rather typical image of a dog mistaken with a similar class (another, similar breed of dog), (c) A mistake due to the occurrence in the class space of 2 highly similar classes, (d) Both labels are true (cooccurence), (e) It is a normal pan, not a wok - mislabeling issues.  }
\label{fig:examples_mistakes_50}
\end{figure}

\begin{figure}[h]
\centering
\subfloat[][\raggedright golden \\ retriever \\ $\downarrow$ \\ dome ]{\includegraphics[height=0.12\textheight]{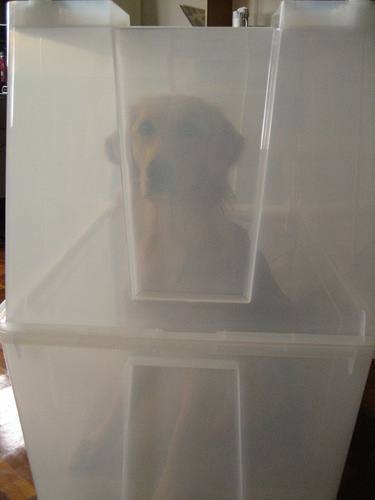}}
\subfloat[][\raggedright golden\\  retriever \\ $\downarrow$ \\ electric \\guitar ]{\includegraphics[height=0.12\textheight]{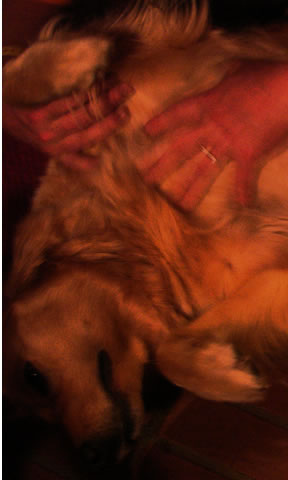}}
\subfloat[][\raggedright ant \\ $\downarrow$ \\ jellyfish \\ ]{\includegraphics[height=0.12\textheight]{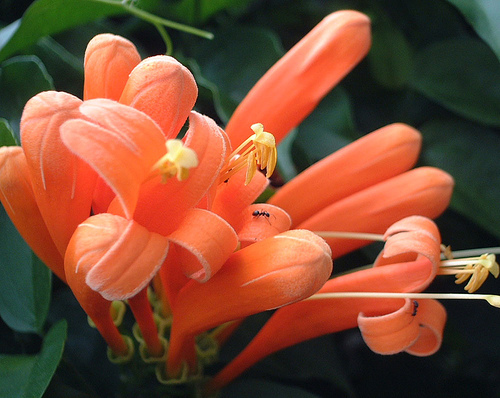}}
\subfloat[][\raggedright wok \\ $\downarrow$ \\ frying pan  \\ ]{\includegraphics[height=0.12\textheight]{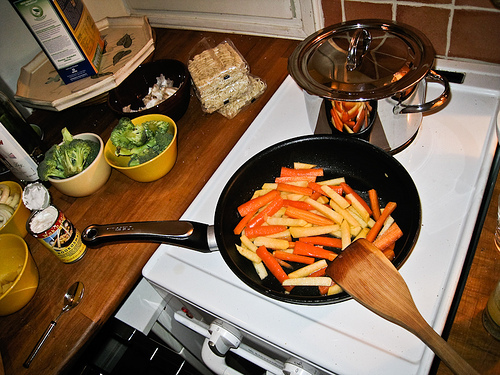}}
\subfloat[][\raggedright wok \\ $\downarrow$ \\ mixing bowl ]{\includegraphics[height=0.12\textheight]{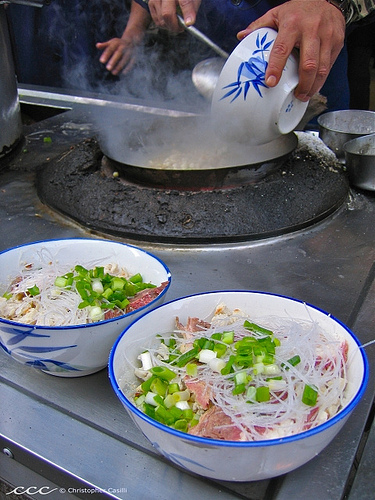}}
\caption{Inspection of MobileNetV2's mistakes on the test set at epoch \textbf{299} and Mini-ImageNet. (a) represents a less typical picture of a dog (it is less visible due to it being closed in a box), (b) a more difficult to categorize picture of a dog due to it being blurry, also - the hands can be connected with a guitar, (c) difficult image, an object is very small and surrounded by different objects, also the flower is very colorful and overwhelms the picture, (d) it is a normal pan, not a wok - mislabeling issues, (e) both labels are true (co-occurence), bowls are more visible.  }
\label{fig:examples_mistakes_299}
\end{figure}

In Figures \ref{fig:examples_mistakes_5}, \ref{fig:examples_mistakes_5} and \ref{fig:examples_mistakes_5}, we present the samples images, for which the network made mistakes after the 5th, 50th and 299th epoch respectively. Figure \ref{fig:examples_mistakes_5} shows that in the early epochs of training, the mistakes of the network are not very reasonable (\emph{e.g.} a missile is named a dalmatian - mistakes between very unrelated objects), which can be expected, but at that point it is also visible that at least some similarities are learned by the network (\emph{e.g.} it mistakes a golden retriever with another dog breed). Close the the IDM peak (see Figure \ref{fig:examples_mistakes_50}), the network mostly make mistakes between similar classes (which is shown by the high IDM values in Section \ref{sec:idm_confusion}). \emph{e.g.}, it makes mistakes between similar dog breeds, between very similar classes (\emph{e.g.} parallel and horizontal bar). The errors are also cause by the mislabeling issues (\emph{e.g.} wok/frying pan), therefore they are not truly the mistakes sometimes. In Fig. \ref{fig:examples_mistakes_299}, we can see that in later epochs (when the IDM errors only variant values drop), the network makes mistakes on the less typical images (\emph{e.g.} a dog hidden in a plastic box), difficult images (\emph{e.g.} blurry images, very small target objects), due to mislabeling issues (\emph{e.g.} a frying pan named a wok) or due to a coocurrence of objects in the same picture. These examples support the hypothesis made earlier.

\section{Inspection of the cyclical nature of the bumps on the network semantic similarity alignment curve}\label{appendix_bumps}

In Section \ref{sec:sai} of the main paper, we aimed to examine how does the network’s similarity perception change during training for CNNs and ViTs and whether it is in line with semantic similarity. In Figure \ref{fig:sai_network_wordnet}, we presented the SAI(NCSM, SCSM) curve (the alignment between the Network Class Similarity Matrices and Semantic Similarity Matrix obtained via WordNet). In this figure, visible bumps can be observed (while the alignment slightly decreases). We named this the similarity perception 'refinement'.

\begin{figure}[h]
\begin{center}
\subfloat[][Train loss, SAI(NCSM, SCSM)]{\includegraphics[width=0.49\textwidth]{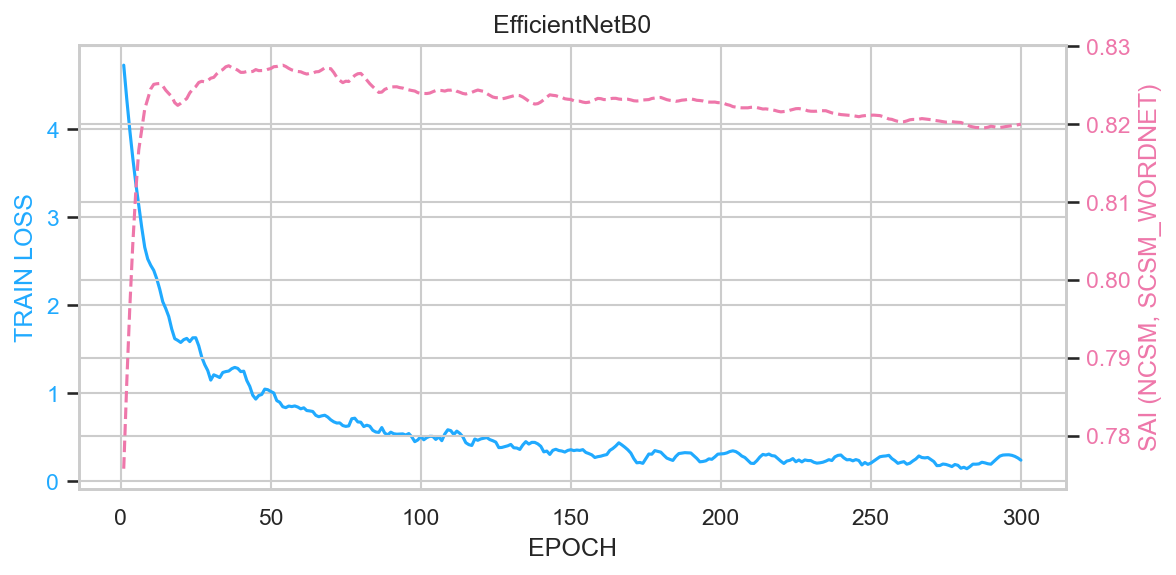}}
\subfloat[][Learning Rate, SAI(NCSM, SCSM)]{\includegraphics[width=0.49\textwidth]{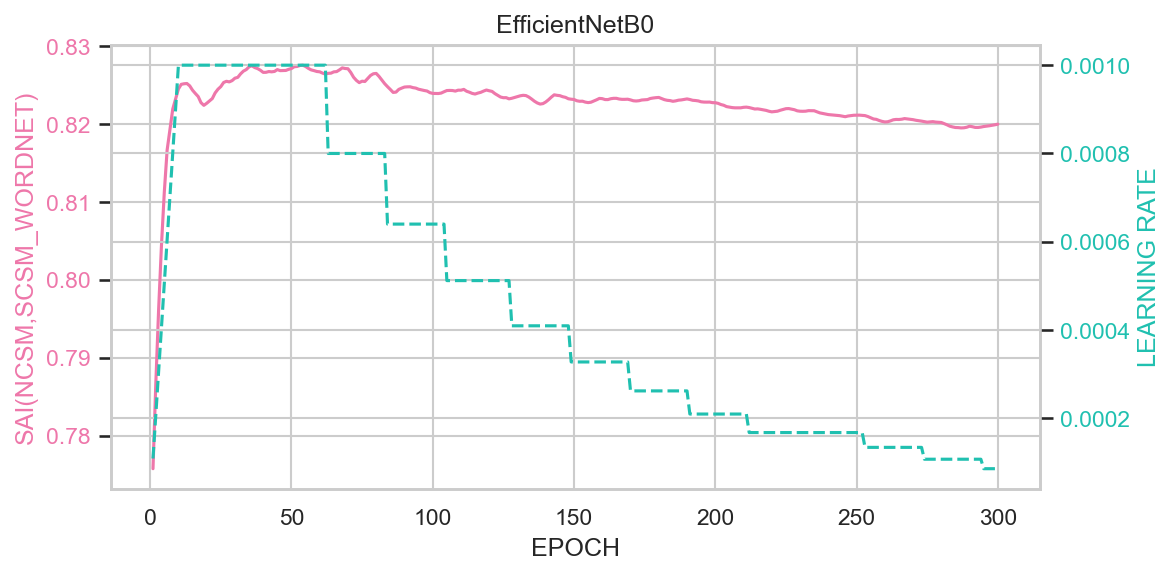}}
\end{center}
\caption{Inspection the cyclical nature of the bumps on the network semantic similarity alignment curve for EfficientNetV2B0 trained on Mini-ImageNet. }
\label{fig:bumps}
\end{figure}

In this section, we provide some additional plots in Fig. \ref{fig:bumps} showing the figures of the train loss and learning rate presented in the same plots together with this SAI variant (for an example network trained for our additional CNN-focused experiments - EfficientNetV2B0). These bumps occur in the phase that the network still dynamically learns (see the train loss plots in Fig. \ref{fig:bumps}). It is visible that the loss and SAI curves are similar but do not present the same thing. While they both steeply increase/decrease in the first stage, in the second stage the loss curve still decreases dynamically. The SAI curve is characterized by a stable trend (however with visible bumps) or slight decreases (and not increase this time). It shows that at this point, it is not the similarity perception learning of the classes, but rather their differentiation. In the loss curve, also some bumps are visible in this phase. These bumps occur in the period, in which the learning  rate scheduler makes the learning rate constant and then utilizes learning rate decay. These abrupt changes can cause the loss and the SAI curve to temporarily spike as the optimizer adjusts. It is visible that the SAI curve in Figure \ref{fig:bumps} stabilizes when learning slows down and the changes in the learning rate become less abrupt. Also, techniques such as dropout add noise during training, so this noise can also cause some temporary instability in the loss/SAI curves.

\section{Similarity inspection beyond the object categorization}\label{appendix_contrastive}

In the main paper, we focused on the similarity inspection of different object recognition networks. In Section \ref{appendix_otherdatasets}, we examined additional networks trained on object detection and segmentation showing that the similarity perception also emerges in such networks in a similar way as it occurs in the object recognition models. Nevertheless, all of these models have a lot of commonalities. \emph{e.g.} even though the object detection and segmentation networks differ in task from the object recognition, they often use the backbones pretrained on object recognition (usually ImageNet). Also, in all of these networks some kind of an object-level classifier occurs (which makes it possible to use our weight-based similarity computation method). They are all trained with the supervised approach and the optimization process of these networks is also similar.

Both from the perspective of computer vision and deep learning, there are many other training objectives for visual learning and understanding, \emph{e.g.} self-supervised and contrastive learning approaches \cite{bib:oquab2023dinov2,bib:chen2020simple,bib:margalit2024unifying}. On the other hand, also models trained on joint text-image objectives are available \cite{bib:radford2021learning}. Such networks learn strong semantically meaningful representations and can produce strong candidate models of the visual processing.

In this section, we present some additional results for two models trained with self-supervised objectives (not for object recognition): DINOv2 \cite{bib:oquab2023dinov2} and CLIP \cite{bib:radford2021learning}. DINOv2 is a model trained with a self-supervised learning framework designed specifically to produce high-quality image representations. CLIP, on the other hand, learns a common representation space for images and text simultaneously, which enables cross-modal tasks and the understanding not only of visual, but also textual semantic relations. As those models, during pre-training do not include a traditional classifier, the classifier's weights cannot be used to produce the Class Similarity Matrix. Nevertheless, even with those models we can enable a similarity-based analysis. To do this, we need the annotated dataset (\emph{e.g.} Mini-ImageNet). In the evaluation step of the training, a dataset is used to extract templates from the network. These templates are aggregated for each class as it is done for object recognition networks in \emph{e.g.} work \cite{bib:huang2021semantic}. This step requires significantly more computational resources, however it can be treated as an alternative for our approach in the cases, in which we cannot use the weights of the classifier, making our method suitable also for self-supervised approaches. We can name this NCSM (Network Class Similarity Matrix) variant Templates-based Network Class Similarity Matrix (TNCSM).

TNCSMs can be obtained also for traditional classifier, however, as mentioned before, they require much more computational resources and are dataset-dependent, therefore are an approximation of the network's similarity perception. In Table \ref{tab:mobile_tcsm}, we present the values of Similarity Alignment Index between the example network's (MobileNetV2's) TNCSM generated for the training and testing set of Mini-ImageNet at different epochs of training (we use the network's feature extractor to to this). We compare the TNCSM-train and TNCSM-test matrices with each other and with the NCSM (based on weights). The results show that the alignment between TNCSM-train and TNCSM-test is very high (almost 1.0) for all epochs. It is also very high, but slightly lower between template-based and network-based matrices (as TNCSM is a dataset-dependent approximation of NCSM). Nevertheless, this alignment is still high, therefore if not possible TNCSMs can be used as a good enough (however more costly) alternative to weights-based CSM. Moreover, it is visible that the values of SAI(NCSM, TNCSM-test) and SAI(NCSM, TNCSM-train) are closer to each other in the earlier epochs of training than in the later epochs revealing the slight overfitting of the network. It suggests the possibility of using the training and testing variants of similarity-based metrics to reveal phenomena such as overfitting in networks.

\begin{table}
\centering
\caption{Mini-ImageNet: Similarity Alignment Index for different versions of direct similarity estimation measures for networks (based on the MobileNetV2 and different epochs) - rounded to 2 decimal places (values 1.0 stand for values very close to 1.0, but not exact 1). While NCSM can be obtained for the networks with classifiers, TCSM can be used as its dataset-dependent approximation. We compute two possible variants of TNCSMs for the train-test split (named in the table as -tr and -tst). }\label{tab:mobile_tcsm}
\resizebox{\textwidth}{!}{\footnotesize
\begin{tabular}{lrrrrrrrrrr}
\toprule
Data split & \multicolumn{10}{c}{Epoch} \\
\cmidrule(lr){2-11}
 &  7 & 10 & 12 & 20 & 25 & 35 & 50 & 100 & 200 & 299 \\
\midrule
SAI(NCSM, TNCSM-tr) &  0.95 & 0.95 & 0.95 & 0.95 & 0.95 & 0.95 & 0.95 & 0.94 & 0.94 & 0.93 \\
SAI(NCSM, TNCSM-tst) &  0.95 & 0.94 & 0.95 & 0.95 & 0.95 & 0.94 & 0.94 & 0.93 & 0.93 & 0.92 \\
SAI(TNCSM-tst, TNCSM-tr) &  1.00 & 1.00 & 1.00 & 1.00 & 1.00 & 1.00 & 1.00 & 1.00 & 1.00 & 1.00 \\
\bottomrule
\end{tabular}}
\end{table}

\begin{figure}[h]
\centering
\subfloat[][DINOv2]{\includegraphics[width=0.4\textwidth]{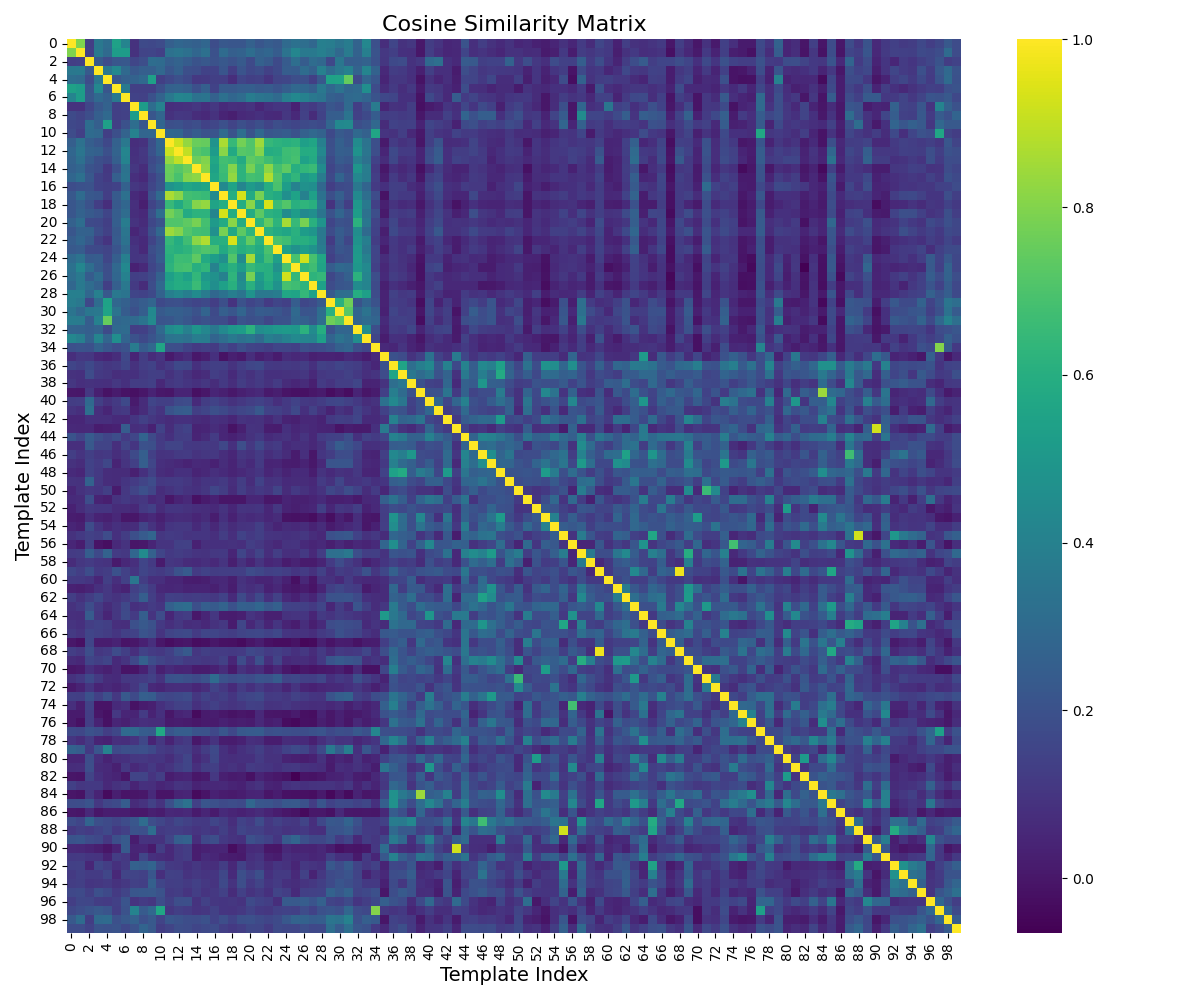}}
\subfloat[][CLIP]{\includegraphics[width=0.4\textwidth]{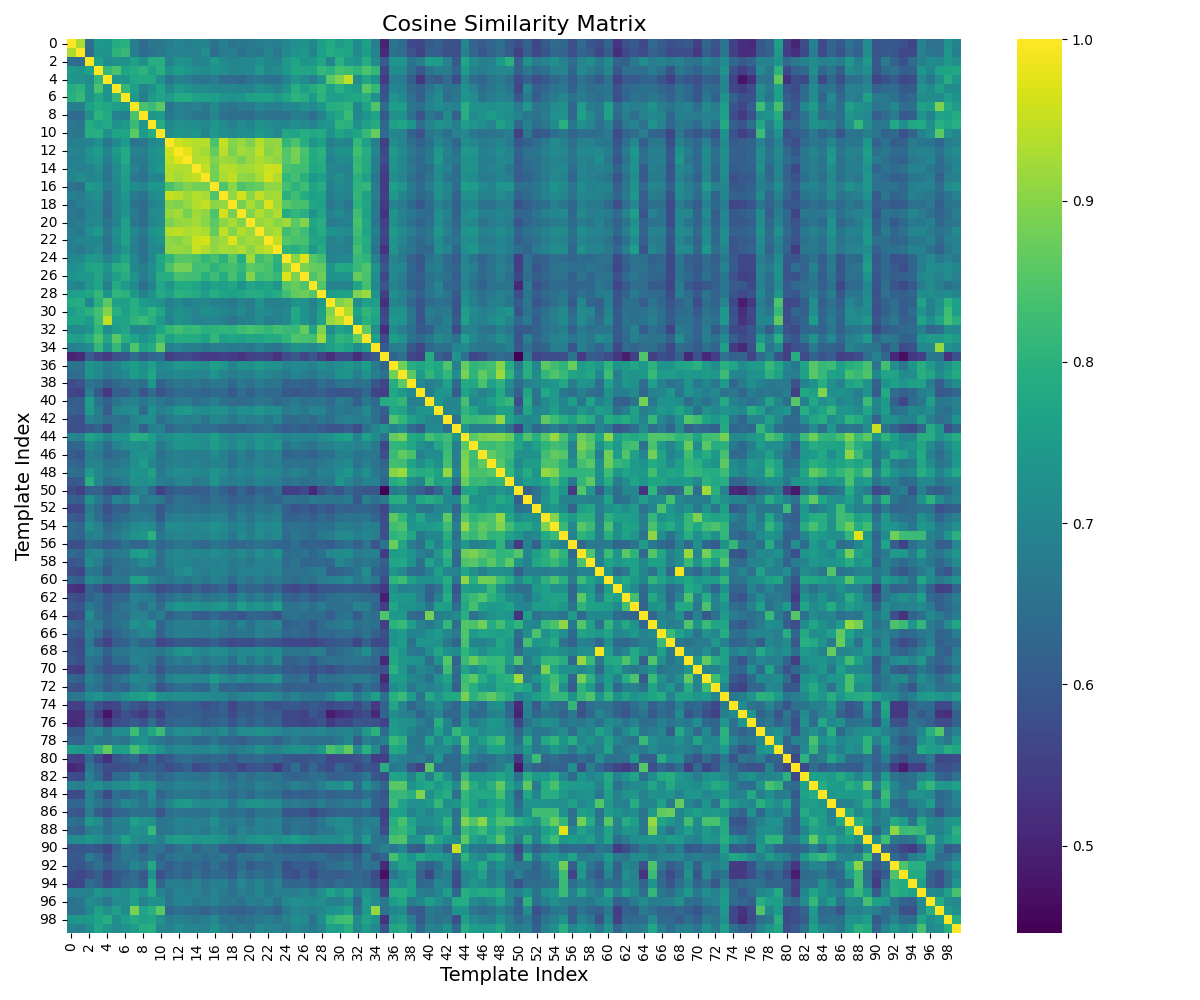}}
\caption{Templates-based Network Class Similarity Matrices obtained for DINOV2 and CLIP - networks trained with representations-focused objectives. }
\label{fig:clip_dino}
\end{figure}

In Figure \ref{fig:clip_dino}, we present the Templates-based Network Class Similarity Matrices obtained for DINOV2 and CLIP for the Mini-ImageNet testing set (the models with weights were taken from \url{https://huggingface.co/docs/transformers/model_doc/clip} and \url{https://huggingface.co/docs/transformers/model_doc/dinov2}. It is visible that the overall structure of the matrices is the same as for all the considered networks analyzed in the main body of the paper (trained for object recognition). The main difference between these two matrices is that the one for CLIP includes more similarities between less related categories (out of the main similarity groups: artifacts and animals), showing the impact of textual semantics on the similarities (\emph{e.g.} in sentences, dogs can occur in the sentence frequently close to some home objects). We also computed the numerical values of the semantic similarity alignment between the TNCSM obtained for DINOv2 and CLIP and Wordnet CSM (SAI(TNCSM, SCSM)). It is \textbf{0.85} for DINOv2 and  \textbf{0.82} for CLIP. The value obtained for DINOv2 is slightly higher than the maximum values obtained for networks and the value obtained for CLIP is similar to the majority of traditional object recognition networks used in our experiments (the highest value was app. 0.84, all higher than 0.8). This surprisingly shows that although these networks use similarity-focused techniques, the final results do not diverge very much from those obtained via traditional training schemes. In our future work, we will more deeply analyze this fascinating finding.

\clearpage

\section{Dictionary of terms used}\label{appendix_metrics}

As our framework includes many metrics, which results in the presence of numerous abbreviations in the main paper, we decided to provide a short description and additional figures to better explain the main terms, metrics and their variants used in the paper.

\paragraph{Class Similarity Matrices (CSMs)}

\begin{itemize}
    \item \textbf{NCSM} -- Network Class Similarity Matrix (based on network weights, image-free). It is created based on weights of the final classifier of the network (data-independent, direct similarity measurements).

    \begin{figure}[H]
    \centering
    \includegraphics[width=0.9\textwidth]{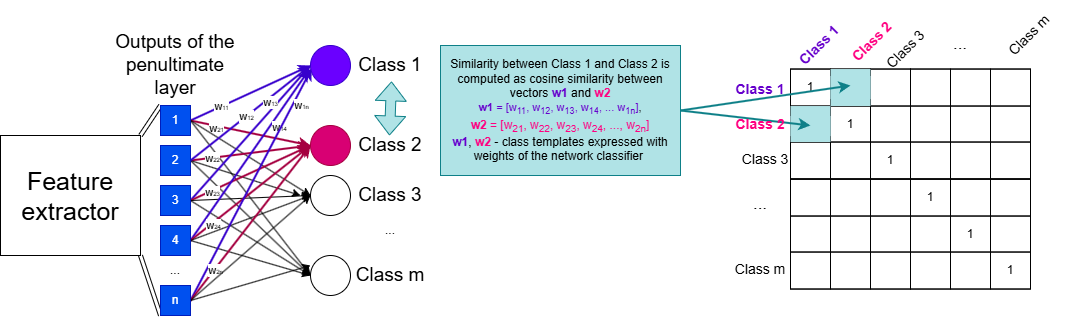}
    \caption{NCSM matrix calculation scheme.}
    \label{fig:dict_ncsm}
    \end{figure}

    \item \textbf{CCSM} -- Confusion matrix-based Class Similarity Matrix (possible train and test variants). It is created as a transformation of the confusion matrix obtained for a given network and a given dataset (data-dependent, indirect similarity measurements).

    \begin{figure}[H]
    \centering
    \includegraphics[width=0.9\textwidth]{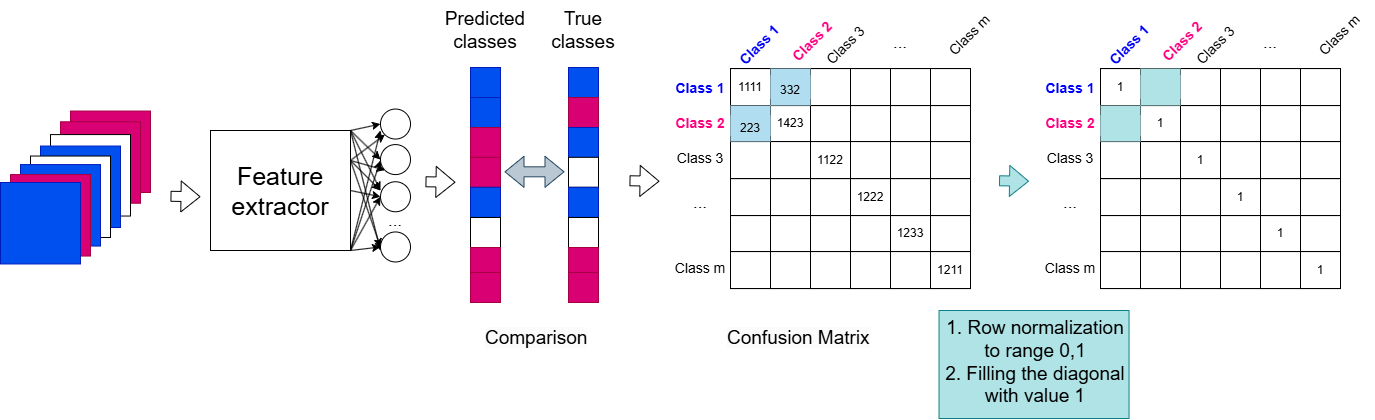}
    \caption{CCSM matrix calculation scheme.}
    \label{fig:dict_ccsm}
    \end{figure}

    \item \textbf{TNCSM} -- Image templates-based Network Class Similarity Matrix (possible train and test variants). It is created based on the features extracted by a given neural network. Features obtained for a particular class are averaged to obtain a general representation of a given class dependent on this dataset (data-dependent, direct similarity measurements).

    \begin{figure}[H]
    \centering
    \includegraphics[width=0.9\textwidth]{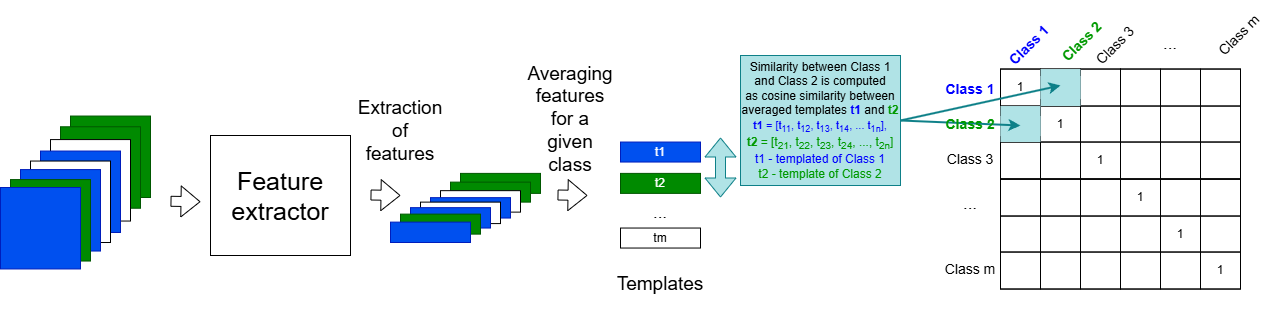}
    \caption{TNCSM matrix calculation scheme.}
    \label{fig:dict_tcsm}
    \end{figure}

    \item \textbf{SCSM/WCSM} -- Semantic/WordNet Class Similarity Matrix (As a reference, Semantic Similarity source is used). It is created based on the WordNet structure and WordNet’s similarity measure -- path (see Appendix~\ref{appendix_wordnet_accuracy} for details of path calculation). It is computed between two concepts in the WordNet tree, thus reflecting the distance between them.

    \begin{figure}[H]
    \centering
    \includegraphics[width=0.9\textwidth]{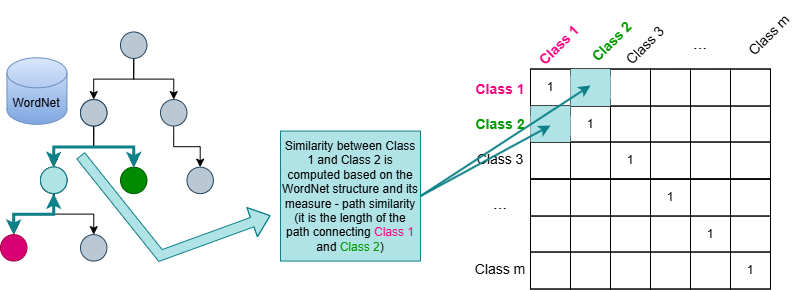}
    \caption{TCSM matrix calculation scheme.}
    \label{fig:dict_wcsm}
    \end{figure}

\end{itemize}

Note that \textbf{CCSM} and \textbf{TNCSM}, as they are the only data-dependent CSMs, are computed per dataset, therefore if train-test split is used, 2 different matrices per each CSM variant can be obtained (i.e. \textbf{CCSM-test}, \textbf{CCSM-train} and \textbf{TNCSM-train}, \textbf{TNCSM-test}).

\paragraph{Possible (Semantic Alignment Index) SAI variants}\mbox{}\\

SAI is computed as a comparison (numerical) between two similarity matrices (see Fig~\ref{fig:dict_sai} below). As their format is the same regardless of the data source used to create them, all CSM variants can be technically used for comparison.

\begin{figure}[H]
\centering
\includegraphics[width=0.9\textwidth]{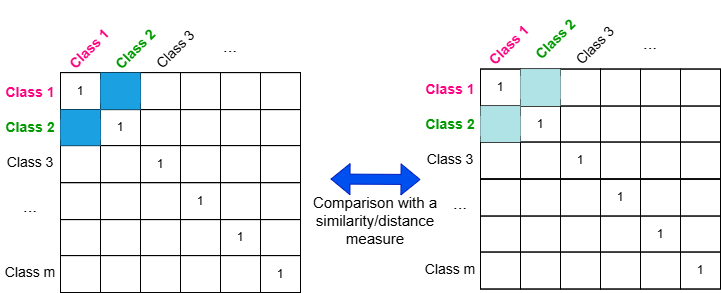}
\caption{SAI calculation scheme.}
\label{fig:dict_sai}
\end{figure}

\paragraph{SAI variants used in paper:}

\begin{itemize}
    \item \textbf{SAI(NCSM, SCSM)} -- Similarity Alignment Index (Network Class Similarity Matrix, Semantic Class Similarity Matrix) -- it determines how well the Network-perceived similarity (based on weights) aligns with semantic similarity.
    \item \textbf{SAI(CCSM, SCSM)} -- Similarity Alignment Index (Confusion-based Class Similarity Matrix, Semantic Class Similarity Matrix) -- it determines how well the Network-perceived similarity (measured indirectly based on confusion on the testing set) aligns with semantic similarity.
    \item \textbf{SAI(CCSM, NCSM)} --  Similarity Alignment Index (Confusion-based Class Similarity Matrix, Network Class Similarity Matrix) -- it determines how well the Network-perceived similarity (based on Confusion on test) aligns with the Network's direct similarity perception (based on weights).
    \item \textbf{SAI(TNCSM-train, TNCSM-test)} -- Similarity Alignment Index (Templates-based Network Class Similarity Matrix for train dataset, Templates-based Network Class Similarity Matrix for test dataset) -- it determines how similar are the Templates-based Network CSMs for the training dataset and testing dataset (measured directly based on templates extracted with a feature extractor).
    \item \textbf{SAI(TNCSM-train, NCSM)} -- Similarity Alignment Index (Tem\-pla\-tes-based Network Class Similarity Matrix for train dataset, Network Class Similarity Matrix) -- it determines how well the Network-perceived similarity (based on templates generated with the train dataset) and the Network-perceived similarity (based on weights) align. To see whether we can use them interchangeably
    \item \textbf{SAI(TNCSM-test, NCSM)} -- Similarity Alignment Index (Tem\-pla\-tes-based Network Class Similarity Matrix for test dataset, Network Class Similarity Matrix) -- it determines how well the Network-perceived similarity (based on templates generated with the test dataset) and the Network-perceived similarity (based on weights) align.
\end{itemize}

\paragraph{Inverse Dissimilarity Metric (IDM)} It is a metric that measures how far in the space defined by an NCSM’s similarity (in terms of the normalized number of classes in the matrix sorted with increasing similarity to a given class - per row) for this network are the predictions of the network from their ground truth labels. It is based on the Dissimilarity Metric (DM) introduced in \cite{bib:filus2023netsat}. The original DM can be computed in the following way. We take the ground truth label $i$ and the post-attack prediction $j$ for each image in the dataset. We check the index of $j$ in row $i$ of the Sorted Class Similarity Matrix (SoCSM).
After gathering predictions on a dataset (with perturbations), DM quantifies the harmfulness of the attack. For each image, let \(i\) represent the ground truth label, and \(j\) the post-attack prediction (in our reformulation - it is the prediction on the clear dataset). We determine the rank \(r_{ij}\) of \(j\) in row \(i\) from the Sorted Class Similarity Matrix. A higher \(r_{ij}\) indicates greater damage, as the post-attack prediction \(j\) is more dissimilar to the ground truth label \(i\) (in our variant, these higher value of this rank, means less reasonable predictions, therefore we need an additional computational step at the very end - $1 - DM$ to obtain higher values for better semantic accuracy). The computational steps of DM are as follows:
\begin{enumerate}
\item Compute the rank \(r_{ij}\) for all images in the dataset.
\item Calculate the mean rank value:
   \[
   \text{Mean Rank} = \frac{1}{N} \sum_{k=1}^{N} r_{ij}^{(k)}
   \]
   where \(N\) is the number of images in the dataset, and \(r_{ij}^{(k)}\) is the rank for the \(k\)-th image.
\item Normalize the mean rank by dividing by the total number of classes minus one:
   \[
   \text{DM} = \frac{\text{Mean Rank}}{C - 1}
   \]
   where \(C\) is the total number of classes. This normalization ensures \(\text{DM}\) being in range \(\langle 0, 1 \rangle\).
\end{enumerate}

The original's DM values can be interpreted as (in terms of the harmfulness of the adversarial attack):
\[ \begin{cases}
      1 & \text{\textit{max harmfulness}} \Rightarrow accuracy = 0\%  \\
      0\leq DM \leq 1 & \text{\textit{the higher the more harmful attack}} \\
      0 & \text{\textit{min harmfulness}} \Longleftrightarrow accuracy = 100\%
   \end{cases}
\]
As we compute the \textit{inverse} of the DM, we transform its values as $IDM = 1 - DM$. Therefore, the IDM's values can be interpreted as (in terms of the accuracy extension):
\[ \begin{cases}
      1 & \text{\textit{max semantic accuracy}} \Longleftrightarrow accuracy = 100\%  \\
      0\leq DM \leq 1 & \text{\textit{the higher the more resonable the mistakes}} \\
      0 & \text{\textit{min semantic accuracy}} \Rightarrow  accuracy = 0\%
   \end{cases}
\]
We also compute the errors-only variant of IDM, in which we only consider samples, for which a network made a mistake. We also propose to include a \textbf{new variant} of IDM - WordNet-based IDM, which uses semantic similarity instead of Network-based similarity to compute ranks.  

\end{document}